\documentclass[11pt]{report}
\usepackage{upennstyle} 
\usepackage{csquotes}
\usepackage{breakcites}
\usepackage{booktabs}
\usepackage{makecell}
\usepackage{pgfplots}
\pgfplotsset{compat=1.18}
\usepackage{pgfplotstable}
\usepackage{enumitem}
\usepackage{adjustbox}
\usepackage{tcolorbox}
\usepackage{listings}
\usepackage{hyperref}
\usepackage{amsmath}
\usepackage{bm}
\usepackage{bbm}
\usepackage{bbold}
\usepackage{algorithm2e}
\usepackage{xurl}
\usepackage{titlesec}

\newcommand{\minisection}[1]{\noindent{\bf #1}\hspace{0.6em}}
\newcommand{\openpi}{\textsc{OpenPI}\xspace}
\newcommand{\openpitwo}{\textsc{OpenPI2.0}\xspace}
\newcommand{\proc}{\textsc{Proc2Pddl}\xspace}
\newcommand{\lego}{\textsc{pddlego}\xspace}

\newcommand{\df}{$\mathbb{DF}$\xspace}
\newcommand{\pf}{$\mathbb{PF}$\xspace}

\newcommand{\pfs}{$\mathbb{PF}$s\xspace}
\newcommand{\txt}{$\mathbb{T}$\xspace}

\usepackage{color}
\definecolor{deepblue}{rgb}{0,0,0.5}
\definecolor{deepred}{rgb}{0.6,0,0}
\definecolor{deepgreen}{rgb}{0,0.5,0}
\definecolor{darkgreen}{RGB}{43,163,39}
\definecolor{bluesquare}{rgb}{126,166,224}
\definecolor{LightGray}{gray}{0.9}
\definecolor{DarkGray}{gray}{0.1}

\definecolor{lightred}{HTML}{CC8685}
\definecolor{darkred}{HTML}{FF0000}

\renewcommand{\tt}[1]{\fontfamily{cmtt}\selectfont #1}

\lstdefinestyle{pythoncode}{
	language=Python,
	otherkeywords={self,join,append,split,write},   
	keywordstyle=\bfseries\color{deepblue},
	emph={__init__},          
	emphstyle=\color{deepred},    
	showstringspaces=false,
	breaklines=true,
	escapeinside=||,
	columns=fullflexible,
	basicstyle=\fontfamily{cmtt}\small,
    belowskip=-\baselineskip,
    aboveskip=-0.7\baselineskip
}

\definecolor{codegreen}{rgb}{0,0.6,0}
\definecolor{codegray}{rgb}{0.5,0.5,0.5}
\definecolor{codepurple}{rgb}{0.58,0,0.82}
\definecolor{backcolour}{rgb}{0.95,0.95,0.92}

\usepackage[nonamebreak,round]{natbib}
\title{STRUCTURED EVENT REASONING WITH LARGE LANGUAGE MODELS}
\author{Li Zhang}
\gradgroup{Computer and Information Science} 
\date{2024} 
\supervisor{Chris Callison-Burch}
\supervisortitle{Associate Professor, University of Pennsylvania}
\gradchair{Mayur Naik, Professor, University of Pennsylvania}
\committee{Dan Roth (chair), Professor, University of Pennsylvania}
\committee{Marianna Apidianaki, Senior Researcher, University of Pennsylvania}
\committee{Mark Yatskar, Assistant Professor, University of Pennsylvania}
\committee{Rada Mihalcea, Professor, University of Michigan}
\committee{Graham Neubig, Associate Professor, Carnegie Mellon University} 

\authorlegal{Li Zhang} 

\dedication{Dedicated to Professor Dragomir Radev, whose generosity and enthusiasm drive me forward.} 

\acknowledgement{My foremost gratitude goes to Chris Callison-Burch, my advisor, who is certainly the primary factor of my success in the PhD program. I am most grateful of the academic freedom Chris gave me so that I could completely unleash my creativity, while he was always there for me in need. To me, this was truly an ideal advisor-student relationship. I learned so much from Chris and would continue his path as an admirable professor. 

I only decided to pursue a PhD after Rada Mihalcea, my former mentor, saw the potential in me. Rada’s positive influence upon me transcends research ability – she makes me a better version of myself. She taught me to be patient and self-content during my early years of urgency and anxiety. Her mindset has always driven me through the most arduous challenges.

I am forever indebted to Dragomir Radev, my first ever mentor in NLP, AI, and research. As Rada would say, Drago sees value in people when few would. I never fully understand why Drago would sit a freshman like me down in his office and teach me basic Linux commands, but I am truly glad he did. I hope I have made him proud.

I never take the unwavering support of my parents, Fan Zhang and Hong Sun, for granted. It is easier said than done to put one’s child’s happiness and well-being before their own, and I will always respect them for always being by my side.

I thank my partner Vivienne Chi, the primary source of joy in my life, not only for her companionship over my ups and downs, but also for her extraordinary wisdom of living. My time with her is like memories of pure light, which dissolve much of my anxiety and self-doubt.

I thank Veronica Lyu, Niket Tandon, and Shuyan Zhou for being amazing collaborators that worked side by side with me for multiple projects, as well as lots of inspiring conversations about life, career, and philosophy. They are the reason why my PhD journey is not lonely at all. 

I thank Peter Clark, Graham Neubig, Steven Wilson, Peter Jansen, Dan Roth, Rui Zhang, Marianna Apidianaki, and Mark Yatskar for their invaluable mentorship on different fronts at different stages of my career.

I thank Ziyang Li and Jiani Huang for being my most stalwart friends in the last five years. Playing music, petting cats, lamenting at poor reviewers, debating about nerusymbolic methods, and slurping Terakawa ramen with them - these are just some of my favorite things.

I thank Jeffrey Cho, Hainiu Xu, Joey Hou, and Tianyi Zhang for the opportunity of mentoring them. Seeing their success is often more stimulating than my own. 

I thank former and present lab members, Ellie Pavlick, Reno Kriz, Daphne Ippolito, Jie Gao, Aditya Kashyap, Artemis Panagopoulou, Yue Yang, Bryan Li, Alyssa Hwang, Liam Dugan, Ajay Patel, Samar Haider, for their mentorship, collaboration, and friendship.

I thank my collaborators and audience of The Protagonists band and Haz Studio channel for their camaraderie and the joy of music, an unexpected windfall of my PhD journey. 
} 

\abstract{Reasoning about real-life events is a unifying challenge in AI and NLP that has profound utility in a variety of domains, while fallacy in high-stake applications could be catastrophic. Able to work with diverse text in these domains, large language models (LLMs) have proven capable of answering questions and solving problems. However, I show that end-to-end LLMs still systematically fail to reason about complex events, and they lack interpretability due to their black-box nature. To address these issues, I propose three general approaches to use LLMs in conjunction with a structured representation of events. The first is a language-based representation involving relations of sub-events that can be learned by LLMs via fine-tuning. The second is a semi-symbolic representation involving states of entities that can be predicted and leveraged by LLMs via few-shot prompting. The third is a fully symbolic representation that can be predicted by LLMs trained with structured data and be executed by symbolic solvers. On a suite of event reasoning tasks spanning common-sense inference and planning, I show that each approach greatly outperforms end-to-end LLMs with more interpretability. These results suggest manners of synergy between LLMs and structured representations for event reasoning and beyond.} 

\begin{document}
\maketitle 
\setcounter{page}{2}

\makecopyright 

\makededication 

\makeacknowledgement 

\makeabstract
\tableofcontents

\clearpage \phantomsection \addcontentsline{toc}{chapter}{LIST OF TABLES} \begin{singlespacing} \listoftables \end{singlespacing}

\clearpage \phantomsection \addcontentsline{toc}{chapter}{LIST OF ILLUSTRATIONS} \begin{singlespacing} \listoffigures \end{singlespacing}


\begin{mainf} 
\chapter{INTRODUCTION}
\label{chap:intro}

Consider this one-sentence story:
\begin{displayquote}
``At the start of a band practice, the violinist was tuning her instrument, while the drummer started playing an elaborate solo.''
\end{displayquote}
This minimal but eventful narrative showcases that natural language frequently involve the communication of \textit{events}. An event is a semantic concept that describes \textit{something that happens} in the world, in some space and time (e.g., a band practice, tuning a violin, playing a drum solo). We humans possess a deep understanding of these events that transcends their face-value. For instance, we can deduce much implicit information, such as the locale (practice room), participants (musicians), their intents (prepare for a show), preceding events (arrival at practice), involved entities (violin, drumsticks), and so on. While some of the above information might be obvious, obtaining others requires \textit{reasoning}. For example, if one were to ask about the outcome of the above event, in the form of a question:
\begin{displayquote}
``What would the violinist likely say in this situation?''
\end{displayquote}
A human who has never been asked this question before but has some background knowledge can likely answer correctly, for example, by making the following deductions:
\begin{enumerate}
    \item It is common knowledge that drum playing is loud.
    \item It is common knowledge that violin tuning requires a quiet environment.
    \item Based on 1 and 2, the violinist cannot effectively tune.
    \item Therefore, the violinist will be annoyed.
\end{enumerate}
Eventually, they may arrive at a possible answer:
\begin{displayquote}
“Could you wait until I finish tuning?”
\end{displayquote}
In this process, the human is not only drawing upon their knowledge of the world, but also systematically integrating pieces of such knowledge to arrive at the correct answer.  Naturally, such an ability is highly sought after in artificial intelligence (AI), machine learning (ML) and natural language processing (NLP) systems.

However, if the same question is posed towards ChatGPT\footnote{A state-of-the-art AI model at the time of writing (\url{chat.openai.com}); the experiment was done in December 2023; response may vary due to the model's non-deterministic nature.}, its response is:
\begin{displayquote}
“That was an amazing solo! You really know how to rock it!”
\end{displayquote}
an unlikely response from any violinist in this situation. Notwithstanding the rapid advancement of models like ChatGPT trained on massive online data, failures like such are still common when reasoning about events, especially those that are less commonly represented in the training data (i.e., the \textit{long-tail problem}). Apart from the toy example above, the lack of reasoning ability in AI models can be catastrophic in high-stake applications. As technology continues to advance, it is increasingly likely that a user will rely on AI models for legal, financial, or medical advice. In these scenarios, each individual's case would be significantly different from and more complex than the available training data. If the model fails to generalize and reason correctly, it will suggest unreasonable actions, leading to a loss of freedom, property, or even life. Less drastically, failure to reason like humans do would result in a lack of trust, where users do not feel confident making decisions using AI models. 

For decades, the AI, ML, and NLP communities have been honing the ability of intelligent systems to reason over events. Historically, such reasoning was performed using symbolic methods. These are highly interpretable but require domain-specific annotations which can be hard to generalize to new domains, especially to the long-tail problems. In contrast, modern efforts have favored data-driven, neural methods that can effectively adapt to domains that are sufficiently represented in the training data, but still struggle with out of domain or out of distribution situations. Moreover, neural methods provide little in the way of interpretability, nor can their output be easily verified and improved, leading to a lack of trustworthiness. 

In my work, I attempt to combine the best of both worlds by reasoning about events using a combination of neural and symbolic methods. I extensively leverage large language models (LLMs), arguably the most powerful tool for most text-based applications. Instead of using LLMs in an end-to-end fashion, I propose a neurosymbolic synergy that combines LLMs with a \textit{structured representation} of events. I explore different forms of such representation as well as different ways to integrate it with LLMs through a variety of downstream tasks (Figure~\ref{fig:thesis_work} and Figure~\ref{fig:usages}). 

\begin{figure}
    \centering
    \includegraphics[width=\columnwidth]{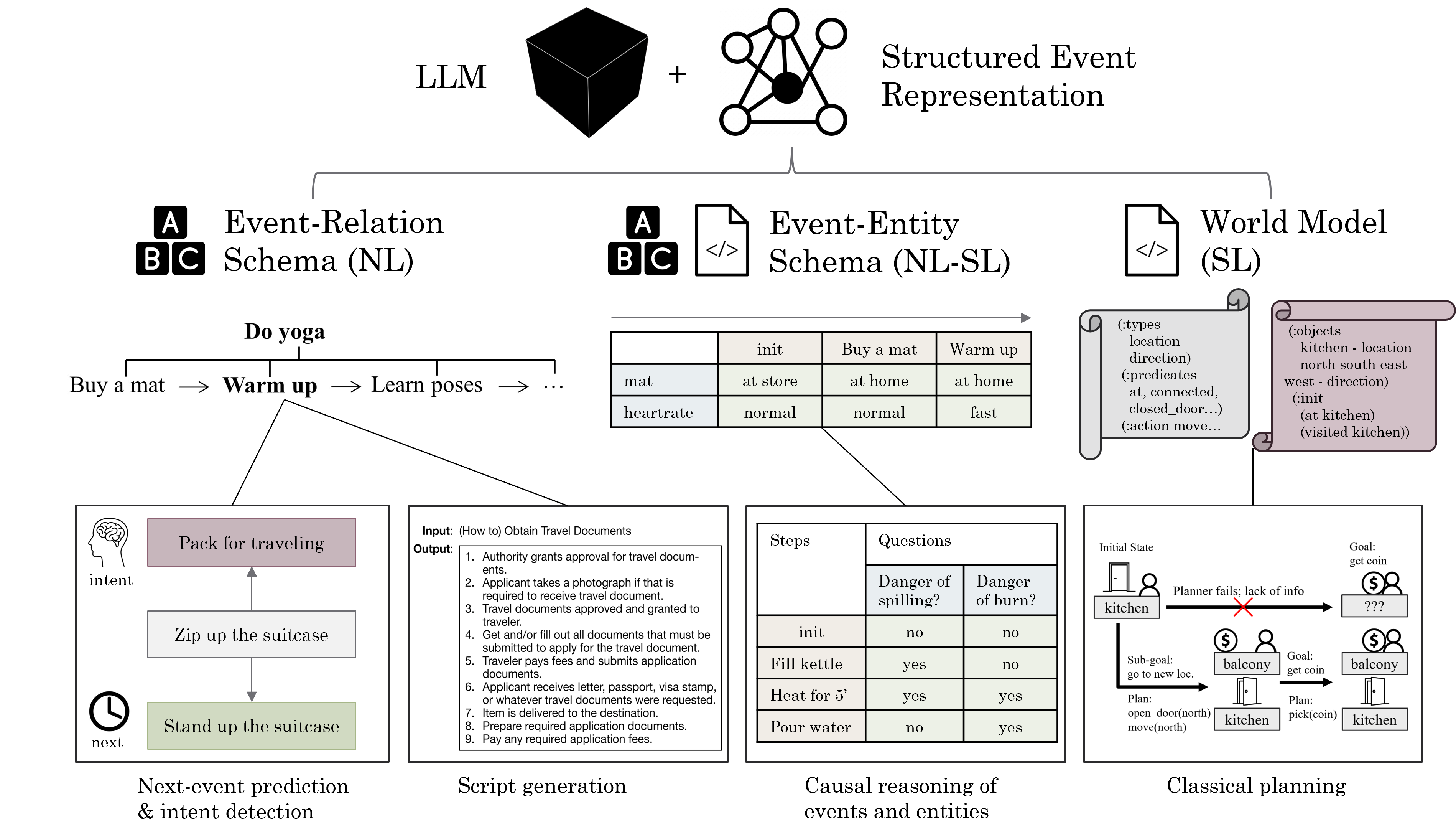}
    \caption{An overview of work discussed in this thesis.}
    \label{fig:thesis_work}
\end{figure}

\begin{figure}[t!]
    \centering
    \includegraphics[width=0.9\columnwidth]{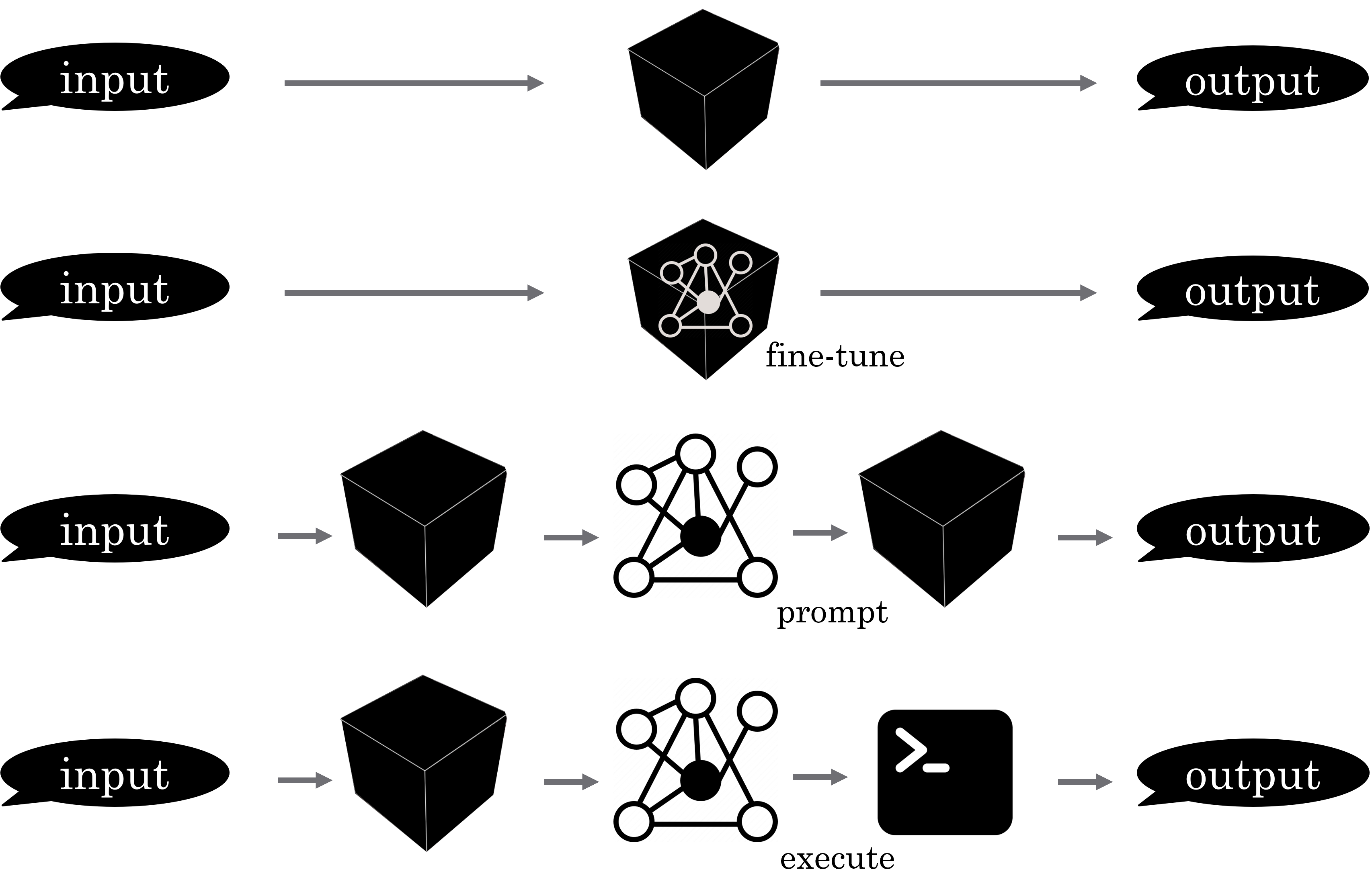}
    \caption{Four approaches of using LLMs for structured reasoning: (from upper to lower) end-to-end usage (no structure), fine-tuning with structure, prompting with structure, and neurosymbolic usage.}
    \label{fig:usages}
\end{figure}

In natural language, an event is often described in relation to other events. Therefore, in Chapter~\ref{chap:relation}, I study a natural language representation by decomposing a complex event into relations of its sub-events. For example, the event ``\textit{do yoga}'' can be represented as a sequence of its sub-events, ``\textit{buy a mat}'', ``\textit{warm up}'', ``\textit{learn poses}'', and so on. The collection of these sub-steps is essentially a procedure with the goal ``\textit{do yoga}'', while each can be seen as a step, thus forming a goal-step relation. Additionally, any pair of two steps specifies the order in which they happen, thus forming a step-step temporal relation. These two relations constitute an \textbf{event-relation schema}, where each event is still expressed with a natural language phrase, while the relations of them are modeled in a structured fashion. 

With these event relations explicitly modeled, I now explore a way to inject such a structured representation into LLMs. In Section~\ref{sec:learn_relations}, I fine-tuning LLMs on a dataset targeting the two event relations. Taking the goal-step relation as an example, a model is given a goal ``\textit{do yoga}'' and must infer the most reasonable step among candidates ``A) \textit{buy a suitable mat}'', ``B) \textit{buy some weights}'', ``C) \textit{deep clean a mat}'', and ``D) \textit{buy a house}.'' While for humans the answer is trivially A, such prediction was non-trivial for LLMs at the time. To train LLMs, I construct a multiple-choice dataset using procedural data from the web. To ensure that the dataset is both clean and challenging, I devise a negative sampling strategy based on LLM-based word vectors to select candidates that are semantically similar but not identical. The resulting dataset is manually verified to have high quality and shows that LLMs are able to learn these event relations, though there is still a performance gap compared to humans.

To ascertain whether the event-relation schema is useful in downstream applications, I apply the previously fine-tuned LLMs on two reasoning tasks in Section~\ref{sec:relation_downstream}. The first is next-event prediction, a task of commonsense inference. For example, given an event ``\textit{the pianist rest her fingers on the keys}'', a model must choose the most reasonable subsequent event among ``\textit{she started playing}'', ``\textit{she stood up}'', ``\textit{she chuckled nervously}'', and ``\textit{she sang a marvelous tune}.'' The second is intent detection, a task that dialog systems like Alexa would need to perform to select the appropriate subroutine. For example, given an utterance ``\textit{I want to make restaurant-quality fried rice}'', a model must choose the most intent among ``\textit{find a recipe}'', ``\textit{recommend a restaurant}'', and so on, before it can call upon an API to fulfill the request. On an array of datasets on both tasks, the models fine-tuned with goal-step and step-step temporal relations, respectively, greatly outperform end-to-end LLMs in few-shot settings. This result demonstrates the benefit of a structured event representation in low-data, long-tail scenarios.

After showing the utility for each individual event relation, I attempt to use both event relations in the event-relation schema at once. In Section~\ref{sec:script_generation}, I tackle the challenging task of script generation. Given a goal such as ``\textit{obtain travel documents}'', the model must generate all the reasonable steps that are not only pertinent but also correctly ordered, such as ``\textit{prepare materials}'', ``\textit{pay fees}'', ``\textit{submit application}'', and so on. I modularize the task into two sequential stages in a pipeline, each tackled by one of the LLMs fine-tuned with a relevant event relation. By both automatic and manual evaluation of the generated scripts, I again show that structured approach is superior to the end-to-end one on a variety of languages. These experiments collectively take advantage of the flexibility of a natural language representation of events and the specificity of the underlying structure, namely the event relations. Despite the improved performance, a natural language representation that interacts with LLMs via fine-tuning lacks trustworthiness. An end-user would have little idea of how the LLMs produce the output, nor can they effectively improve or correct the models without extensively modifying the data.

Previously, I have chosen to express events using natural language because of their richness and volatility in real-life texts. In comparison, the entities involved in the events are much more finite and grounded. Therefore, in Chapter~\ref{chap:entity}, I switch to a semi-symbolic representation by decomposing a complex event into dynamic state changes of entities. For example, the event ``\textit{do yoga}'' can be represented as the states of involved entities: the \textit{mat} was \textit{in the store} before the step ``\textit{buy a mat}'', but \textit{at home} afterwards. The three-dimensional matrix with the axes being the steps, entities, and attributes and the value being the states is referred to as an \textbf{event-entity schema}. Although each entity is superficially expressed with a natural language phrase, they are finite and can be grounded in an embodied environment.

How can one equip LLMs with the knowledge of entities, just like that of event relations discussed above? In Section~\ref{sec:openpi}, I contribute in constructing a dataset of entity state tracking in procedural texts. I facilitate the grounding of entities by canonicalizing them into symbols. Specifically, this entails clustering their mentions. For example, \textit{coffee maker}, \textit{espresso dispenser}, or simply \textit{machine} may refer to the same entity in the procedure of ``\textit{make coffee with an espresso machine}.'' This task is much more challenging than typical coreference resolution or paraphrase detection because procedural texts are highly contextual. By prompting state-of-the-art LLMs with in-context learning, I fully canonicalize an entity tracking dataset that can evaluate entity state tracking models more fairly. These models can thus predict an event-entity schema given a procedure. 

Intuitively, the states of entities causally give rise to the occurrence of events. To see if LLMs can similarly leverage the event-entity schema to make predictions about events, in Section~\ref{sec:crepe}, I define the task of causal reasoning task of events and entities. For example, over the course of ``\textit{boiling water}'', it will be \textit{dangerous to overturn a kettle} when the kettle is filled and heated up, but not before the kettle is heated or after the water is poured out from the kettle. The above judgement causally hinges on whether the kettle contains hot water. Naturally, I leverage a predicted event-entity schema by providing it to LLMs as an in-context prompt. Notwithstanding the symbolic nature of the schema, I find that a pseudo-code form (one that expresses events and entities as Python functions and class objects) respectively achieves much better performance than a natural language form (e.g., at the end of step ``\textit{heat up the kettle}'', the \textit{kettle} is \textit{filled with hot water}). In addition to illustrating the potency of the event-entity schema, this observation leaves the open-question of whether such a code-like form better evokes the reasoning ability of modern LLMs trained on a mixture of code and text.

To answer this question, in Section~\ref{sec:symbolic_form}, I design prompts, both in a typical natural language or textual form and the previously proposed symbolic language or code-like form, on a dozen of general NLP tasks beyond event reasoning. However, despite some contemporaneous work that argues for the idea, I observe no conclusive trend. Nevertheless, regardless of which form to take, the event-entity schema assumes a symbolic nature, and is proven effective when fed to LLMs as in-context input. This mechanism of synergy between the structured representation and LLMs is highly flexible, and provides more interpretability and user control over the fine-tuning mechanism in the last Chapter. 

In the two previous Chapters, my proposed structured representation, either in natural or symbolic language, is eventually provided to LLMs as input. However, I have observed that as the reasoning task becomes more complex, LLMs fall short and might not be the optimal tool. In Chapter~\ref{chap:world}, I introduce an alternative, neurosymbolic methodology by which the structured representation and LLMs interact. Instead of feeding the LLM-generated symbolic representation into another LLM, I rather use an algorithmic solver that can parse and execute it. Because the solver is both deterministic and well-constructed, its output is guaranteed to be correct provided that the input symbolic representation is correct. This way, LLMs are relieved of the task of problem solving, but are only responsible for generating the correct interim representation of the context, or namely a \textbf{world model}. By construction, this approach is more trustworthy than the two previous alternatives, in that a user can verify, interpret, and correct the output by interacting with the structured world model. Because the world model is no longer input to LLMs but rather to symbolic solvers, the representation also must be fully symbolic. In other words, the LLM that generates the world model must ground the involved concepts (e.g., events, entities, etc.) to some provided environment such that they can be consumed by the symbolic solver. 

An ideal downstream task to instantiate the above idea is classical planning, explored in Section~\ref{sec:planning}. At any point of time, a state of the environment is symbolically defined by a collection of entity states (e.g., \texttt{in(you, kitchen)}, \texttt{closed\_door(kitchen,balcony)}, \texttt{in(coin,balcony)}. Also defined is a domain model of permitted actions with pre-conditions and effects. Collectively, these two pieces of information constitute a world model. The goal of classical planning is thus to find a sequence of such actions that drive the state of the environment in a given way. In the real life, the environment is rarely defined symbolically, but in natural language (e.g., you are in the kitchen, the kitchen is connected to the balcony by a closed door, a coin is on the balcony, here are a list of actions you can perform...). Given such a description, my approach is not to have LLMs generate a plan, but to generate a representation in planning domain definition language (PDDL) that can be deterministically solved. 

Can LLMs indeed generate a symbolic world model given textual description? 
In Section~\ref{sec:plan_procedure}, I focus on planning for procedural texts. Given a textual description as above, I use LLMs to generate a domain model before feeding it into a PDDL solver to arrive at the plan. Even though the LLMs' task is reduced to just translating the environment description to a world model but not actual planning, such generation turns out to be highly challenging and even impossible for many LLMs given their weakened ability to generate low-resource, domain-specific languages. Using a combination of approaches to modularize the prompting process, I end up with a model that can generate solvable domain model more than 30\% of the time. 

In Section~\ref{sec:plan_interaction}, I shift my focus to planning for interactive textual environments that emulate a robotic application. Here, the environment itself is a symbolically defined state-transition model. Therefore, my approach is to learn a world model through exploration and interaction with the environment. However, unlike the previous setup where the entirety of the environment has been described, many entity states are initially unobserved. Thus, the world model cannot be completely learned, and a plan cannot be derived. To tackle this challenge, I decompose the end goal into sub-goals that \textit{can} be planned for via the current partial world model. Through making progress towards the sub-goals, more of the environments are explored, resulting in a more complete world model, that eventually can support planning for the end goal. I show that this neurosymbolic approach is drastically more effective than having end-to-end LLMs predicting the plan in various interactive environments.

In summary, in this thesis, I introduce three structured event representations: a language-based event-relation schema, a semi-symbolic event-entity schema, and a fully-symbolic world model. I explored ways that these representations may interact with LLMs, including fine-tuning with them, in-context learning with them, and generating them to be solved symbolically. On a variety of downstream tasks of event reasoning, I show that my proposed approaches are superior to end-to-end LLMs, both on performance and trustworthiness.




\newpage
\section{Thesis Statement}
I argue that using LLMs in conjunction with structured representations of events lead to improved performance in reasoning tasks. This thesis focuses on three types of such structured representations: (1) a language-based event-relation schema, (2) a semi-symbolic event-entity schema, and (3) a fully-symbolic world mode. Through experiments on a variety of approaches to leverage these representations and a diverse set of downstream tasks, I show that the benefit of structured event reasoning using LLMs.
\newpage
\section{Publications Presented}
The work described in this thesis has been published as the following conference papers in the Association for Computational Linguistics (ACL) Anthology\footnote{\url{https://aclanthology.org/}}:

(*equal contribution; $^\dagger$mentored student)

\begin{enumerate}
    \item Zhang, L.*, Lyu, Q.*, and Callison-Burch, C. (2020b). Reasoning about goals, steps, and temporal ordering with WikiHow. In Proceedings of the 2020 Conference on Empirical Methods in Natural Language Processing (EMNLP), Online. Association for Computational Linguistics.
    \item Zhang, L., Lyu, Q., and Callison-Burch, C. (2020a). Intent detection with WikiHow. In Proceedings of the 1st Conference of the Asia-Pacific Chapter of the Association for Computational Linguistics and the 10th International Joint Conference on Natural Language Processing, Suzhou, China. Association for Computational Linguistics.
    \item Lyu, Q.*, Zhang, L.*, and Callison-Burch, C. (2021). Goal-oriented script construction. In Proceedings of the 14th International Conference on Natural Language Generation, Aberdeen, Scotland, UK. Association for Computational Linguistics.
    \item Zhang, L.*, Xu, H.*$^\dagger$, Yang, Y., Zhou, S., You, W., Arora, M., and Callison-Burch, C. (2023). Causal reasoning of entities and events in procedural texts. In Findings of the Association for Computational Linguistics: EACL 2023, Dubrovnik, Croatia. Association for Computational Linguistics.
    \item Zhang, L.*, Dugan, L.*, Xu, H.*$^\dagger$, and Callison-Burch, C. (2023). Exploring the curious case of code prompts. In Proceedings of the 1st Workshop on Natural Language Reasoning and Structured Explanations (NLRSE), Toronto, Canada. Association for Computational Linguistics.
    \item Zhang, L., Xu$^\dagger$, H., Kommula, A., Zhou, Callison-Burch, C, and Tandon, N. (2024). OpenPI2.0: An Improved Dataset for Entity Tracking in Texts. In Proceedings of the 18th Conference of the European Chapter of the Association for Computational Linguistics, St. Julians, Malta. Association for Computational Linguistics.
    \item T. Zhang*$^\dagger$, L. Zhang*, Z. Hou, Z. Wang, Y. Gu, P. Clark, C. Callison-Burch, and N. Tandon. PROC2PDDL: Open-Domain Planning Representations from Texts. In Proceedings of the 2nd Workshop on Natural Language Reasoning and Structured Explanations (NLRSE), Bankok, Thailand. Association for Computational Linguistics.
    \item L. Zhang, P. Jansen, P. Clark, C. Callison-Burch, and N. Tandon. PDDLego: Iterative Planning in Textual Environments. In Proceedings of the 13th Joint Conference on Lexical and Computational Semantics (*SEM 2024). Association for Computational Linguistics.
\end{enumerate}

These publications are licensed under Creative Commons 4.0 BY, listed below. Therefore, parts of the relevant discussions are quoted directly from said publications, with the explicit approval of all co-authors, my thesis committee, and the Graduate Group Chair. None of these publications have been or will be extensively discussed in any of my collaborators' theses. All work was completed jointly with collaborators at the University of Pennsylvania, Carnegie Mellon University, and Allen Institute for Artificial Intelligence. At the end of each chapter, I summarize my primary contribution to the work. Refer to the above publications for more details about each described work. 

\chapter{RELATED WORK}
Assuming basic knowledge of machine learning, this thesis pertains to three concepts:
\begin{enumerate}
    \item Events and procedures, the task
    \item Large language models, the tool
    \item Machine reasoning, the framework
\end{enumerate}
This chapter provides background knowledge to these concepts, focusing on empirical rather theoretical aspects.

\newpage

\section{Events and Procedures}
\label{sec:related_work_events_procedures}

\subsection{Events}

\textbf{Event} is a historic and important concept in linguistics and NLP. In this thesis, an event is something that \textit{happens} at some space and time. For example, ``play drums'', ``jazz concert'', ``starts raining'', or ``glacier movement'' are all natural language expressions that describe events, ubiquitously in texts. Regardless of whether these are noun phrases or verb phrases, the common thread is that all of them can \textit{happen}. Some non-examples include ``drummer'', ``venue'', ``raindrop'', or ``Eurasian Plate''. These cannot \textit{happen}, and are in fact \textbf{entities} that \textit{participate} in events, which we will study in more details later. 

In literature, the concept of event has been thoroughly studied, leading to many possible definitions. In NLP, the closest neighboring field is semantic role labeling (SRL), the task of assigning roles to participants (similar to entities) according to a predicate (similar to events) \citep{10.5555/555733}. The definitions of events in SRL is similar to what I adopt, but I do not focus  exclusively on the arguments of events. In theoretical linguistics and formal semantics, the event semantics have a history of decades \citep{tenny2000history,maienborn2011semantics}. One of the most acceptable formalism of events is the Davidsonian event semantic \citep{davidson1967logical}, defining events as ``concrete particulars with a location in space and time.'' Later in this thesis, I will show some work that targets the time-feature of events. The mainstream event formalism was later replaced by the Neo-Davidsonian formalism \citep{higginbotham1985semantics}, which expands the definition of events on several fronts (such as not restricting events to verb phrases. Later, this expanded definition also encompasses processes (or procedures) and states, which I will discuss in details \citep{mourelatos1978events, bach1986algebra}. In this thesis, I do not focus on any particular formalism, but intentionally take on a loose and pragmatic definition with the purpose of solving problems and tackling tasks in NLP.

Events can be multimodal (e.g., ``alarm clock ringing'' can not only be described, but also be seen and experienced), but I will only discuss them in textual terms under the context of NLP. Because events are a semantic construct, the study of natural language events often falls under the umbrella disciplines of semantics or natural language understanding. As an example of a task that is out of my scope, part-of-speech tagging of the phrase ``alarm clock ringing'' is not in our scope because it does not pertain to semantics, nor is translating this phrase to French because it does not focus on any particular feature of an event.

My work primarily falls under the sub-field of \textit{event-centric NLP} \citep{chen-etal-2021-event}, which includes efforts that span a number of fronts. To start with, information extraction or semantic parsing \citep{mausam-etal-2012-open,liu-etal-2018-jointly,yang-etal-2019-exploring,du-cardie-2020-event}, the task of extracting relational tuples from texts, is largely concerned with events. For example, for the text ``a pirate ship was sunken by the coastal guards'', a model might need to identify an \textit{attack} event with the guard and pirate as the \textit{belligerents}. Typically, systems perform information extraction with a given ontology, specifying the information of interest. In case of event extraction, there is often a categorization of events (e.g., material, vehicle, weapon) and parameters associated with each (e.g., agent, beneficiary, manner). In my work, I do not attempt to extract events from texts, but rather assume that they are readily available either via web resources or crowdsourced datasets. Instead, I focus on applying those available event information to downstream applications. However, the formulation of relational tuples aligns with my notion of structured representation, especially those described in Section~\ref{chap:relation}. Notably, unlike typical information retrieval tasks, I do not assume any given ontology with regard to said relational representation.

Similarly, much work has focused on tracking and summarizing events over time \citep{allan1998topic,laban-hearst-2017-newslens,saravanakumar-etal-2021-event}. For example, over the \textit{COVID pandemic} that lasted years, a line of events such as \textit{universities shut-down}, \textit{stay-at-home ordinance}, \textit{dispensing sanitizers}, \textit{masking}, \textit{vaccination} can be collectively summarized as the precautions that were adopted over time. Similar to this line, my work also emphasizes a collection of events bound by temporal relations instead of singular events. In contrast to referenced work that primarily studies events in news, my work focuses on procedures (discussed later), in which events happen in a much smaller scale, but are more intentional and tighter-knit in a more qualified environment. Similar to the discussion on information extraction, this line of work is often concerned with extracting information from unstructured text, while my work is not.

Many efforts branded as {\it commonsense inference} \citep{mostafazadeh-etal-2016-story,zellers-etal-2018-swag,zhou-etal-2020-temporal} mainly target event reasoning, the family of tasks that are core to this thesis. These tasks are diverse in nature. For example, the next-event prediction task might provide an event such as \textit{the host taps on the glass in a banquet hall} as input, and a model needs to select the most reasonable subsequent event from \textit{everyone goes silent}, \textit{the guests start singing}, \textit{the crowd bursts into a roar of laughter}, and so on. Because `commonsense' encompasses additional reasoning beyond the event reasoning that I focus on --including things like social commonsense or physical commonsense--, I refrain from this term. While my exact notion of reasoning will be further defined in Section~\ref{sec:machine_reasoning}, this line of work is concerned with different aspects of events described in the beginning of Chapter~\ref{chap:intro}. Many of the involved datasets are also discussed and used in following sections.

A \textit{script} is a standardized sequence of events about stereotypical activities \citep{feigenbaum1981handbook}. For example, ``\textit{go to a restaurant}'' typically involves ``\textit{order food}'', ``\textit{eat}'', ``\textit{pay the bill}'', etc. Such script knowledge has long been proposed as a way to enhance AI systems \citep{schank_scripts_1977}. Most work in script learning focuses on narrative scripts, where declarative or descriptive knowledge is distilled from narrative texts like news or stories \citep{mujtaba2019recent}. Such scripts are descriptions of sequential events (e.g. a traffic accident involves a collision, injuries, police intervention, etc.). \citet{chambers-jurafsky-2008-unsupervised} introduced the classic Narrative Cloze Test, where a model is asked to fill in the blank given a script with one missing event. Following the task, a few papers made extensions on representation \citep{chambers-jurafsky-2009-unsupervised, pichotta-mooney-2014-statistical} or modeling \citep{jans-etal-2012-skip, pichotta-mooney-2016-statistical,pichotta-mooney-2016-using}, achieving better performance on Narrative Cloze. Meanwhile, other work re-formalized Narrative Cloze as language modeling (LM) \citep{rudinger-etal-2015-script} or multiple-choice \citep{granroth2016happens} tasks. Alternative to the narrative scripts, procedural scripts are those whose events are unified under a common goal. Those are equivalent to procedures and will be discussed in details later. In Section~\ref{sec:script_generation}, I tackle the task of generating a procedural script.

To summarize, the event-centric NLP targets the special semantics of events, which span various conventional sub-fields in NLP. Zooming into events has much practical impact financially \citep{ding2015deep}, scientifically \citep{berant-etal-2014-modeling}, and medically \citep{zhang2020diagnostic}. Naturally, it has attracted much attention in the community in the recent years.

\subsection{Procedures}
Most of my thesis work focuses on \textit{procedures}, a concept that may be thought of as a subset of events. A procedure, or a process, is a compound event, (e.g., ``learn about NLP''), which can be broken down into multiple events (e.g., ``read papers'', ``take classes'', ``attend seminars'', ...). A procedure consists of a \textbf{goal} (or intent, motivation) event, and some \textbf{step} events to achieve this goal. Studying procedures has many additional benefits than studying general events \citep{li-tutorial}, especially from a human-centered perspective, placing an emphasis on humans' behavior and cognition.

The earliest work on procedures in NLP dates back to \citet{miller1976natural,momouchi-1980-control}. At that time, procedural understanding is closely tied to AI \textit{planning} \citep{schank_scripts_1977}. Specifically, a substantial body at that time focused on natural language generation from plans \citep{mellish1989natural, wahlster1993plan}, many relying on instructional texts as data \citep{kosseim1994content,paris1995support}, which are structured and limited in scope. However, such work did not attempt learning from procedural texts, until later when \citep{10.1145/584955.584977} created a human-in-the-loop tool for procedural knowledge acquisition. Note that such knowledge extracted from procedural texts can be transformed to plans (the reverse of ``natural language generation from plans'' mentioned above), which is a substantial body of subsequent research \citep{macmahon2006walk, branavan-etal-2009-reinforcement,10.5555/2900423.2900560,artzi-zettlemoyer-2013-weakly,kiddon-etal-2015-mise}. One primary use of such procedural language generation is to answer how-to questions, the second most sought-after type of queries on the internet at that time \citep{deRijke2005QuestionAW}. To that end, there were an array of efforts to identify \textbf{instructional texts} on the web \citep{takechi-etal-2003-feature}, automatically generate them \citep{paris_automatically_2005}, converting them to executables \citep{gil2011tellme,fritz2011formal}, study their linguistic idiosyncrasies \citep{https://doi.org/10.1111/0824-7935.00118,bielsa2002semantic,aouladomar2005preliminary,10.1145/2531920}, and extract components such as titles \citep{delpech-saint-dizier-2008-investigating}. 

Much like information extraction in events discussed above, much work extracts structured knowledge from procedures \citep{lau2009interpreting,Addis2011FromUW}. \citet{zhang-etal-2012-automatically} proposes a standard representation of procedures that emphasizes goals, steps, pre- and post-conditions, much like my formulation in this thesis (Section~\ref{sec:learn_relations} and Section~\ref{sec:planning}), but was much simpler and set the basis of procedural representation in subsequent research. \citep{maeta-etal-2015-framework, kiddon-etal-2015-mise} took a different approach, and focused on the relations among entities, which are also a center piece in information extraction \citep{doddington-etal-2004-automatic,ellis2014overview}. The entity-based representation heavily inspires my work in Chatper~\ref{chap:entity}.

Under the context of script learning, a line of work focuses on procedural scripts, where events happen in a scenario, usually in order to achieve a goal. For example, to ``visit a doctor'', one should ``make an appointment'', ``go to the hospital'', etc. To obtain data, Event Sequence Descriptions (ESD) are collected usually by crowdsourcing, and are cleaned to produce procedural scripts. Thus, most such datasets are small-scale, including OMICS \citep{singh2002open}, SMILE \citep{regneri-etal-2010-learning}, the \citet{li2012crowdsourcing} corpus, and DeScript \citep{wanzare-etal-2016-crowdsourced}. The evaluation tasks are diverse, ranging from event clustering, event ordering \citep{regneri-etal-2010-learning}, text-script alignment \citep{ostermann-etal-2017-aligning} and next event prediction \citep{nguyen-etal-2017-sequence}. There are also efforts on domain extensions \citep{yagcioglu-etal-2018-recipeqa, berant-etal-2014-modeling} and modeling improvements \citep{frermann-etal-2014-hierarchical, modi-titov-2014-inducing}. All of the above are possible candidates for the data source to study procedures. 

In our work, the primary source of procedural data is \textbf{wikiHow}\footnote{\url{wikihow.com}} (previously eHow\footnote{\url{https://www.wikihow.com/wikiHow:History-of-wikiHow}}), a website of how-to instructions for many tasks. As the structure and writing style are consistent, wikiHow has been leveraged by NLP researchers since its inception \citep{10.1145/988672.988750,Addis2011FromUW,10.1145/2567948.2578846}. In recent years, wikiHow has grown massively in size (now more than 110k articles), diversity (19 languages, hundreds of categories) and quality (editorial process\footnote{\url{https://www.wikihow.com/wikiHow:Delivering-a-Trustworthy-Experience}}). As a result, it has fueled much research on procedures from various aspects \citep{pareti2018representation}, such as linking actions \citep{pareti2014integrating,10.1145/2872518.2890585,lin-etal-2020-recipe,donatelli-etal-2021-aligning,zhou-etal-2022-show}, what-if reasoning \citep{tandon-etal-2019-wiqa,rajagopal-etal-2020-ask}, entity tracking \citep{tandon-etal-2020-dataset}, next-event prediction \citep{nguyen-etal-2017-sequence,zellers-etal-2019-hellaswag,zhang-etal-2020-analogous}, intent reasoning \citep{dalvi-etal-2019-everything}, goal-step reasoning \citep{zhou-etal-2019-learning-household,park2018learning,zhang-etal-2020-reasoning,yang-etal-2021-visual}, procedure generation \citep{sakaguchi-etal-2021-proscript-partially,lyu-etal-2021-goal}, simulation \citep{puig2018virtualhome}, summarization \citep{DBLP:journals/corr/abs-1810-09305,ladhak-etal-2020-wikilingua} and so on.

As more work has explored procedures \citep{9070972}, research agendas become more diverse. 

\newpage

\section{Large Language Models}

\subsection{A Brief History of Language Models}
\label{sec:llm_history}
For a complete survey on LLMs and their predecessors, readers are redirected to \citet{zhao2023survey}. In essence, a language model (LM) is one that assigns generative likelihood of vocabulary in a passage. Its revolution over the past decades might be summarized as the following four phases.

\begin{figure}
    \centering
    \includegraphics[width=\columnwidth]{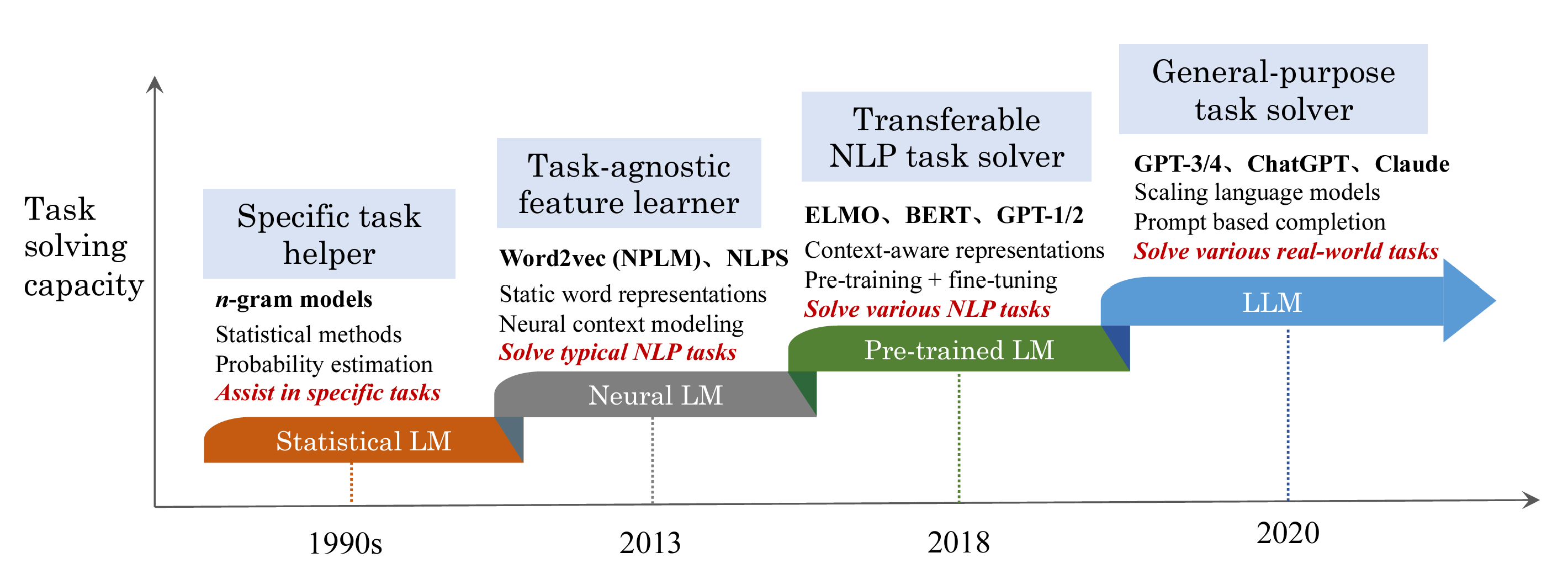}
    \caption{The evolution of language models \citep{zhao2023survey}.}
    \label{fig:lm-history}
\end{figure}

Around 1990, \textbf{statistical LMs}, or n-gram models, based on the Markov assumption, were proven useful in very specific tasks such as part-of-speech tagging \citep{thede1999second}. One typically use these statistical LMs for \textbf{probability estimation}: e.g., predicting the tag with the highest probability associated with a token. In these tasks, the probabilistic distribution of the in-domain vocabulary was simple enough to be modeled by the limited order of these statistical LMs. In much more complex tasks, the magnitude of the transition probability becomes intractable. 

Around 2013, said limitation was addressed by \textbf{neural LMs}. Deep learning approaches such as word2vec \citep{mikolov2013distributed} enabled the concept of a distributed representation, condensing information from the context. One typically use these neural LMs for \textbf{feature extraction}: e.g., concerting text into a vector representation which is then fed to a model, such as KNN or LSTM. Due to their task-agnostic nature, these neural LMs had gained dominance in many NLP tasks. However, supervised neural models are data-hungry, and thus it is challenging to annotate sufficient data for most tasks, especially the low-resource ones.

Around 2018, this pain point was alleviated by \textbf{pre-trained LMs}. Exemplified by ELMO \citep{peters-etal-2018-deep}, BERT \citep{devlin-etal-2019-bert}, and GPT-2 \citep{radford2019language}, these general-purpose models were trained with large-scale unlabeled corpora and could effectively transfer to individual tasks requiring a relatively small amount data. This is referred to as the \textbf{pre-train then fine-tune paradigm} which had since become mainstream. With the new-gained power of domain adaptation, pre-trained LMs redefined state-of-the-art for many NLP tasks. Notably, the might of these LMs to preserve knowledge during large-scale pre-training has long been attributed to the Transformer architecture \citep{vaswani2017attention}, which has persisted for most offspring NLP models. However, in computer vision, there has been competing arguments that scale plays a more important role than architecture \citep{smith2023convnets}.

Around 2020, pre-trained LMs (e.g., BERT with 330M parameters) were scaled up to become \textbf{large language models} (e.g., GPT-3 with 175B parameters, PaLM \citep{chowdhery2023palm} with 540B parameters). The leap in model size showed emergent abilities \citep{wei2022emergent}. The abilities allow for solving complex tasks that had not been possible before (e.g., many of the reasoning tasks will be discussed in this thesis). Also importantly, these LLMs that are pre-trained on a mixture of code and text can generate well-formed symbolic data (e.g., Python, Javascript, JSON), an ability that will be extensively discussed in Section~\ref{section:code_prompt} onward. Moreover, they enable the \textbf{in-context learning paradigm}, where previously required fine-tuning data is reduced to just a handful of few-shot learning exemplars, in some cases, zero-shot. In addition to the remarkable performance in many mainstream NLP tasks on par with the data-heavy fine-tuning paradigm, this new paradigm of interacting with large language models is a game-changer for many practitioners, for science and business alike, who have very limited budget for data annotation. In 2023, when I started writing this thesis, products like ChatGPT\footnote{\url{https://openai.com/blog/chatgpt}} have become a term as commonplace as `iPhone' for many ordinary people, and a go-to solution provider for many businesses.

It is plain to see that in the past decade alone, LMs have evolved from a probability estimator to a problem solver, drifting away from the original perception of ``language modeling.'' Indeed, most task formats in NLP can be reduced to generative language modeling. For example, solving a question answering (QA) example is equivalent to generating the answers given the question as the context. Conversely, generating text using LMs is also equivalent to answering the question ``what would be the appropriate response to ...'' Hence, the task \textit{formats} often discussed in the NLP community (e.g., QA, dialog, generation, information retrieval, etc.) is not nearly as important as the \textit{core} of the tasks (e.g., event-centric learning, logical reasoning, etc.). 

This thesis includes work that spans 2018 to 2024, and thus will primarily focus on the last two types of language models, collectively referred to as LLMs\footnote{After all, 330 million (BERT's number of parameters) is \textit{large} by most people's standards.}. Due to the highly empirical nature of work discussed in this thesis, it is sufficient to understand the usage of these LLMs (i.e., the two paradigms) without grasping the internal details of these models.

\subsection{LLMs are Black-Boxes}

A black-box is a mechanism that takes in an input, produces an output, while the user cannot interpret what goes on within. Regardless of performance, a black-box gives rise to multiple problems at once: 
\begin{enumerate}
    \item Trustworthiness: a user cannot trust the black-box by knowing that its underlying mechanism aligns with their expectation\footnote{Though a user might still trust a black-box that empirically performs satisfactorily (e.g., by statistical or anecdotal evidence).};
    \item Verifiability: a user cannot systematically examine if the output is correct;
    \item Improvability: a user cannot do things differently for a better performance.
    
\end{enumerate}

As discussed above, LLMs in the scope of this thesis are neural models, which are long known to be known as black-boxes \citep{benitez1997artificial}. Notwithstanding decade-long efforts on looking \textit{inside} the black-box (i.e., understanding what neural models do under the hood) \citep{dayhoff2001artificial}, I instead take a more macro view and focus on the interpretability of the higher-order pipeline that involves the black-box. In other words, I am interested in \textit{how to use} black-box LLMs so that a user gains more empirical insights into their decision-making process.

The most intuitive and least interpretable approach to use an LLM for most NLP tasks is \textbf{end-to-end}. The two ``ends'' here specify the input and output, agnostic of the task format. For a QA format, the input is the question while the output is the answer; for a dialog format, the input is the conversation history while the output is the utterance; for a translation format, the input is a source passage while the output is a target passage. The end-to-end usage is also agnostic to the two paradigms, pre-train then fine-tune and in-context learning. For the former, both the training and testing data are tuples of input and output; for the latter, both the in-context and to-infer examples are tuples of input and output. With the advancement of large-scale pretraining, the straightforward end-to-end usage demonstrates impressive performance and flexibility, as demonstrated in the big tables of results in mainstream LLM papers \citep{devlin-etal-2019-bert,liu2019roberta,radford2019language,raffel2020exploring,brown2020language,chowdhery2023palm,touvron2023llama}. It is arguably the default way to interact LLMs for most. Naturally, the end-to-end usage falls short on the three criteria above. On improvability, specifically, a user's hands are frustratingly tied. They may either increase the quantity or quality of the fine-tuning data (for the first paradigm), or engineer a different prompt (for the second paradigm), both of which come at an obvious cost. 

Later, I will discuss existing attempts to address these issues.

\newpage
\section{Machine Reasoning}
\label{sec:machine_reasoning}

\subsection{Reasoning Frameworks}
\label{sec:reasoning_framework}

Reasoning is an intriguing and crucial ability of human cognition \citep{rips1990reasoning}. Naturally, machine reasoning is a well sought-after ability in AI systems to demonstrate human-like intelligence. In the communities of mathematics, AI, machine learning, and NLP, the concept of reasoning is historical and varies greatly \citep{mccarthy1959programs, pearl1988probabilistic, duan-etal-2020-machine}. For a brief survey on the topic, readers are referred to \citet{duan-etal-2020-machine}. For the purpose of this thesis, it is crucial to understand two frameworks for machine reasoning: symbolic reasoning and textual reasoning. 

\textbf{Symbolic reasoning} works by manipulating knowledge in the form of symbolic logic using inference algorithms. Such inference algorithm can either be deterministic (Good, Old-Fashioned AI) or probabilistic \citep{pearl1988probabilistic,richardson2006markov}. Either way, the input must be formalized as symbols, which in itself is a non-trivial process for many applications. On the other hand, since the age of deep learning and LLMs, \textbf{textual reasoning} that only works with textual input has been made possible. By 2023, hundreds of papers with `reasoning' in their titles have been published\footnote{\url{https://aclanthology.org/}} but very few use any symbolic reasoning technique, but primarily use LLMs to answer \textit{reasoning questions} \citep{nlrse-2023-natural}. These include commonsense reasoning \citep{davis2015commonsense}, numerical reasoning \citep{cobbe2021gsm8k}, and of course event-centric reasoning that will be the focus on this thesis. 

As discussed before, the default methodology is end-to-end language modeling, which, in conjunction with various techniques, has achieved dominant performance in many of these reasoning tasks.

Comparing the two frameworks for machine reasoning, symbolic reasoning is highly interpretable and may even guarantee correctness given a well-defined symbolic input, much unlike the finicky black-box neural models. Conversely, symbolic reasoning is also highly rigid, requiring extensive training or annotation within a domain, a style, and even a singular dataset, much unlike modern LLMs which can work with rich expressions in diverse domains. Such an impasse of formal vs. neural methods is historical and ongoing. 

\textbf{Neural-symbolic reasoning} attempts to combine the best of both worlds and has a long history in literature \citep{hitzler2022neuro}. Indeed, there are many ways in which one can simultaneously work with symbols and neural networks, including knowledge graphs \citep{bordes2013translating,teru2020inductive}, neural semantic parsing \citep{dong-lapata-2016-language,finegan-dollak-etal-2018-improving}, etc. In this thesis, the notion of \textit{structured reasoning} is a subset of neural-symbolic reasoning, specifically designed to tackle event-centric reasoning. The core idea is to combine LLMs with structured representation of events. We will explore this approach in details later. 

\subsection{Reasoning Tasks}
In this thesis, I will focus on reasoning tasks in a broad and empirical sense. In NLP literature, `reasoning' is a highly overloaded term: what counts as reasoning as what does not is far from clear. What I have in mind is close to the sub-field of \textbf{multi-hop reasoning}, either QA from a passage \citep{welbl-etal-2018-constructing, talmor-berant-2018-web, yang-etal-2018-hotpotqa} or utilizing multihop information in the form of symbolic data \citep{de-cao-etal-2019-question, ding-etal-2019-cognitive, cao-etal-2019-bag, fang-etal-2020-hierarchical, thayaparan-etal-2019-identifying}. Here, a question has to be answered (or, a problem has to be solved) using at least two pieces of knowledge and one algorithmic operation. However, reasoning processes are subjective. To answer the question ``does Ella Fitzgerald typically sing with swing\footnote{A rhythmic feel that is iconic in jazz music.}'' probably requires just a single hop for jazz lovers, but at least two hops in addition to acquiring missing knowledge for the rest. In this thesis, I take an eclectic view and do not enforce what counts or does not count as reasoning, with much overlap with tasks like inference, entailment, commonsense, and QA. 

This thesis focuses on the family of tasks called \textbf{event reasoning} that targets either extracting the knowledge or answering the questions described above. In NLP, this work includes includes extracting knowledge from instructional texts \citep{10.1145/584955.584977,delpech-saint-dizier-2008-investigating, zhang-etal-2012-automatically}, reasoning about relations among events \citep{takechi-etal-2003-feature,tandon-etal-2019-wiqa,rajagopal-etal-2020-ask}, event knowledge-base construction \citep{jung2010automatic, chu2017distilling,park2018learning,zhou-etal-2022-show}, or various downstream applications \citep{zhang-etal-2020-analogous,dalvi-etal-2019-everything,chen-etal-2020-hybridqa}. 

\section{Interplay}
In summary, my work discussed in this thesis can be situated in the overlap of the three concepts above: LLMs, events, and reasoning. To tackle event reasoning tasks, I use a conjunction LLMs and structured event representation.  

\chapter{EVENT-RELATION SCHEMA: A NATURAL LANGUAGE REPRESENTATION}
\label{chap:relation}

Event reasoning is an umbrella term for many tasks and applications. A few event reasoning tasks that I will discuss in this Chapter are: predicting the next event, predicting the intent of an action, and predicting the steps to take in a procedure. In human communication, these tasks almost always involve input and output in the form of natural language. Traditionally, for machines to consume the data and learn the task, it has been imperative to represent events in a symbolic fashion (discussed in Section~\ref{sec:related_work_events_procedures}). For example, previous work has explored \textbf{event schemata} \citep{baker1998berkeley,li-etal-2020-cross} or \textbf{procedure schemata} \citep{momouchi-1980-control,zhang-etal-2012-automatically,kiddon-etal-2015-mise} that identify the key components in representing events and procedures. While these schemata are focused and effective for certain tasks, their construction is prone to errors and their ability to generalize to unseen domains is limited. On the other hand, by the time this research project was performed, LLMs have shown superior performance on many of the event reasoning tasks exemplified above. Seemingly, LLMs eliminate the need for any such structured representation, because they can take natural language as input and generate natural language texts directly. Nevertheless, I will show that end-to-end LLMs cannot take advantage of specific knowledge that could be encoded in the schemata, so their performance still has room for improvement. 

To push the limit of automated event reasoning, I attempt to bridge the two attempts by defining a language-based structured representation that not only retains a considerable amount of the structured knowledge of events, but also can be consumed and leveraged by LLMs (see Figure~\ref{fig:pipeline_nl}). Previously described symbolic schemata (for example, see Figure~\ref{fig:zhang2012}) are not a good fit with LLMs which are trained to exclusively work with natural language. To strike a balance between a symbolic vs. natural language representation, I propose a minimal \textbf{relational event schema} that includes two important event relations: hierarchical \textsc{goal-step relation} between an event and its sub-events, and a \textsc{step-step temporal relation} among the sub-events. The idea is explored under the DARPA KAIROS project\footnote{\url{https://www.darpa.mil/program/knowledge-directed-artificial-intelligence-reasoning-over-schemas}} and its related work \citep{li-etal-2021-future} (see Figure~\ref{fig:wikihow_goal_steps} for an illustration). Compared to existing event schemata such as Figure~\ref{fig:zhang2012}, the proposed one has two advantages. First, it is lightweight but versatile, which I will demonstrate through improvements on three downstream tasks later in this Chapter. Second, the operands of each relation (i.e., the atomic unit) are events simply expressed as natural language phrases, which are much more higher-level and flexible than symbolic atomic units. As these natural language event expressions can be consumed and interpreted by LLMs, generalization among domains becomes much more likely. 

\begin{figure}[t!]
    \centering
    \includegraphics[scale=0.6]{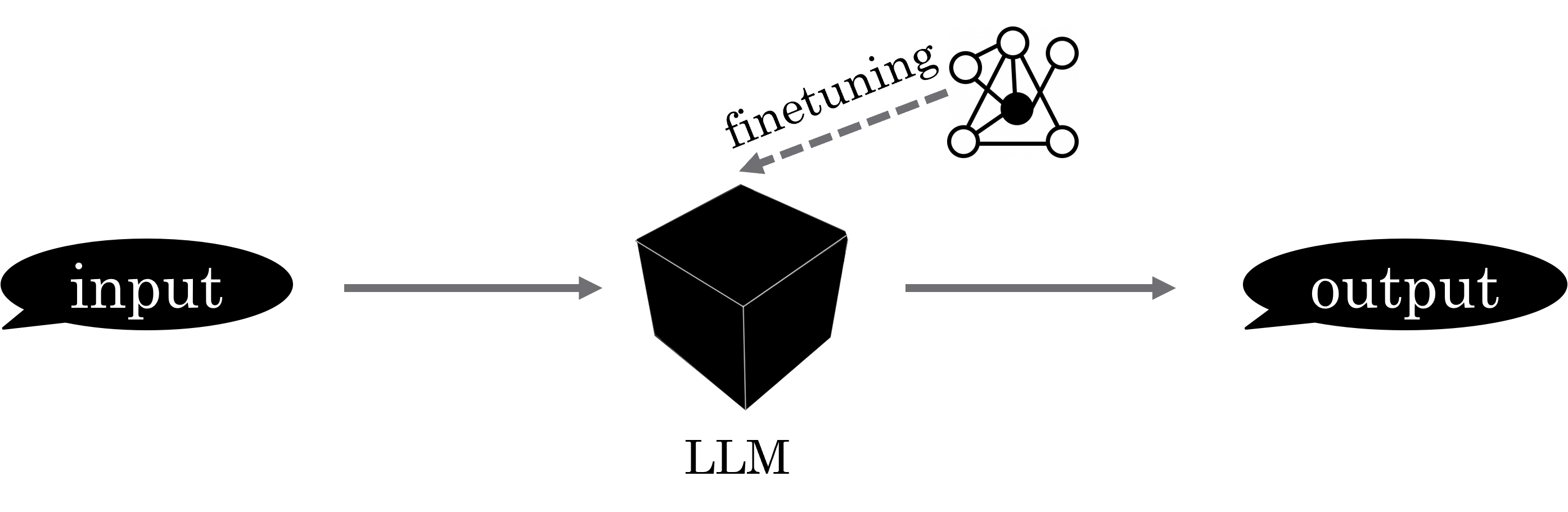}
    \caption{An illustration of my proposed pipeline leveraging a natural language representation of entities. The LLM is fine-tuned with data of that representation.}
    \label{fig:pipeline_nl}
\end{figure}

\begin{figure}[t!]
    \centering
    \includegraphics[width=\columnwidth]{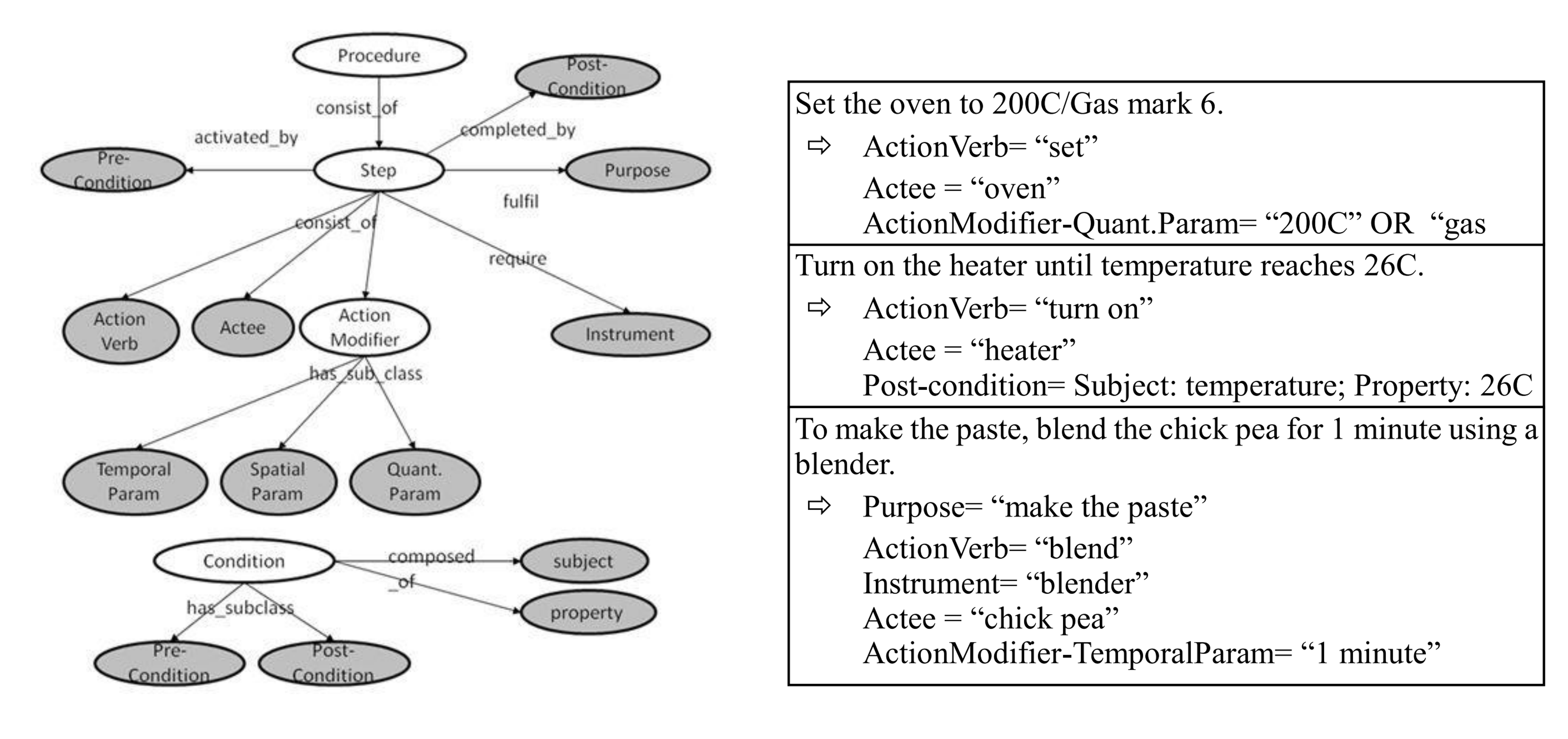}
    \caption{A fine-grained procedure representation (schema) proposed by \citet{zhang-etal-2012-automatically}.}
    \label{fig:zhang2012}
\end{figure}

Next, I describe work done by my collaborators and myself to first learn such a relational event schema data from the web (Section \ref{sec:learn_relations}). We use this data to equip LLMs with such structured knowledge via fine-tuning and apply them to downstream tasks (Section \ref{sec:relation_downstream} and Section \ref{sec:script_generation}).

\section{Learning Event Relations}
\label{sec:learn_relations}

\begin{figure}
    \centering
    \includegraphics[scale=0.7]{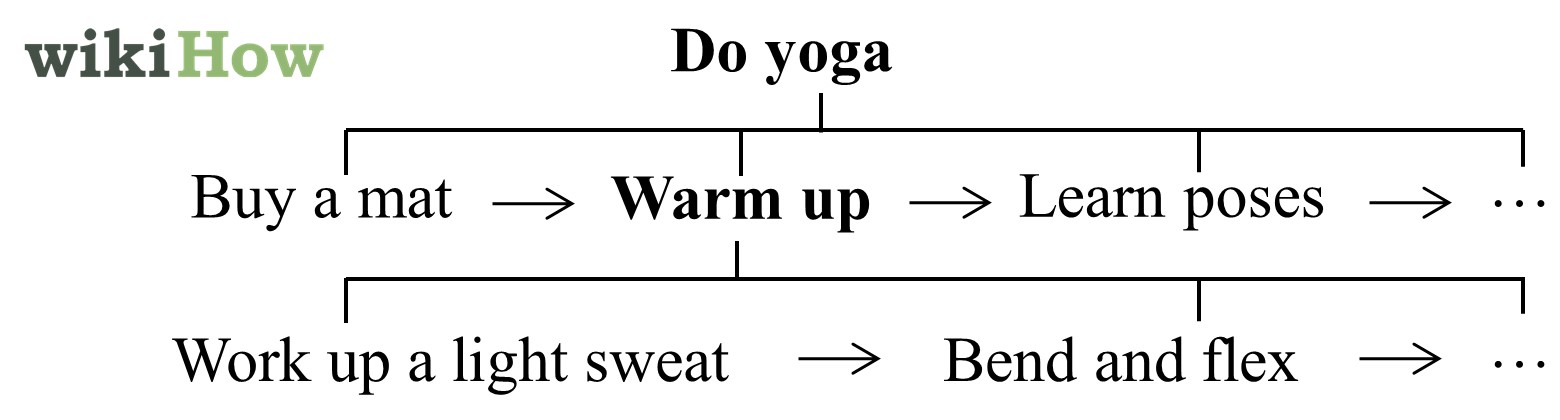}
    \caption{Goals and steps (slightly paraphrased) from wikiHow articles ``How to Do Yoga'' and ``How to Warm Up''. The lines denote goal-step relations; the arrows denote step-step temporal relations.}
    \label{fig:wikihow_goal_steps}
\end{figure}

To construct a large-scale dataset of procedures, we crawl wikiHow\footnote{\url{wikihow.com}} which is a how-to website containing more than 110k procedures at the time the work was done in 2020. It has been leveraged by researchers since its inception \citep{10.1145/988672.988750,Addis2011FromUW,10.1145/2567948.2578846}. Procedural texts are suitable data for inducing hierarchical and temporal relations, because the hierarchical relation exist organically between a \textbf{goal} and its \textbf{steps} (i.e., \textsc{goal-step relation}), and the \textsc{temporal relation} can be found among the steps. Each wikiHow article represents a procedure and contains a title, which we extract as a goal, and some step paragraphs, of which we extract the headlines as steps (see the top part of Figure~\ref{fig:wikihow_goal_steps} for an example). Additionally, not used in this work, our data includes related articles, references, Q\&A, tips and warnings, links to images, and videos aligned with texts. 

wikiHow has articles from a broad range of domains, with 19 top-level categories: Arts and Entertainment, Cars \& Other Vehicles, Computers and Electronics, Education and Communications, Family Life, Finance and Business, Food and Entertaining, Health, Hobbies and Crafts, Holidays and Traditions, Home and Garden, Personal Care and Style, Pets and Animals, Philosophy and Religion, Relationships, Sports and Fitness, Travel, Work World, and Youth. We plot the distribution of the top eight categories in Figure~\ref{fig:cat_dist_app}.

\begin{figure}
    \centering
    \includegraphics[scale=0.3]{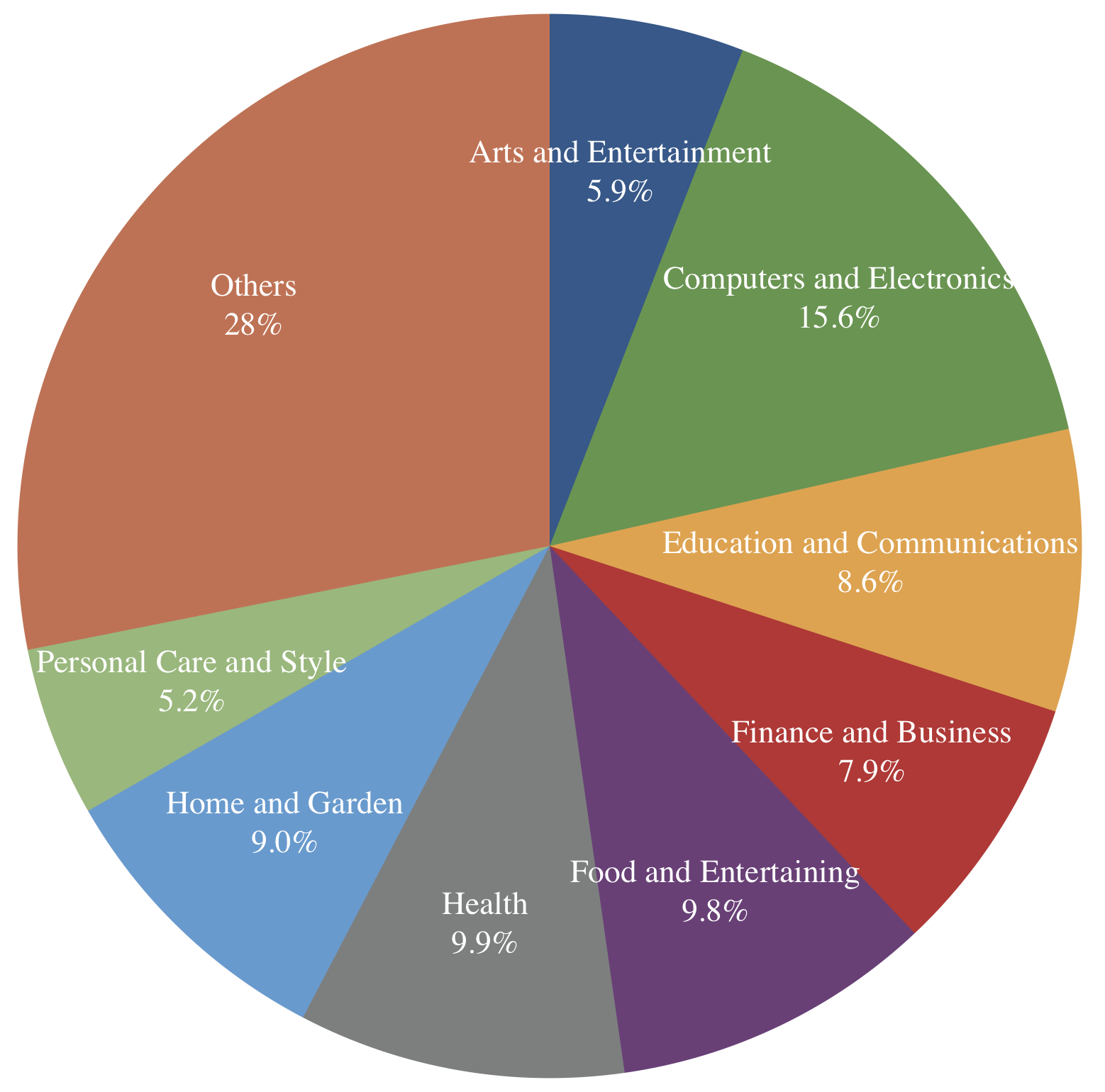}
    \caption{Category distribution of wikiHow articles.}
    \label{fig:cat_dist_app}
\end{figure}

Next, our goal is to learn an relational event schema including two relations: hierarchical relation between an event and its sub-events, and temporal relation among the sub-events. In procedures, these are equivalent to the \textsc{goal-step} relation, and the step-step temporal relation. Using the wikiHow corpus, we construct a dataset of three event relation inference tasks. 

\subsection{Step Inference Task}
\label{sec:step-inference-task}
We first introduce the \textit{Step Inference} task, targeting \textsc{goal-step} relations between events. We formulate this as a 4-choose-1 multiple choice format evaluated using accuracy.

In this task, a system is given a goal and 4 candidate steps and needs to choose the step that helps achieve the goal. For example, given the goal ``Prevent Coronavirus'' and the candidate steps:
\begin{center}
A. wash your hands \hspace{20pt} B. wash your cat \\
C. clap your hands \hspace{24pt} D. eat your protein\\
\end{center}
the correct step would be A. 

Obtaining the goal and the positive candidate is straightforward, as we sample them iteratively from each how-to article. However, it is challenging to sample negative candidates \citep{chao-etal-2018-negative,zellers2019vcr} which should have high semantic relatedness with the positive candidate (or the question becomes trivial) while being incorrect answers. We first map each step in wikiHow to a vector representation by taking the average of the BERT embeddings \citep{devlin-etal-2019-bert} of the verbs. Given the positive step, we then choose 3 steps under different wikiHow categories with the highest cosine similarity to it as the negative candidates. The nearest-neighbors are computed using FAISS \citep{JDH17}.

It has recently become clear that the latest NLP models can exploit statistical artifacts from a dataset \citep{poliak-etal-2018-hypothesis,Si2019WhatDB,zellers-etal-2019-hellaswag}. To prevent the model from learning the negative sampling strategy and relying on just the candidates, we randomly reassign one of the candidates as positive, and the others as negative. Then, we replace the provided goal with the goal attached to the new positive candidate. This strategy ensures that any model performs no better than chance when given access to only the candidates and not the context. 

For each step in wikiHow, we create an example by using it as the positive candidate, followed by the negative sampling and label reassignment processes as described above. Then, we apply a collection of hand-crafted filters to remove low-quality examples.

\subsection{Goal Inference Task}
\label{sec:goal-inference-task}
Next, we introduce the \textit{Goal Inference} task, formulated in a similar way as Step Inference. 

In this task, a system is given a prompt step and 4 candidate goals and needs to choose the correct goal which the step helps achieve. For example, given the step ``choose a color of lipstick'' and the candidate goals:
\begin{center}
A. Get Pink Lips \hspace{18pt} B. Read One's Lips \\
C. Lip Sync \hspace{39pt} D. Draw Lips
\end{center}
the correct goal would be A.

For each goal in wikiHow, we create the set of 4 candidates by using it as the positive candidate, followed by the negative sampling, label reassignment, and filtering processes as in Step Inference. For each positive candidate goal, we use each of its steps to create an example. 

\subsection{Step Ordering Task}
Finally, we introduce the \textit{Step Ordering} task, targeting \textsc{step-step temporal} relations between events. This task is in a 2-choose-1 multiple choice format evaluated using accuracy.

In this task, given a prompt goal and 2 steps, a system needs to determine which step temporally precedes the other. For example, given the goal ``Clean Silver'' and the steps:
\begin{center}
A. dry the silver \hspace{10pt} B. handwash the silver
\end{center}
the correct answer would be B precedes A.

Unfortunately, not all steps in every wikiHow article follow an underlying order. We observe that there are 2 types of wikiHow articles. One is \textit{unordered}, where the steps are parallel alternatives, such as ways to ``Stay Healthy'' (``develop an exercise routine", ``get enough sleep", ``eat a healthy diet'', etc.). The other is \textit{ordered}, such as recipes for cooking or manuals for fixing appliances, where most steps should be taken sequentially. 

We ask 3 annotators to label 1,000 wikiHow articles as ordered or not as a coarse-grained approximation for whether their steps are ordered. We finetune a pre-trained RoBERTa model using 5-fold cross-validation, finding an average precision of 88\%. We then ask a 4\textsuperscript{th} annotator to label another 40 articles as the held-out test set, where the finetuned model achieves 100\% precision. Finally, we only consider articles that the model predicts as ordered (around 40\%) for the Step Ordering task. 

For each goal in wikiHow, we create a set of examples by using it as the prompt and sampling every pair of its adjacent steps as candidates. Then, we randomly shuffle the candidates, so each appears first with 50\% chance. 

\subsection{Test Set Construction and Validation}\label{crowdsourcing-validation}
There exists some noise in our automatically generated examples, because some of them do not have a single correct answer. Errors can be introduced when a sampled negative candidate is in fact correct. For example, in the Goal Inference task, consider an example where the give step is ``practice swings'', the expected positive candidate step is ``Play Golf'', and a candidate negative example is ``Play Drums''. ``Play Drums'' is sampled due to its high embedding similarity with ``Play Golf'' and is also a reasonable goal for ``practice swings (of the drumsticks)''. This is an ambiguous example and should be excluded from the test set. Therefore, we ask crowd workers to validate a subset of the examples. 

\begin{figure*}
    \centering
    \includegraphics[width=\linewidth]{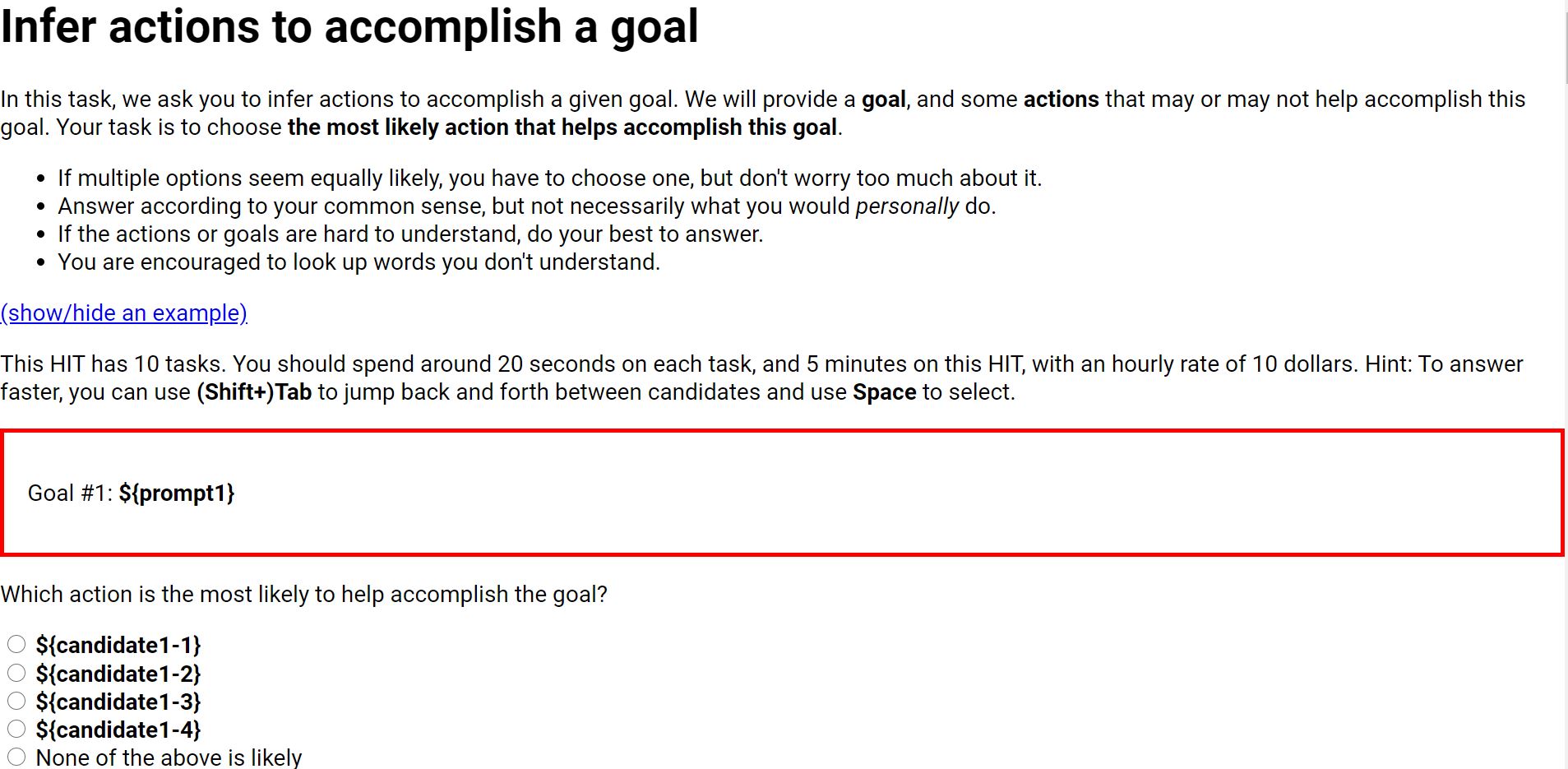}
    \caption{Screenshot of the HIT design for the Step Inference task.}
    \label{fig:step-hit}
\end{figure*}

We perform crowdsourcing on Amazon Mechanical Turk, requiring Master Qualification and a lifetime HIT approval rate over 90\%. The HIT design is shown in Figure~\ref{fig:step-hit}.

For each of Step Inference and Goal Inference, we randomly sample 4,800 examples as input, and for each example we ask 3 crowd workers to choose the most likely candidate. Every HIT includes 15 examples with a pay of \$0.83, estimated to be completed in 5 minutes, equivalent to an hourly rate of \$9.96.

For Step Ordering, we randomly sample 9,300 examples, and for each example we ask 3 crowd workers to order the events (with a ``neutral'' option). Every HIT includes 30 examples with a pay of \$0.83, estimated to be completed in 5 minutes, equivalent to an hourly rate of \$9.96. 

In the test set, we only keep examples where all 3 workers agree with the gold label as our benchmark. We remove all examples from the automatically generated ones whose prompt or candidates appear in the test set, and use the remaining data as the training set.

\subsection{In-Domain Evaluation}
\label{sec:goal_step_in_domain}
To see if LLMs can effectively learn to predict thse two event relations, we finetune pretrained BERT \citep{devlin-etal-2019-bert}, XLNet \citep{yang2019xlnet}, RoBERTa \citep{liu2019roberta} and GPT-2 \citep{radford2019language} models on the training set and report accuracy on the test set. To benchmark human performance, two authors each annotate 100 random examples from the test set and report the average accuracy. The results are shown in Table~\ref{tab:goal-step-performance}, indicating a satisfactory performance and still a gap of 10\% to 20\% between human and models trained on all available in-domain data. 

At the time this project was carried out, the above models represented the state-of-the-art. Later LLMs with in-context learning abilities would demonstrate an even stronger performance \citep{srivastava2022beyond}. However, as these larger models are trained on internet data with a later cut-off date than this project, one cannot rule out the possibility that the strong performance is attributed to data contamination (i.e., the models have already seen the test set and labels of our dataset on the web).

\begin{table}
\centering
\begin{tabular}[t]{lllll}
\toprule
 & \makecell{Step\\Infer.} & \makecell{Goal\\Infer.} & \makecell{Step\\Ordering} \\ \midrule
Train size & 374,278 & 185,231 & 836,128 \\
Test size & 2,250 & 1,703 & 3,100 \\ \midrule
BERT & .874 & .798 & .819 \\
XLNet & .867 & .783 & .826 \\ 
RoBERTa & .882 & .820 & .835 \\
GPT-2 & .836 & .686 & .801 \\ 
\midrule
Human & .965 & .980 & .975 \\
\bottomrule
\end{tabular}
\caption{The accuracy of state-of-the-art models on the test sets after being finetuned on our training sets.}
\label{tab:goal-step-performance}
\end{table}

\subsection{Open-Ended Examples}
\label{sec:goal_step_open_ended}
In addition to quantitatively evaluating models on our multiple-choice tasks, we perform qualitative evaluation on some open-ended examples from wikiHow unseen during training, using RoBERTa.

For Step Inference, we rank 100 steps with high embedding similarity for their likelihood of helping achieve a given goal. For example, for the goal ``Eat in Islam'', the top 3 ranked steps are ``understand what type of meats are permissible'' (correct), ``start by adding mild spices to your food,'' and ``gather supplies and ingredients.'' Similarly for Goal Inference, we rank 100 goals against some steps. For example, for the steps ``spend the holiday with your beloved, eat KFC, check out the light displays,'' the top 3 ranked goals are ``Celebrate a Japanese Christmas'' (correct)\footnote{KFC and light displays are Japanese Christmas traditions \citep{kimura2005christmas}.}, ``Celebrate a Czech Christmas,'' and ``Celebrate a British Christmas.'' These examples show that the model trained on our data can retrieve texts based on \textsc{goal-step} relations, beyond simply semantic relevance. 

For Step Ordering, the model can perfectly order some articles with as much as 10 steps. For example, given the goal ``Clean a Trumpet,'' the first 5 predicted, ordered steps are ``gather your materials,'' ``disassemble your trumpet,'' ``fill up a bathtub,'' ``place a towel down in the tub,'' and ``set your trumpet parts down to soak.'' This shows that the model trained on our data can order certain long sequences of events based on \textsc{step-step temporal} relations. 

At this point, we have a suite of LLMs that can discriminate the \textsc{goal-step} relation and \textsc{step-step temporal} relation, given two events. To achieve this, we construct a dataset of procedural knowledge from the web, and fine-tune LLMs on examples that illustrate the above two relations, which collectively constitute the relation event schema I proposed at the beginning of this Chapter. Next, I will show evidence that these LLMs can effectively perform on various event-centric reasoning tasks.

The work above was published in \citet{zhang-etal-2020-reasoning}, in which my collaborator Qing Lyu and I contributed equally in roughly all components. I have obtained approval from all collaborators to exclusively include this work in this thesis. 

\newpage
\section{Application of Event Relation}
\label{sec:relation_downstream}

Using the new-gained ability of LLMs to predict the edges in a graph of the relational event schema described in the beginning of Section~\ref{sec:learn_relations}, we will now apply these two relations to two major NLP tasks, intent detection (a component of dialog systems) and next event prediction (a component of commonsense inference). 
For both cases, if models equipped with the knowledge of relational event schema outperforms those which are end-to-end, we would have shown that our structured representation of events is practically beneficial.

\subsection{Intent Detection}
\label{sec:intent_detection}

Task-oriented dialog systems like Apple's Siri, Amazon Alexa, and Google Assistant have become pervasive in smartphones and smart speakers. To support a wide range of functions, dialog systems must be able to map a user's natural language instruction onto the desired skill or API. Performing this mapping is called intent detection.

Intent detection is usually formulated as a sentence classification task. Given an utterance (e.g. ``wake me up at 8''), a system needs to predict its intent (e.g. ``Set an Alarm''). Most modern approaches use neural networks to jointly model intent detection and slot filling \citep{6707709,liu2016attentionbased,goo-etal-2018-slot,zhang-etal-2019-joint}. In response to a rapidly growing range of services, more attention has been given to zero-shot intent detection \citep{ferreira2015zero,ferreira15,yazdani-henderson-2015-model,chen2016zero,kumar2017zero}. While most existing research on intent detection proposed novel model architectures, few have attempted data augmentation. One such work \citep{10.1145/1526709.1526773} showed that models can learn much knowledge that is important for intent detection from massive online resources such as Wikipedia.

The core idea of our proposed methodology is to transfer the knowledge of \textsc{goal-step} event relation to the intent detection task, based on the observation that a goal can approximate an intent, and each step in it can approximate an associated utterance. Hence, we reuse the finetuned models described in Section~\ref{sec:goal_step_in_domain} with some minor differences. To enable multilingual settings, we fine-tune a pretrained RoBERTa model for the English datasets and a pretrained XLM-RoBERTa model \citep{conneau-etal-2020-unsupervised} for the multilingual datasets. In test time, we cast the instances of the intent detection datasets into a multiple-choice format, where the utterance is the input and the full set of intents are the possible candidates, consistent with our wikiHow pretraining task. For each model, we append a linear classification layer with cross-entropy loss to calculate a likelihood for each candidate, and output the candidate with the maximum likelihood. 

For each intent detection dataset in any language, we consider the following settings:\\
\textbf{+in-domain} (+ID): a model is only trained on the dataset's in-domain training data;\\
\textbf{+wikiHow +in-domain} (+WH+ID): a model is first trained on our wikiHow data in the corresponding language, and then trained on the dataset's in-domain training data;\\
\textbf{+wikiHow zero-shot} (+WH 0-shot): a  model is trained only on our wikiHow data in the corresponding language, and then applied directly to the dataset's evaluation data.

For non-English languages, the corresponding wikiHow data might suffer from smaller sizes and lower quality. Hence, we additionally consider the following cross-lingual transfer settings for non-English datasets:\\
\textbf{+en wikiHow +in-domain} (+enWH+ID), a model is trained on wikiHow data in English, before it is trained on the dataset's in-domain training data;\\
\textbf{+en wikiHow zero-shot} (+enWH 0-shot), a model is trained on wikiHow data in English, before it is directly applied to the dataset's evaluation data.

We consider the 3 following benchmarks: \\
\textbf{The Snips dataset} \citep{coucke2018snips} is a single-turn English dataset. It is one of the most cited dialog benchmarks in recent years, containing utterances collected from the Snips personal voice assistant. While its full training data has 13,784 examples, we find that our models only need its smaller training split consisting of 2,100 examples to achieve high performance. Since Snips does not provide test sets, we use the validation set for testing and the full training set for validation. Snips involves 7 intents, including \textit{Add to Playlist}, \textit{Rate Book}, \textit{Book Restaurant}, \textit{Get Weather}, \textit{Play Music}, \textit{Search Creative Work}, and \textit{Search Screening Event}. Some example utterances include ``Play the newest melody on Last Fm by Eddie Vinson,'' ``Find the movie schedule in the area,'' etc.\\
\textbf{The Schema-Guided Dialogue dataset} (SGD) \citep{rastogi2020towards} is a multi-turn English dataset. It is the largest dialog corpus to date spanning dozens of domains and services, used in the DSTC8 challenge \citep{rastogi2020schema} with dozens of team submissions. Schemas are provided with at most 4 intents per dialog turn. Examples of these intents include \textit{Buy Movie Tickets for a Particular show}, \textit{Make a Reservation with the Therapist}, \textit{Book an Appointment at a Hair Stylist}, \textit{Browse attractions in a given city}, etc. At each turn, we use the last 3 utterances as input. An example: ``That sounds fun. What other attractions do you recommend? There is a famous place of worship called Akshardham.''\\
\textbf{The Facebook multilingual datasets} (FB-en/es/th) \citep{schuster-etal-2019-cross-lingual} is a single-turn multilingual dataset. It is the only multilingual dialog dataset to the best of our knowledge, containing utterances annotated with intents and slots in English (en), Spanish (es), and Thai (th). It involves 12 intents, including \textit{Set Reminder}, \textit{Check Sunrise}, \textit{Show Alarms}, \textit{Check Sunset}, \textit{Cancel Reminder}, \textit{Show Reminders}, \textit{Check Time Left on Alarm}, \textit{Modify Alarm}, \textit{Cancel Alarm}, \textit{Find Weather}, \textit{Set Alarm}, and \textit{Snooze Alarm}. Some example utterances are ``Is my alarm set for 10 am today?'' ``Colocar una alarma para mañana a las 3 am,'' \includegraphics[height=14pt]{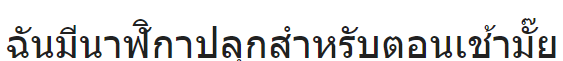}  etc.

\begin{table}
\centering
\begin{tabular}{lrrrr}
\toprule
 & \makecell{Training\\Size} & \makecell{Valid.\\Size} & \makecell{Test\\Size} & \makecell{Num.\\Intents} \\ \midrule
Snips & 2,100 & 700 & N/A & 7  \\
SGD & 163,197 & 24,320 & 42,922 & 4  \\ 
FB-en & 30,521 & 4,181 & 8,621 & 12  \\
FB-es & 3,617 & 1,983 & 3,043 & 12  \\ 
FB-th & 2,156 & 1,235 & 1,692 & 12  \\ 
\bottomrule
\end{tabular}
\caption{Statistics of the dialog benchmark datasets.}
\label{tab:intent_datasets}
\end{table}

Statistics of the datasets are shown in Table~\ref{tab:intent_datasets}.

We compare our models with the previous state-of-the-art results of each dataset: \\
\textbullet\ \citet{9082602}
proposed a Siamese neural
network with triplet loss, achieving state-of-the-art results on Snips and FB-en; \\
\textbullet\ \citet{8907842} used multi-task learning to jointly learn intent detection and slot filling, achieving state-of-the-art results on FB-es and FB-th;  \\
\textbullet\ \citet{ma2019endtoend} augmented the data via back-translation to and from Chinese, achieving state-of-the-art results on SGD.

\begin{table}
\centering
\begin{tabular}{llll}
\toprule
          & Snips & SGD & FB-en  \\ \midrule
\citep{9082602}      & .993  & N/A & .993 \\
\citep{ma2019endtoend}      & N/A  & .948 & N/A \\ \midrule
+in-domain (+ID)  & .990  & .942 & .993 \\
(ours) +WH+ID   & \textbf{.994}  & \textbf{.951$\dagger$} & \textbf{.995$\dagger$} \\
(ours) +WH 0-shot & .713  & .787 & .445 \\ \midrule
Chance    & .143  & .250 & .083 \\
\bottomrule
\end{tabular}
\caption{The accuracy of intent detection on English datasets using RoBERTa. State-of-the-art performances are in bold; $\dagger$ indicates statistically significant improvement from the previous state-of-the-art.}
\label{tab:intent_eng_results}
\end{table}

\begin{table}
\centering
\begin{tabular}{lllll}
\toprule
          & FB-en & FB-es & FB-th  \\ \midrule
\citep{9082602} & .993  & N/A & N/A \\
\citep{8907842} &  N/A  & .978 & .967 \\ \midrule
+in-domain (+ID) & .993  & .986 & .962 \\
(ours) +WH+ID   & \textbf{.995}  & .988 & .971 \\
(ours) +enWH+ID   & \textbf{.995}  & \textbf{.990$\dagger$} & \textbf{.976$\dagger$} \\
(ours) +WH 0-shot & .416  & .129 & .119 \\
(ours) +enWH 0-shot & .416  & .288 & .124 \\ \midrule
Chance    & .083  & .083 & .083 \\
\bottomrule
\end{tabular}
\caption{The accuracy of intent detection on multilingual datasets using XLM-RoBERTa. }
\label{tab:intent_multi_results}
\end{table}

\begin{figure*}\small
\centering
\begin{tikzpicture}[scale=0.8]
\begin{axis}[
    title={Snips (RoBERTa)},
    axis x line*=bottom,
    axis y line*=left,
    legend style={nodes={scale=0.7, transform shape}},
    xmode=log,
    log ticks with fixed point,
    ymin=0, ymax=1.05,
    ytick={0,.2,.4,.6,.8,1},
    x filter/.code=\pgfmathparse{#1 + 6.90775527898214},
    legend pos=north west,
    xticklabels={,,}
]
\addplot[mark=triangle,densely dotted,mark options={solid}] 
table {
0.01	0.117
0.05	0.137
0.1	0.47
0.5	0.84
1	0.963
}; 
\addplot[mark=square*,mark options={scale=0.8}] 
table {
0.01	0.823
0.05	0.9
0.1	0.953
0.5	0.974
1	.994
}; 
\draw ({rel axis cs:0,0}|-{axis cs:0,0.143}) -- ({rel axis cs:1,0}|-{axis cs:0,0.143});
\addplot[
    only marks,
    visualization depends on=\thisrow{alignment} \as \alignment,
    nodes near coords,point meta=explicit symbolic,every node near coord/.style={anchor=\alignment}
    ] table [
     meta index=2
     ] {
x       y       label       alignment
0.1 0.953 .953 -90
0.1 0.47 .470 150
};
\end{axis}
\end{tikzpicture}%
\hskip 1pt 
\begin{tikzpicture}[scale=0.8]
\begin{axis}[
    title={SGD (RoBERTa)},
    axis x line*=bottom,
    axis y line*=left,
    legend style={nodes={scale=0.7, transform shape}},
    xmode=log,
    ymin=0, ymax=1.05,
    ytick={0,.2,.4,.6,.8,1},
    log ticks with fixed point,
    x filter/.code=\pgfmathparse{#1 + 6.90775527898214},
    legend pos=north west,
    xticklabels={,,},
    yticklabels={,,}
]
\addplot[mark=triangle,densely dotted,mark options={solid}] 
table {
0.01	0.441
0.05	0.493
0.1	0.531
0.5	0.586
1	0.817
}; 
\addplot[mark=square*,mark options={scale=0.8}] 
table {
0.01	0.629
0.05	0.694
0.1	0.755
0.5	0.829
1	.860
}; 
\draw ({rel axis cs:0,0}|-{axis cs:0,0.250}) -- ({rel axis cs:1,0}|-{axis cs:0,0.250});
\addplot[
    only marks,
    visualization depends on=\thisrow{alignment} \as \alignment,
    nodes near coords,point meta=explicit symbolic,every node near coord/.style={anchor=\alignment}
    ] table [
     meta index=2
     ] {
x       y       label       alignment
0.1 0.531 .531 90
0.1 0.755 .755 -90
};
\end{axis}
\end{tikzpicture}%
\hskip 1pt 
\begin{tikzpicture}[scale=0.8]
\begin{axis}[
    title={FB-en (RoBERTa)},
    axis x line*=bottom,
    axis y line*=left,
    legend style={nodes={scale=0.7, transform shape}},
    xmode=log,
    ymin=0, ymax=1.05,
    ytick={0,.2,.4,.6,.8,1},
    log ticks with fixed point,
    x filter/.code=\pgfmathparse{#1 + 6.90775527898214},
    legend pos=north west,
    xticklabels={,,},
    yticklabels={,,}
]
\addplot[mark=triangle,densely dotted,mark options={solid}] 
table {
0.01	0.383
0.05	0.458
0.1	0.458
0.5	0.911
1	0.981
}; 
\addplot[mark=square*,mark options={scale=0.8}] 
table {
0.01	0.359
0.05	0.8
0.1	0.884
0.5	0.962
1	0.982
}; 
\draw ({rel axis cs:0,0}|-{axis cs:0,0.083}) -- ({rel axis cs:1,0}|-{axis cs:0,0.083});
\addplot[
    only marks,
    visualization depends on=\thisrow{alignment} \as \alignment,
    nodes near coords,point meta=explicit symbolic,every node near coord/.style={anchor=\alignment}
    ] table [
     meta index=2
     ] {
x       y       label       alignment
0.1 0.458 .458 150
0.1 0.884 .884 -40
};
\end{axis}
\end{tikzpicture}%
\hskip 1pt 
\begin{tikzpicture}[scale=0.8]
\begin{axis}[
    title={FB-en (XLM-RoBERTa)},
    axis x line*=bottom,
    axis y line*=left,
    legend style={nodes={scale=0.7, transform shape}},
    xmode=log,
    ymin=0, ymax=1.05,
    ytick={0,.2,.4,.6,.8,1},
    log ticks with fixed point,
    x filter/.code=\pgfmathparse{#1 + 6.90775527898214},
    legend pos=north west,
]
\addplot[mark=triangle,densely dotted,mark options={solid}] 
table {
0.01	0.405
0.05	0.459
0.1	0.481
0.5	0.75
1	0.951
}; 
\addplot[mark=square*,mark options={scale=0.8}] 
table {
0.01	0.667
0.05	0.789
0.1	0.894
0.5	0.964
1	0.98
}; 
\draw ({rel axis cs:0,0}|-{axis cs:0,0.083}) -- ({rel axis cs:1,0}|-{axis cs:0,0.083});
\addplot[
    only marks,
    visualization depends on=\thisrow{alignment} \as \alignment,
    nodes near coords,point meta=explicit symbolic,every node near coord/.style={anchor=\alignment}
    ] table [
     meta index=2
     ] {
x       y       label       alignment
0.1 0.894 .894 -40
0.1 0.481 .481 150
};
\end{axis}
\end{tikzpicture}%
\hskip 1pt 
\begin{tikzpicture}[scale=0.8]
\begin{axis}[
    title={FB-es (XLM-RoBERTa)},
    axis x line*=bottom,
    axis y line*=left,
    legend style={nodes={scale=0.7, transform shape}},
    xmode=log,
    ymin=0, ymax=1.05,
    ytick={0,.2,.4,.6,.8,1},
    log ticks with fixed point,
    x filter/.code=\pgfmathparse{#1 + 6.90775527898214},
    legend pos=north west,
    yticklabels={,,},
]
\addplot[mark=triangle,densely dotted,mark options={solid}] 
table {
0.01	0.29
0.05	0.138
0.1	0.349
0.5	0.646
1	0.902
}; 
\addplot[mark=square*,mark options={scale=0.8}] 
table {
0.01	0.143
0.05	0.244
0.1	0.633
0.5	0.804
1	0.925
}; 
\addplot[mark=o,densely dashed,mark options={solid}] 
table {
0.01	0.405
0.05	0.638
0.1	0.845
0.5	0.945
1	0.978
}; 
\draw ({rel axis cs:0,0}|-{axis cs:0,0.083}) -- ({rel axis cs:1,0}|-{axis cs:0,0.083});
\addplot[
    only marks,
    visualization depends on=\thisrow{alignment} \as \alignment,
    nodes near coords,point meta=explicit symbolic,every node near coord/.style={anchor=\alignment}
    ] table [
     meta index=2
     ] {
x       y       label       alignment
0.1 0.845 .845 -50
0.1 0.663 .663 -100
0.1 0.349 .349 150
};
\end{axis}
\end{tikzpicture}%
\hskip 1pt 
\begin{tikzpicture}[scale=0.8]
\begin{axis}[
    title={FB-th (XLM-RoBERTa)},
    axis x line*=bottom,
    axis y line*=left,
    legend style={nodes={scale=0.7, transform shape}},
    xmode=log,
    ymin=0, ymax=1.05,
    ytick={0,.2,.4,.6,.8,1},
    log ticks with fixed point,
    x filter/.code=\pgfmathparse{#1 + 6.90775527898214},
    legend style={at={(0.52,0.02)},anchor=south west},
    yticklabels={,,},
]
\addplot[mark=triangle,densely dotted,mark options={solid}] 
table {
0.01 .275
0.05 .161
0.1  .341
0.5  .517 
1    .881
}; \addlegendentry{+ID}
\addplot[mark=square*,mark options={scale=0.8}] 
table {
0.01	0.216
0.05	0.602
0.1	0.851
0.5	0.912
1	0.966
}; \addlegendentry{(ours)+WH+ID}
\addplot[mark=o,densely dashed,mark options={solid}] 
table {
0.01	0.151
0.05	0.665
0.1	0.853
0.5	0.919
1	0.97
}; \addlegendentry{(ours)+enWH+ID}
\draw ({rel axis cs:0,0}|-{axis cs:0,0.083}) -- ({rel axis cs:1,0}|-{axis cs:0,0.083});
\addlegendimage{}
\addlegendentry{Chance}
\addplot[
    only marks,
    visualization depends on=\thisrow{alignment} \as \alignment,
    nodes near coords,point meta=explicit symbolic,every node near coord/.style={anchor=\alignment}
    ] table [
     meta index=2
     ] {
x       y       label       alignment
0.1 0.853 .853 -40
0.1 0.851 .851 150
0.1 0.341 .341 -40
};
\end{axis}
\end{tikzpicture}
\caption{Learning curves of models in low-resource settings. The vertical axis is the accuracy of intent detection, while the horizontal axis is the number of in-domain training examples of each task, distorted to log-scale.}
\label{fig:intent_learning_curve}
\end{figure*}

The performance of RoBERTa on the English datasets (Snips, SGD, and FB-en) are shown in Table~\ref{tab:intent_eng_results}. We repeat each experiment 20 times, report the mean accuracy, and calculate its p-value against the previous state-of-the-art result, using a one-sample and one-tailed t-test with a significance level of 0.05. Our models achieve state-of-the-art results using the available in-domain training data. Moreover, our wikiHow data enables our models to demonstrate strong performances in zero-shot settings with no in-domain training data, implying our models' strong potential to adapt to new domains. 

The performance of XLM-RoBERTa on the multilingual datasets (FB-en, FB-es, and FB-th) are shown in Table~\ref{tab:intent_multi_results}. Our models achieve state-of-the-art results on all 3 languages. While our wikiHow data in Spanish and Thai does improve models' performances, its effect is less salient than the English wikiHow data.

Our experiments above focus on settings where all available in-domain training data are used. However, modern task-oriented dialog systems must rapidly adapt to burgeoning services (e.g. Alexa Skills) in different languages, where little training data are available. To simulate low-resource settings, we repeat the experiments with exponentially increasing number of training examples up to 1,000. We consider the models trained only on in-domain data (+ID), those first pretrained on our wikiHow data in corresponding languages (+WH+ID), and those first pretrained on our English wikiHow data (+enWH+ID) for FB-es and FB-th. 

The learning curves of each dataset are shown in Figure~\ref{fig:intent_learning_curve}. Though the vanilla transformers models (+ID) achieve close to state-of-the-art performance with access to the full training data (see Table~\ref{tab:intent_eng_results} and \ref{tab:intent_multi_results}), they struggle in the low-resource settings. When given up to 100 in-domain training examples, their accuracies are less than 50\% on most datasets. In contrast, our models pretrained on our wikiHow data (+WH+ID) can reach over 75\% accuracy given only 100 training examples on all datasets. 

As our model performances exceed 99\% on Snips and FB-en, the concern arises that these intent detection datasets are ``solved''. We address this by performing error analysis and providing future outlooks for intent detection. 

Our model misclassifies 7 instances in the Snips test set. Among them, 6 utterances include proper nouns on which intent classification is contingent. For example, the utterance ``please open Zvooq'' assumes the knowledge that Zvooq is a streaming service, and its labelled intent is ``Play Music.'' 

Our model misclassifies 43 instances in the FB-en test set. Among them, 10 has incorrect labels: e.g. the labelled intent of ``have alarm go off at 5 pm'' is ``Show Alarms,'' while our model prediction ``Set Alarm'' is in fact correct. 28 are ambiguous: e.g. the labelled intent of ``repeat alarm every weekday'' is ``Set Alarm,'' whereas that of ``add an alarm for 2:45 on every Monday'' is ``Modify Alarm.'' We only find 1 example an interesting edge case: the gold intent of ``remind me if there will be a rain forecast tomorrow'' is ``Find Weather,'' while our model incorrectly chooses ``Set Reminder.''

By performing manual error analyses on our model predictions, we observe that most misclassified examples involve ambiguous wordings, wrong labels, or obscure proper nouns. Our observations imply that Snips and FB-en might be too easy to effectively evaluate future models. 

State-of-the-art models now achieve greater than 99\% percent accuracy on standard benchmarks for intent detection. 
However, intent detection is far from being solved. The standard benchmarks only have a dozen intents, but future dialog systems will need to support many more functions with intents from a wide range of domains. To demonstrate that our pretrained models can adapt to unseen, open-domain intents, we hold out 5,000 steps (as utterances) with their corresponding goals (as intents) from our wikiHow dataset as a proxy of an intent detection dataset with more than 100,000 possible intents (all goals in wikiHow). 

For each step, we sample 100 goals with the highest embedding similarity to the correct goal, as most other goals are irrelevant. We then rank them for the likelihood that the step helps achieve them. Our RoBERTa model achieves a mean reciprocal rank of 0.462 and a 36\% accuracy of ranking the correct goal first. As a qualitative example, given the step ``find the order that you want to cancel,'' the top 3 ranked steps are ``Cancel an Order on eBay'', ``Cancel an Online Order'', ``Cancel an Order on Amazon.'' This hints that our pretrained models' can work with a much wider range of intents than those in current benchmarks, and suggests that future intent detection research should focus on open domains, especially those with little data.

In conclusion, by pretraining language models on wikiHow, we attain state-of-the-art results in 5 major intent detection datasets spanning 3 languages. The wide-ranging domains and languages of our pretraining resource enable our models to excel with few labelled examples in multilingual settings, and suggest open-domain intent detection is now feasible. 

The work above was published in \citet{zhang-etal-2020-intent}, in which I primarily contributed to all components. I have obtained approval from all collaborators to exclusively include this work in this thesis.

\subsection{Next-Event Prediction}
\label{sec:next_event_prediction}

In the previous Section, I have demonstrated that the \textsc{goal-step} event relation can effectively transfer to the task of intent detection. This provides evidence that injecting structured knowledge in LLMs leads to performance gain compared to end-to-end usage. Next, we will see how the \textsc{step-step temporal} event relation can transfer to next-event prediction tasks.

\begin{figure}
    \centering
    \includegraphics[width=\columnwidth]{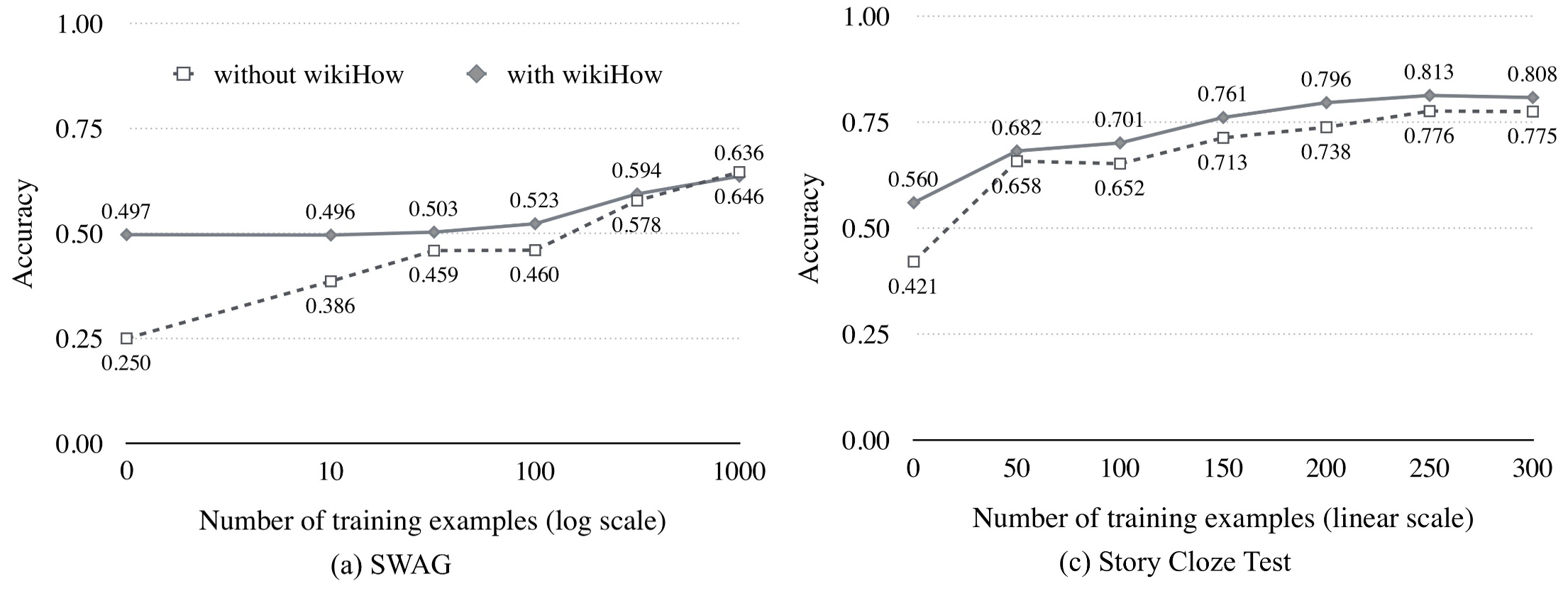}
    \caption{Accuracy of RoBERTa on SWAG and Story Cloze Test with different training set sizes, with and without being previously fine-tuned on our data of \textsc{step-step temporal} relation.}
    \label{fig:goal-step-evaluation}
\end{figure}

We consider two datasets in different domains.
\textbf{SWAG} \citep{zellers-etal-2018-swag} is a \textit{commonsense inference} dataset constructed from video captions. Given a context, a system chooses one event most likely to happen from four candidates. For transfer learning, we use up to 1,000 examples for training and the standard validation set. We use the model trained on our Step Inference task to transfer to this task. 

\textbf{Story Cloze Test} \citep{mostafazadeh-etal-2016-corpus} is a \textit{story understanding} dataset in the fiction domain, where a system chooses an ending to a 4-sentence-story from 2 candidates. We use up to 314 examples for training and 1,571 examples for validation, from the 2016 and 2018 data releases after removing duplicates. We use the model trained on our Step Ordering task to transfer to this task. To mimic the ``next sentence prediction'' format, we convert each example in our task to a ``next step prediction'' question with 4 prompt steps and 2 candidate steps, exactly one of which happens after the prompt.

Both datasets come with a sizeable training set, which may easily lead to overfitting. Hence, we only use a subset of their training data to simulate a low-resource scenario. Therefore, we are not comparing to the state-of-the-art performances involving the entire in-domain training sets. For each target task, we finetune a vanilla RoBERTa model and one pretrained on our data described in Section~\ref{sec:goal_step_in_domain} on increasingly larger portions of the target training set, and observe accuracy on the validation set, as the test set labels are not publicly available.

Figure~\ref{fig:goal-step-evaluation} shows the learning curves of the downstream tasks with an increasing number of their training samples, demonstrating a clear advantage of using our training data in low-resource settings. For SWAG, the model trained on our data has a zero-shot performance 24\% over chance, outperforming the vanilla model when up to 1,000 training examples are given. For Snips, the model trained on our data boasts an impressive 78\% zero-shot performance, approaching perfect accuracy rapidly after some in-domain training. For the Story Cloze Test which has the largest domain-mismatch with our data, the model still benefits from the knowledge learned from it consistently, given any portion of in-domain training data up to the full size in our experiment. These results show that the model learns real-world procedural knowledge from our wikiHow-based tasks, which can be readily applied to various domains and writing styles. 

In this section, I have respectively demonstrated the utility of the \textsc{goal-step} relation and the \textsc{step-step temporal} relation on the intent detection task and the next-event prediction task. Both applications illustrate the idea of \textit{structured event reasoning}, as opposed to end-to-end, structure-agnostic LLMs. For example, for intent detection, the key insight to success is that the utterance and intent constitute a portion of the relational event schema. Therefore, models equipped with that structured knowledge are on track to perform better.

\newpage
\section{Application of Event Schema}
\label{sec:script_generation}

\begin{figure}
    \centering
    \includegraphics[scale=0.2]{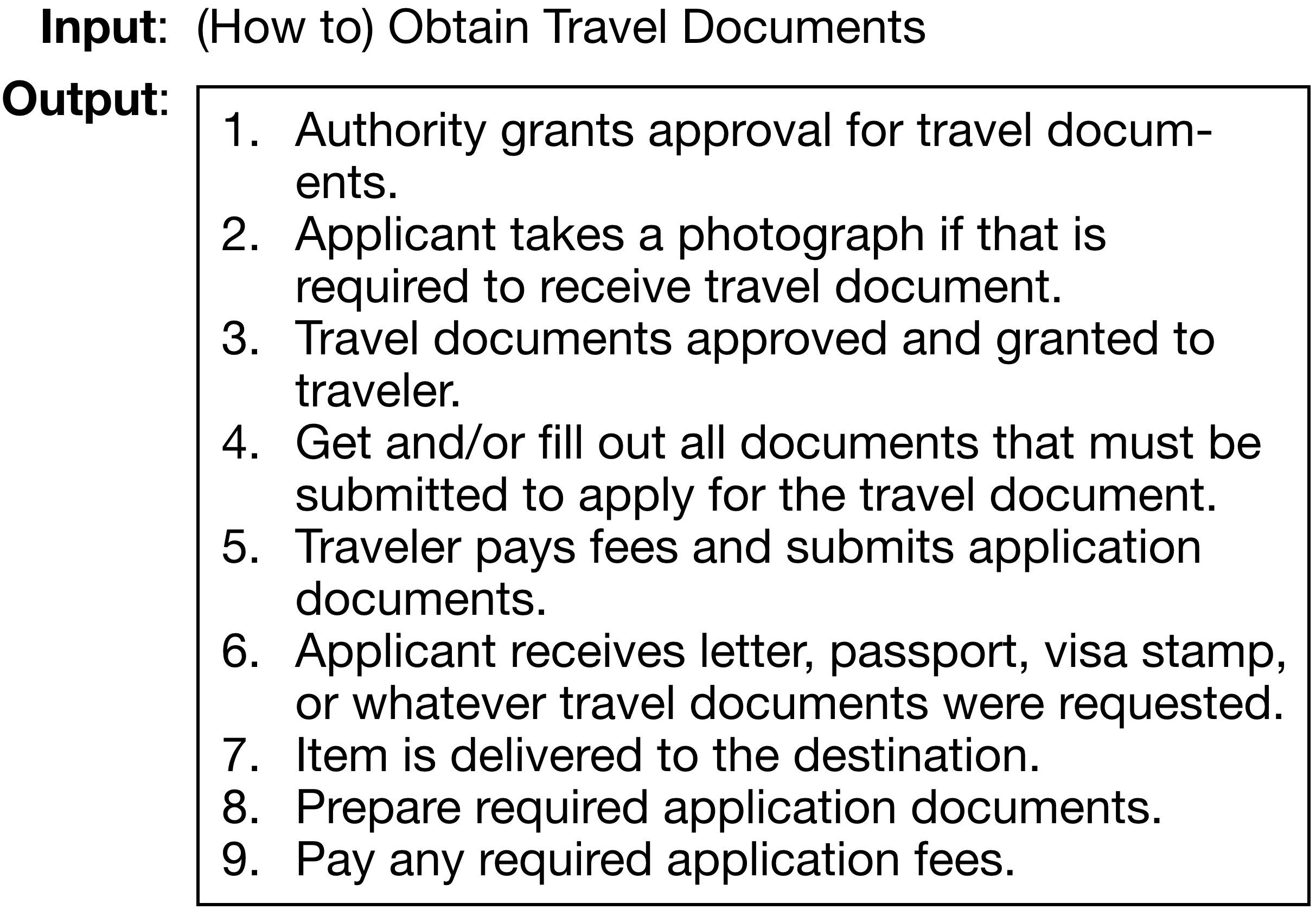}
    \caption{An example script constructed by our Step-Inference-Ordering pipeline in a zero-shot manner. The input is a \textit{goal}, and the output is an ordered list of steps.}
    \label{fig:thumbnail_script}
\end{figure}

In the two previous sections, I have shown that LLMs finetuned on the two event relations, collectively constituting the event schema, can outperform end-to-end usages in multiple downstream tasks and datasets. In this section, we will attempt to combine both abilities to tackle the challenging task of procedural \textbf{script generation}, generating a sequence of coherent steps given a goal. This process is also equivalent to generating an event schema tree of depth of 1, in which the nodes are the sub-steps or events and the edges are the two relations between each pair of the nodes.

\subsection{What is Script Learning?}

In both lines, it still remains an open problem what kind of automatic task most accurately evaluates a system's understanding of scripts. Most prior work has designed tasks focusing on various fragmented pieces of such understanding. For example, Narrative Cloze assesses a model's knowledge for completing a close-to-finished script. A related concept, Event Sequence Descriptions (ESD), on the other hand, evaluates script learning systems with the aforementioned variety of tasks, each touching upon a specific piece of script knowledge nonetheless (see Section~\ref{sec:related_work_events_procedures}). Recent work has also brought forth generation-based tasks, but mostly within an open-ended/specialized domain like story or recipe generation \citep{fan-etal-2018-hierarchical, xu-etal-2020-megatron}. 

\subsection{Goal-Oriented Script Construction}

We propose the task of \textit{Goal-Oriented Script Construction} (GOSC) to \textit{holistically} evaluate a model's understanding of scripts.Given a \textit{goal} (or the name of a script), we ask the model to construct the sequence of \textit{steps} (or events in a script) to achieve the goal. This task targets a model's ability to narrate an entire script, subsuming most existing evaluation tasks. Our rationale is that a model that \textit{understands} some scripts (e.g. how to ``\textit{travel abroad}'' and ``\textit{go to college}'') should be able to produce \textit{new} ones (e.g. how to ``\textit{study abroad}'') using the absorbed knowledge, close to how humans learn. 

Concretely, given a \textit{goal} $g$, a system constructs a complete script as an ordered list of \textit{steps} $S$, with a ground-truth reference $T$. As a hint of the desired level of granularity, we also provide an expected number of steps (or length of the script), $l$, as input. Depending on whether the set of possible candidate steps are given in advance, GOSC can happen in two settings: Generation or Retrieval. 

While almost all prior script learning work has focused on English, we leverage our wikiHow corpus to enable multilingual settings, just as Section~\ref{sec:intent_detection}. 

\subsection{Models}

\begin{figure*}
    \centering
    \includegraphics[scale=0.47]{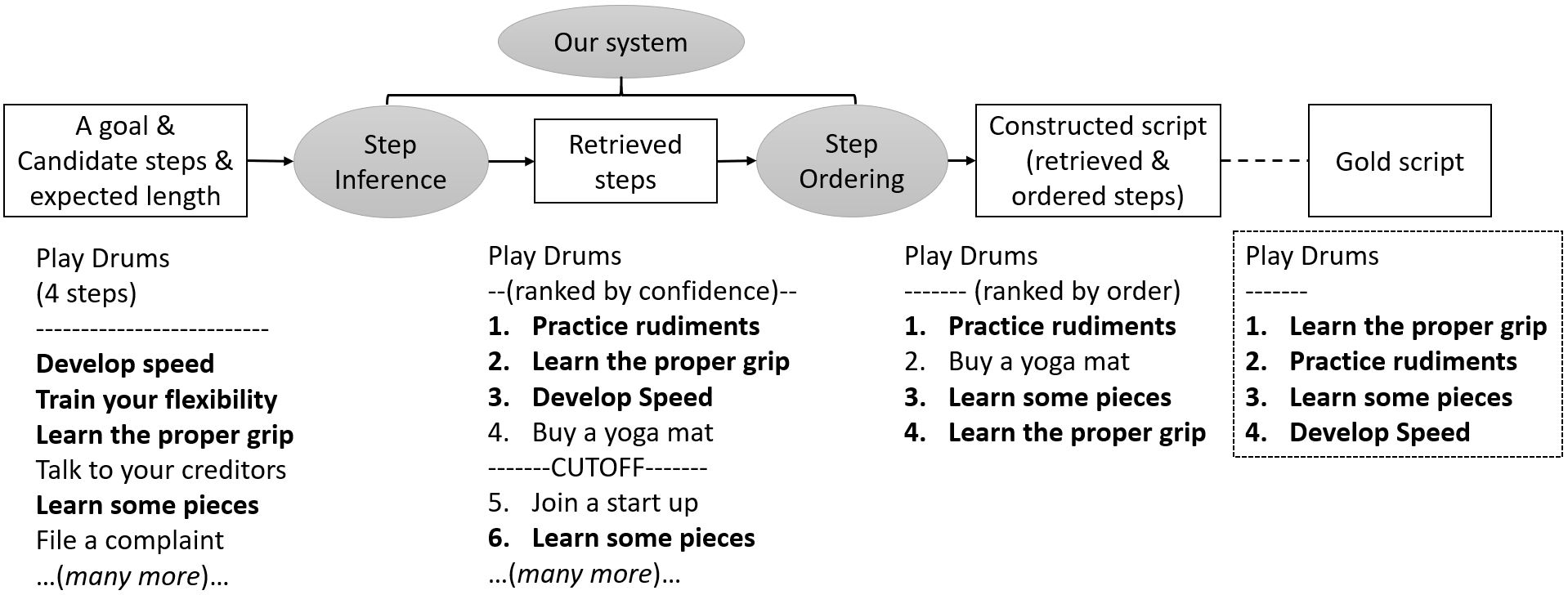}
    \caption{Our Step-Inference-Ordering pipeline for the GOSC Retrieval task. An example \textit{ordered} script is shown with example steps in the input and output. Those that appear in the ground-truth script is in bold.}
    \label{fig:script_pipeline}
\end{figure*}

We develop two systems based on state-of-the-art Transformers for the GOSC task.

\paragraph{Generation Approach}

For the Generation setting, we finetune mT5 \citep{xue2020mt5}, a pretrained generation model that is not only state-of-the-art on many tasks but also the only available massively multilingual one to date. 

During fine-tuning, we provide the goal of each article in the training set as a prompt, and train the model to generate the sequence of all the steps conditioned on the goal. Therefore, the model's behavior is similar to completing the task of inferring relevant steps and sorting them at once. At inference time, the model generates a list of steps given a goal in the test set. 

\paragraph{Retrieval Approach}

We then implement a \textit{Step-Inference-Ordering pipeline} for the Retrieval setting. Our pipeline contains a Step Inference model to first gather the set of desired steps, and a Step Ordering model to order the steps in the set. These models are based on our previous work described in Section~\ref{sec:goal_step_in_domain}. Under the hood, the models are pretrained XLM-RoBERTa \citep{conneau-etal-2020-unsupervised} or mBERT \citep{devlin-etal-2019-bert} for binary classification, both state-of-the-art multilingual representations.

Our Step Inference model takes a goal and a candidate step as input, and outputs whether the candidate is indeed a step toward the goal with a confidence score. During training, for every script, its goal forms a positive example along with each of its steps. We then randomly sample 50 steps from other scripts within the same wikiHow category and pair them with the goal as negative examples. The model predicts a label for each goal-step pair with a cross-entropy loss. During evaluation, for each script in the test set, every candidate step is paired with the given goal as the model input. We then rank all candidate steps based on the model confidence scores decreasingly. Finally, the top $l$ steps are retained, where $l$ is the required length. 

Our Step Ordering model takes a goal and two steps as input, and outputs which step happens first. During training, we sample every pair of steps in each ordered script as input to the model with a cross-entropy loss. During evaluation, we give every pair of retrieved steps as input, and count the total number of times that a step is ranked before others. We then sort all steps by this count to approximate their complete ordering.

\begin{figure*}
    \centering
    \includegraphics[scale=0.15]{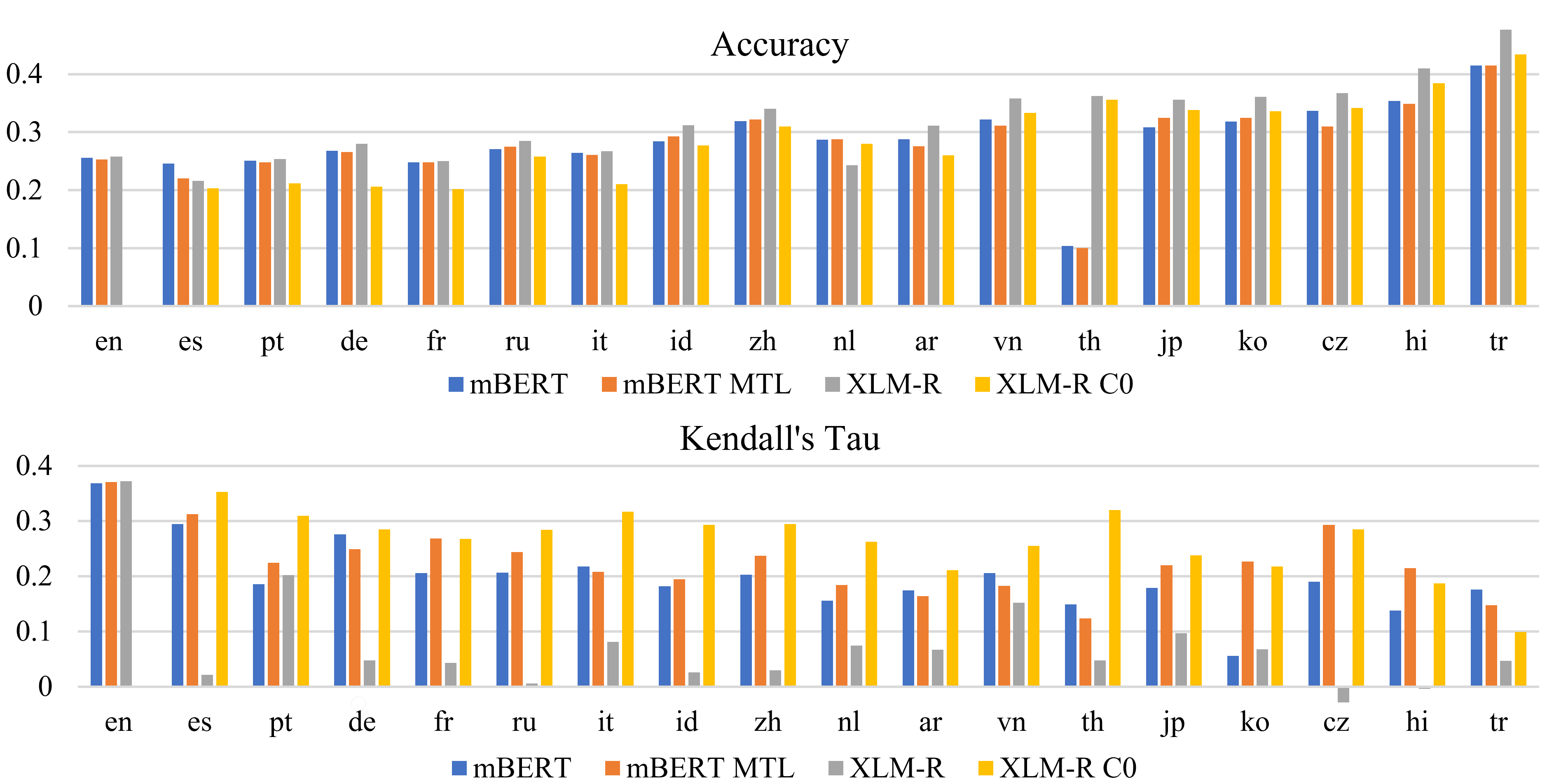}
    \caption{Detailed performance on each language from Table~\ref{table:retrieval_avg_results}.}
    \label{fig:script_lang_perf}
\end{figure*}

An illustration of our Step-Inference-Ordering pipeline is shown in Figure~\ref{fig:script_pipeline}. We also consider two additional variations.\\
\textbf{Multitask Learning} (MTL): The Step Inference and the Step Ordering models share the encoder layer, but have separate classifier layers. During training, the MTL system is then presented with a batch of examples from each task in an alternating fashion. During evaluation, the corresponding classifier is used. \\
\textbf{Cross-Lingual Zero-Shot Transfer} (C0): While there are abundant English training scripts, data in some other languages are scarce. Hence, we also attempt to directly evaluate the English-trained models on non-English data. 

\subsection{In-Domain Evaluation}
\label{section:in_domain_eval}
To demonstrate the performance of models on the GOSC task, we evaluate them on our multilingual wikiHow dataset using both automatic metrics and human judgments. The ultimate utility for this task is the extent to which a human can follow the constructed steps to accomplish the given goal. As direct user studies might be costly and hard to standardize, we carefully choose measures that adhere to this utility. By default, all models are trained and evaluated on the same language.

\begin{table}
\centering
\begin{tabular}{ cccccccccc } 
\toprule
Lang. & \hspace{0.2em} en & es & pt & de & fr & ru  \\
\midrule
Perp. & \hspace{0.2em} 17 & 11 & 24 & 97 & 46 & 79  \\
Bert. &  .823 & .702 & .682 & .677 & .718 & .682  \\
\midrule
\midrule
Lang. & \hspace{0.2em} it & id & zh & nl & ar & vn \\
\midrule
Perp. & \hspace{0.05em} 116 & 269 & 13,249 & 955 & 746 & 97  \\
Bert. &  .653 & .692 & .667 & .690 & .701 & .695  \\
\midrule
\midrule
Lang. & \hspace{0.2em} th & jp & ko & cz & hi & tr \\
\midrule
Perp. & 29,538 & 73,952 & 2,357 & 1,823 & 2,033 & 36,848  \\
Bert. & .701 & .679 & .692 & .682 & .704 & .665  \\
\bottomrule
\end{tabular}
\caption{Auto evaluation results for the Generation setting (Perplexity and BERTScore F1 measure). The performance of multilingual T5 is reported.}
\label{table:generation_ppl}

\end{table}

\begin{table}
\centering
\begin{tabular}{ l|cc|cccc|c|cc } 
 \toprule
\multirow{2}{1.6cm}{Model} & \multicolumn{2}{c|}{English only}  & \multicolumn{2}{c}{Avg. all lang.s} \\
& Acc. & Kendall's $\tau$  & Acc.   & Kendall's $\tau$   \\
 \midrule
mBERT & .256 & .369 & .286 & .198\\
mBERT MTL  & .253  & .371 & .283 & .226\\
XLM-R & .258  & .372 & .317 & .075\\
XLM-R C0  & -  & - & .291 & .264\\
\bottomrule
\end{tabular}%
\caption{Auto evaluation results for the Retrieval setting (Accuracy and Kendall's Tau). The performance of mBERT and XLM-RoBERTa, along with their multitask (MTL) and crosslingual zero-shot transfer (C0) variations, are reported. Multitask XLM-R and cross-lingual zero-shot mBERT are found to perform a lot worse and thus omitted.}
\label{table:retrieval_avg_results}
\end{table}

\paragraph{Auto Evaluation for Generation Setting}
To automatically evaluate models in the Generation Setting, we report \textbf{perplexity} and \textbf{BERTScore} \citep{zhang2019bertscore}, as two frequently used metrics for evaluating text generation.

The mean perplexity of mT5 on the test set of each language is shown in Table~\ref{table:generation_ppl}. The results show a large range of variation. To see if perplexity correlates with the data size, we conduct a Spearman's rank correlation two-tailed test. We find a Spearman's $\rho$ of $-0.856$ and a p-value of $1e-5$ between the perplexity and the number of articles in each language in our dataset; we find a Spearman's $\rho$ of $-0.669$ and a p-value of $2e-4$ between the perplexity and the number of tokens in each language in the mC4 corpus where mT5 is pretrained on. These statistics suggest a significant correlation between perplexity and data size, while other typological factors are open to investigation.

Table~\ref{table:generation_ppl} also shows the BERTScore F1 measure of the generated scripts compared against the gold scripts. Except for English (.82), the performance across different languages varies within a relatively small margin (.65 - .72). However, we notice that as a metric based on the token-level pairwise similarity, BERTScore may not be the most suitable metric to evaluate scripts. It is best designed for \textit{aligned} texts (e.g. a machine-translated sentence and a human-translated one), whereas in scripts, certain candidate steps might not have aligned reference steps. Moreover, BERTScore does not measure whether the ordering among steps is correct. To address these flaws, we further perform human evaluation later.

\paragraph{Auto Evaluation for Retrieval Setting}
To automatically evaluate models in the Retrieval Setting, we first calculate accuracy, i.e. the percentage of predicted steps that exist in the ground-truth steps. To account for the ordering of steps, we also compute Kendall's $\tau$ between the overlapping steps in the prediction and the ground-truth.

The performance of our Step Inference-Ordering pipeline using mBERT and XLM-RoBERTa\footnote{XLM-RoBERTa is not able to converge on the training data for Step Ordering for all but 3 languages using a large set of hyperparameter combinations.} on all 18 languages are shown in Figure~\ref{fig:script_lang_perf}. 

Across languages, the results are generally similar with a large room for improvement. On average, our best system constructs scripts with around $30\%$ accuracy and around $0.2$ Kendall's $\tau$ compared to the ground-truth. Compared to the baseline, our multitask and cross-lingual zero-shot variations demonstrate significant improvement on ordering. This is especially notable in low-resource languages. For example, MTL on Korean and C0 on Thai both outperform their baseline by 0.17 on Kendall's $\tau$.

\paragraph{Human Evaluation}

To complement automatic evaluation, we ask 6 annotators\footnote{The annotators are graduate students and native or proficient speakers of the language assigned.} to each \textit{edit} 30 output scripts by the Step-Inference-Ordering pipeline and mT5 in English, French, Chinese, Japanese, Korean and Hindi, respectively. The \textit{edit} process consists of a sequence of two possible actions: either 1) delete a generated step entirely if it is irrelevant, nonsensical or not a reasonable step of the given goal, or 2) move a step somewhere else, if the order is incorrect. Then, the generated script is evaluated against the edited script in 3 aspects:\\
\textbf{Correctness}, approximated by the length (number of steps) of the edited script over that of the originally constructed script (c.f. precision);\\
\textbf{Completeness}, approximated by the length of the edited script over that of the ground-truth script (c.f. recall);\\
\textbf{Orderliness}, approximated by Kendall's $\tau$ between overlapping steps in the edited script and the generated script.\footnote{In this formulation, the correctness and completeness of a retrieval-based model are equal, since the length of its constructed script is equal to that of the ground truth script by definition.}

The results are shown in Table~\ref{table:generation_human_results}. While the constructed scripts in the Retrieval setting contain more correct steps, their ordering is significantly worse than those in the Generation setting. This suggests that the generation model is better at producing \textit{fluent} texts, but can easily suffer from hallucination.

\begin{table}
\centering
\begin{tabular}{ l|cccccc } 
\toprule
\multicolumn{7}{c}{Retrieval: Step-Inference-Ordering pipeline} \\
\midrule
Language & en & fr & zh & jp & ko & hi  \\
Correctness & .70 & .39 & .50 & .49 & .45 & .82  \\
Completeness & .70 & .39 & .50 & .49 & .45 & .82  \\
Orderliness & .45 & .38 & .16 & .12 & .10 & .75  \\
\bottomrule
\toprule
\multicolumn{7}{c}{Generation: mT5} \\
\midrule
Language & en & fr & zh & jp & ko & hi  \\
Correctness & .39 & .51 & .46 & .40 & .37 & .49  \\
Completeness & .35 & .40 & .46 & .30 & .36 & .41  \\
Orderliness & .82 & .46 & .60 & .81 & .69 & .88  \\
\bottomrule
\end{tabular}
\caption{Human judgments of correctness, completeness and orderliness of the output of the Step-Inference-Order pipeline and the mT5 model for the same set of 30 gold scripts, in six languages. }
\label{table:generation_human_results}

\end{table}

\begin{figure*}
    \centering
    \includegraphics[scale=0.19]{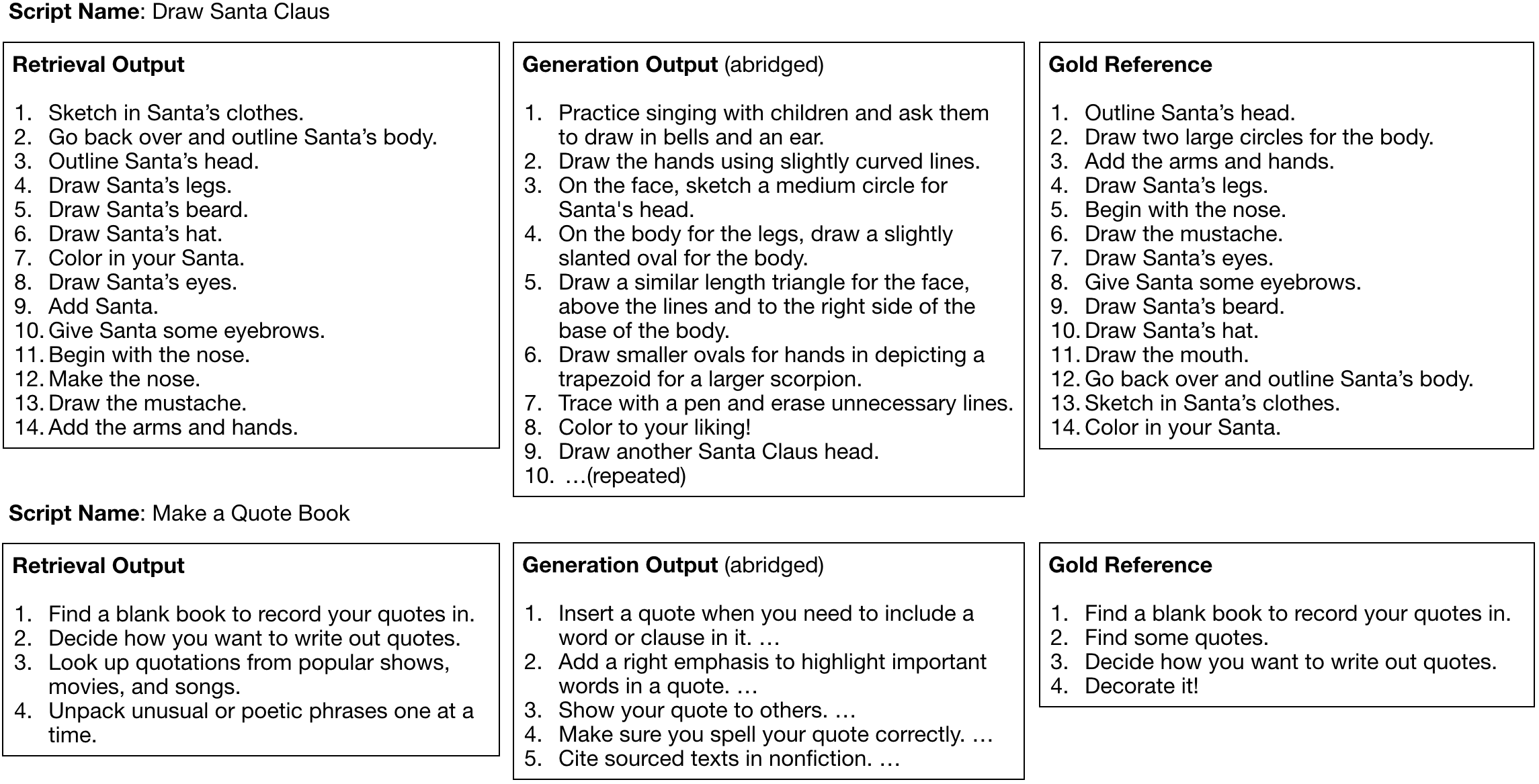}
    \caption{Two example scripts constructed by our Retrieval and Generation approaches.}
    \label{fig:qual_example}
\end{figure*}

\paragraph{Qualitative Examples}
\label{section:in_domain_qual_examples}
To understand models' behavior, we present two representative scripts produced by the mBERT Retrieval model and the mT5 Generation model side by side, accompanied by the ground-truth script, shown in Figure~\ref{fig:qual_example}.

The retrieved ``Draw Santa Claus'' script has a high step accuracy (85\%), with a reasonable ordering of drawing first the outline and then details. The generation output is more off-track, hallucinating irrelevant details like ``singing'' and ``scorpion'', despite being on the general topic of drawing. It also generates more repetitive steps (e.g. the head is drawn twice), most of which are abridged. 

As for ``Make a Quotebook'', the retrieved script has a 50\% step accuracy. The third step, though not in the gold reference, is similar enough to ``find some quotes'', suggesting that our exact match evaluation isn't perfect. In the generated script, all steps are also generally plausible, but some essential steps are missing (e.g. find a book, find quotes). This suggests that the generation model dwells too much on the details, ignoring the big picture.

These patterns in the two scripts are common in the model outputs, a larger sample of which is included in the Supplementary Materials.

\subsection{Out-Domain Evaluation}

To show the potential of our model for transfer learning, we use the retrieval-based Step-Inference-Ordering pipeline finetuned on wikiHow to construct scripts for other datasets and domains. We quantitatively evaluate our model on 4 other script learning corpora, and qualitatively analyze some constructed scripts in a case study.

\paragraph{Quantitative Evaluation}
Since no multilingual script data are available yet, we perform transfer learning experiments on 4 other English script corpora, OMICS \citep{singh2002open}, SMILE \citep{regneri-etal-2010-learning}, DeScript  \citep{wanzare-etal-2016-crowdsourced}\footnote{The above 3 corpora are all obtained from \url{http://www.coli.uni-saarland.de/projects/smile/}}, and the KAIROS Schema Learning Corpus (LDC2020E25). The first 3 pertain to human activities, while the last is in the military and political domain. They are all in the format of different \textit{scenarios} (e.g. ``eat in a restaurant'', similar to our \textit{goal}) each with a number of \textit{event sequence descriptions} (ESDs, similar to our \textit{steps}). Statistics for each corpus are in Table~\ref{table:transfer_results}.

\begin{table}
\small
\centering
\begin{tabular}{ l|cc|cccc|c|cc } 
 \toprule
\multirow{2}{*}{\begin{tabular}[c]{@{}l@{}}Corpus \end{tabular}} & \multicolumn{2}{c|}{Corpus Stats.}  & \multicolumn{2}{c}{Results} \\
& Scenarios & ESDs  & Acc.   & Kendall's $\tau$   \\
 \midrule
SMILE & 22 & 386 & .435 & .391\\
OMICS  & 175  & 9044 & .346 & .443\\
DeScript & 40  & 4000 & .414 & .418\\
KAIROS  & 28  & 28 & .589 & .381\\
\bottomrule
\end{tabular}
\caption{The zero-shot GOSC Retrieval performance of XLM-RoBERTa finetuned on wikiHow on 4 target corpora.}
\label{table:transfer_results}

\end{table}

For each dataset, we select the ESD with the most steps for every scenario as a representative script to avoid duplication,
thus converting the dataset to a GOSC evaluation set under the Retrieval setting. We then use the XLM-RoBERTa-based Step-Inference-Ordering pipeline trained on our English wikiHow dataset to directly construct scripts on each target set, and report its zero-shot performance in Table~\ref{table:transfer_results}. We see that $30\%-60\%$ steps are accurately retrieved, and around $40\%$ are correctly ordered. This is close to or even better than the in-domain results on our English test set. As a comparison, a random baseline would have only 0.013 Accuracy and 0.004 $\tau$ on average. Both facts indicate that the script knowledge learned from our dataset is clearly non-trivial.

\paragraph{KAIROS Case Study: The \textit{Bombing Attack} Scripts}
\label{section:case_study}
To explore if the knowledge about \textit{procedural} scripts learned from our data can also facilitate the zero-shot learning of \textit{narrative} scripts, we present a case study in the context of the DARPA KAIROS program\footnote{\href{https://tinyurl.com/yxwztj3j}{{\tt www.darpa.mil/program/knowledge- directed-artificial-intelligence-reasoning -over-schemas}}}. One objective of KAIROS is to automatically induce scripts from large-scale narrative texts, especially in the military and political domain. We show that models trained on our data of commonplace events can effectively transfer to vastly different domains.

With the retrieval-based script construction model finetuned on wikiHow, we construct five scripts with different granularity levels under the \textit{Improvised Explosive Device (IED) attack} scenario: ``Roadside IED attack'', ``Backpack IED attack'', ``Drone-brone IED attack'', ``Car bombing IED attack'', ``IED attack''. 
We take the name of each script as the input \textit{goal}, and a collection of related documents retrieved from Wikipedia and Voice of America news as data sources for extracting step candidates. 

Our script construction approach has two components. First, we extract all events according to the KAIROS Event Ontology from the documents using OneIE \citep{lin-etal-2020-joint}. The ontology defines 68 event primitives, each represented by an \textit{event type} and multiple \textit{argument types}, e.g. a Damage-type event has arguments including Damager, Artifact, Place, etc. OneIE extracts all event instances of the predefined primitives from our source documents. Each event instance contains a \textit{trigger} and several \textit{arguments} (e.g. Trigger: ``destroy'', Damager: ``a bomber'', Artifact: ``the building'', ... ). All event instances form the candidate pool of steps for our target script.

Since the events are represented as trigger-arguments tuples, a conversion to the raw textual form is needed before inputting them into our model. This is done by automatically instantiating the corresponding event type template in the ontology with the extracted arguments. If an argument is present in the extracted instance, we directly fill it in the template; else, we fill in a placeholder word (e.g.``some'', ``someone'', depending on the argument type). For example, the template of Damage-type events is ``$\langle arg1 \rangle$ damaged $\langle arg2 \rangle$ using $\langle arg3 \rangle$ instrument'', which can be instantiated as ``A bomber damaged the building using some instrument''). Next, we run the Step Inference-Ordering Pipeline described before on the candidate pool given the ``goal''. The only modification is that since we don't have a gold reference script length in this case, all retrieved steps with a confidence score higher than a threshold (default=0.95) are retained in the final script.

We manually evaluate the constructed scripts with the metrics defined in Section~\ref{section:in_domain_eval}, except \textit{Completeness} as we don't have gold references. The 5 constructed scripts have an average \textit{Correctness} of 0.735 and \textit{Orderliness} of 0.404. Despite the drastic domain shift from wikiHow to KAIROS, our model can still exploit its script knowledge to construct scripts decently. An example script, ``Roadside IED attack'', is shown in Figure~\ref{fig:IED_example}. All the steps retrieved are sensible, and most are ordered with a few exceptions (e.g. the ManufactureAssemble event should precede all others).\footnote{More details on the format of the script, all five constructed scripts, the event ontology, and a list of news documents used can be found in the Supplementary Materials.} 

\begin{figure}
    \centering
    \includegraphics[scale=0.2]{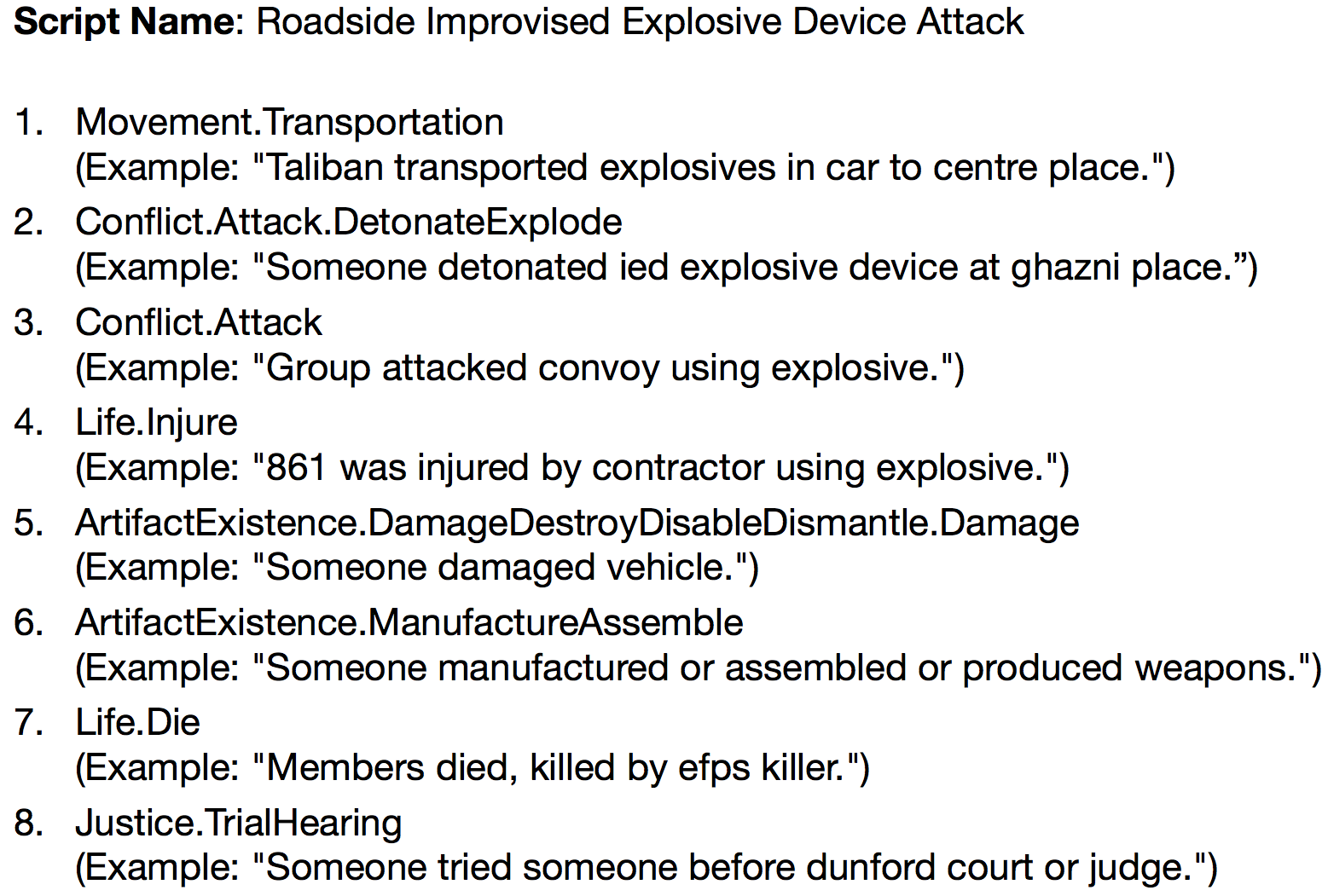}
    \caption{An example narrative script produced by our retrieval-based pipeline trained on wikiHow. Each event is represented by its Event Type and an example sentence.}
    \label{fig:IED_example}

\end{figure}

The work above was published in \citet{lyu-etal-2021-goal}, in which my collaborator Qing Lyu and I contributed equally in roughly all components. I have obtained approval from all collaborators to exclusively include this work in this thesis. 

\section{Summary}
Until this point, I have defined a relational event schema using two event relations. Recall that in the very beginning of this thesis, I outlined the two desirable metrics of any method: performance of trustworthiness. By fine-tuning pre-trained LLMs on a natural language representation of these relations, I have shown an increase of performance over end-to-end usage. For end-users, this methodology also brings about an increased sense of trustworthiness. For example, to tackle an intent detection task, a user using an end-to-end LLM would not know why or how the model succeeds or fails, whereas a user using an LLM finetuned on the \textsc{goal-step} event relation at least knows that the model is equipped with a particular piece of knowledge helpful to tackle the current task. For the script generation task, our proposed pipeline is much more transparent (c.f., neural module networks described in Section~\ref{sec:reasoning_framework}), as the model outputs of the two stages (step retrieval and step ordering) are mechanically combined as the final output. Such is a case of structured reasoning argued in this thesis. For a variety of event-centric tasks, one may represent an event as our proposed schema, and use either its components of itself as a whole to solve various problems.

However, there are still two fundamental issues. First, the atomic unit of the current event schema is still events (sub-events). As a result, reasoning tasks that involve more fine-grained information cannot be tackled. This calls for a more general representation of events. Second, our approach of fine-tuning LLMs is reliant on a sizeable amount of training data. Moreover, it is challenging for end-users to understand how an LLM generates an output based on a large amount of training data. Both issues lead to a lack of user-control, or the ability to improve and trust the model. In the next chapter, I attempt to tackle both issues.

\chapter{EVENT-ENTITY SCHEMA: A SEMI-SYMBOLIC REPRESENTATION}
\label{chap:entity}

Imagine teaching a kid or a robot how to cook. Midway through the process, you are asked the following question. 
\begin{displayquote} \centering
    \textit{Is it safe to touch (the center of) the pan?}
\end{displayquote}

Essentially, this question calls for reasoning about the event ``\textit{touch the pan}'' given certain circumstances. As discussed before, one may naively pose this question directly to an end-to-end LLM, a baseline approach that I have shown to have many shortcomings. As before, a reasonable alternative may be using a structured event representation. Unfortunately, the relational event schema described in the last chapter would not work here, because the event ``\textit{touch the pan}'' is neither a goal nor a step. Fundamentally, the answer to this question does not depend on the relation among its sub-events or any other events. To answer this question, most humans would realize a connection to an \textbf{entity state}:
\begin{displayquote} \centering
    \textit{Is it safe to touch the pan only if the pan is cool.}
\end{displayquote}
The above statement describes a logic statement (i.e., inference) that can be written semi-formally as:
\begin{equation} \label{eq:inference}
    \texttt{entity\_has\_attribute} (pan, cool) \Rightarrow \texttt{event\_can\_happen} (safe\ to\ touch\ the\ pan)
\end{equation}
The above formula indicates two items that must be inferred: 1. whether the entity \textit{pan} is \textit{hot}, and 2. the causality between the state of an entity and the plausibility of an event. In this particular instance, the former can be inferred from the context, while the latter is commonsense. In real life involving complex scenarios, inferring either of the two items is highly challenging, and therefore LLMs are reasonably good tools to do so. Instead of using LLMs to directly answer the question, we have now decomposed the task and use LLMs in a modular fashion. 

The thought process above demonstrates the neurosymbolic methodology. The neural part naturally refers to the LLMs whose input is free-formed natural language. The symbolic part refers to how we decompose the task, grounding the concepts of \textit{pan}, \textit{cool}, \texttt{entity\_has\_attribute}, and even the inference operation $\Rightarrow$ into symbols. Therefore, answering the question ``\textit{is it safe to touch the pan?}'' becomes a two
-hop reasoning problem, where each sub-problem can either be answered formally with symbols, or using LLMs with natural language. For example, one might use LLMs to figure out that the \textit{pan} is \textit{cool}, a symbolic expression, based on some textual description. At this point, if the causality in Equation~\ref{eq:inference} is conveniently provided or otherwise inferred, we can deterministically deduce the final answer ``\textit{it is safe to touch the pan}'' (more in Chapter~\ref{chap:world}). Otherwise, we may also express ``\textit{the pan is cool}'' as a half-language, half-symbolic expression and feed it to LLMs to help them arrive at the answer. In either approach, we leverage the strong-suits of both neural and symbolic methods, whereas in this Chapter, I will focus on a semi-symbolic representation that is predicted by one LLM and fed into another to answer questions (see Figure~\ref{fig:pipeline_semi_sl}).

\begin{figure}
    \centering
    \includegraphics[scale=0.6]{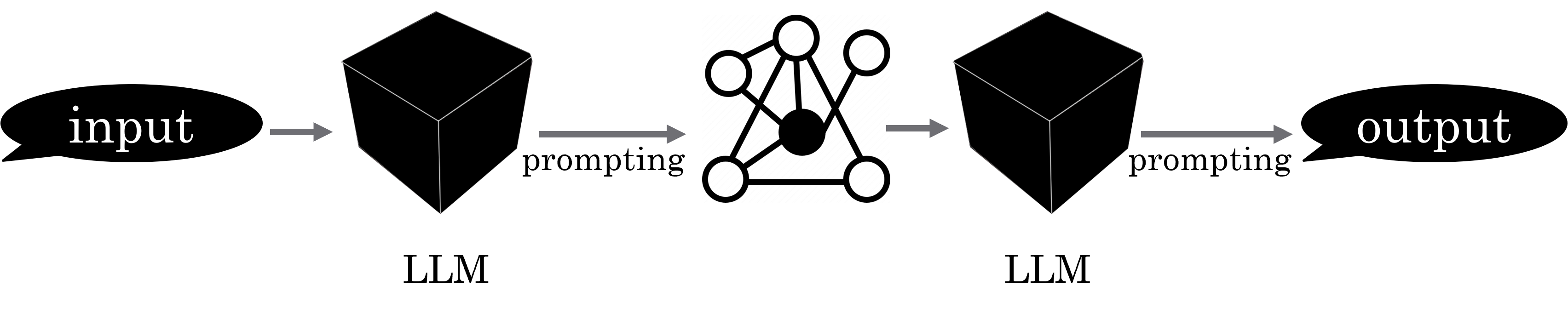}
    \caption{An illustration of my proposed pipeline leveraging a semi-symbolic representation of entities. The LLMs are interacted via few-shot prompting.}
    \label{fig:pipeline_semi_sl}
\end{figure}

I propose a semi-symbolic representation based on entities called an \textbf{entity schema} (Figure~\ref{fig:openpi}) that builds upon the previously discussed event-relation schema which models an event as a procedure including a goal event (e.g., \textit{defog a window using potatoes}) and a sequence of ordered step events (e.g., \textit{rub the cut side of potato on the window}). For each step, the schema models an array of 4-tuples describing an entity state change. Each 4-tuple contains an entity, an attribute, a state before the step, and a state after the step (e.g., \textit{the window}'s \textit{texture} was \textit{smooth} before and \textit{sticky} after). Essentially, the entity schema is a matrix where the axes are step, entity, and attribute, while the value is the before and after states. Unlike the previously discussed relational event schema where events are expressed as natural language, here, the entities are both textual and symbolic (i.e., semi-symbolic), for they can not only be consumed by LLMs (this Chapter), but also possibly be consumed by symbolic algorithms or grounded to some environment (Chapter~\ref{chap:world}). To learn the entity schema, I first enhance an existing dataset \citep{tandon-etal-2020-dataset} by canonicalizing the entities, so that different mentions of \textit{window}, \textit{glass}, \textit{pane} all fall under a unified symbol (Section~\ref{sec:openpi}). Using this enhanced dataset, I tune LLMs that can predict an entity schema given a procedure. I demonstrate their utility via a downstream task to causally reason about entities and events. There, I show that modern LLMs trained on a mixture of code and text can effectively leverage the event schema in the form of a Python-code (Section~\ref{sec:crepe}). Finally, I show that such a code-like form does not perform equally well on a variety of other NLP tasks (Section~\ref{sec:symbolic_form}). 

In the Chapter~\ref{chap:relation}, I have been using the pre-train then fine-tune paradigm with the BERT-family LLMs. In contrast, the discussions in this Chapter will revolve around the in-context learning paradigm made possible by the state-of-the-art GPT-family LLMs (see categorization described in Section~\ref{sec:llm_history}), where the LLMs are interfaced via few-shot prompting. This paradigm shift is preferred as it eliminates the need of a sizeable fine-tuning dataset, offering more flexibility.

\begin{figure}
    \centering
    \includegraphics[scale=0.6]{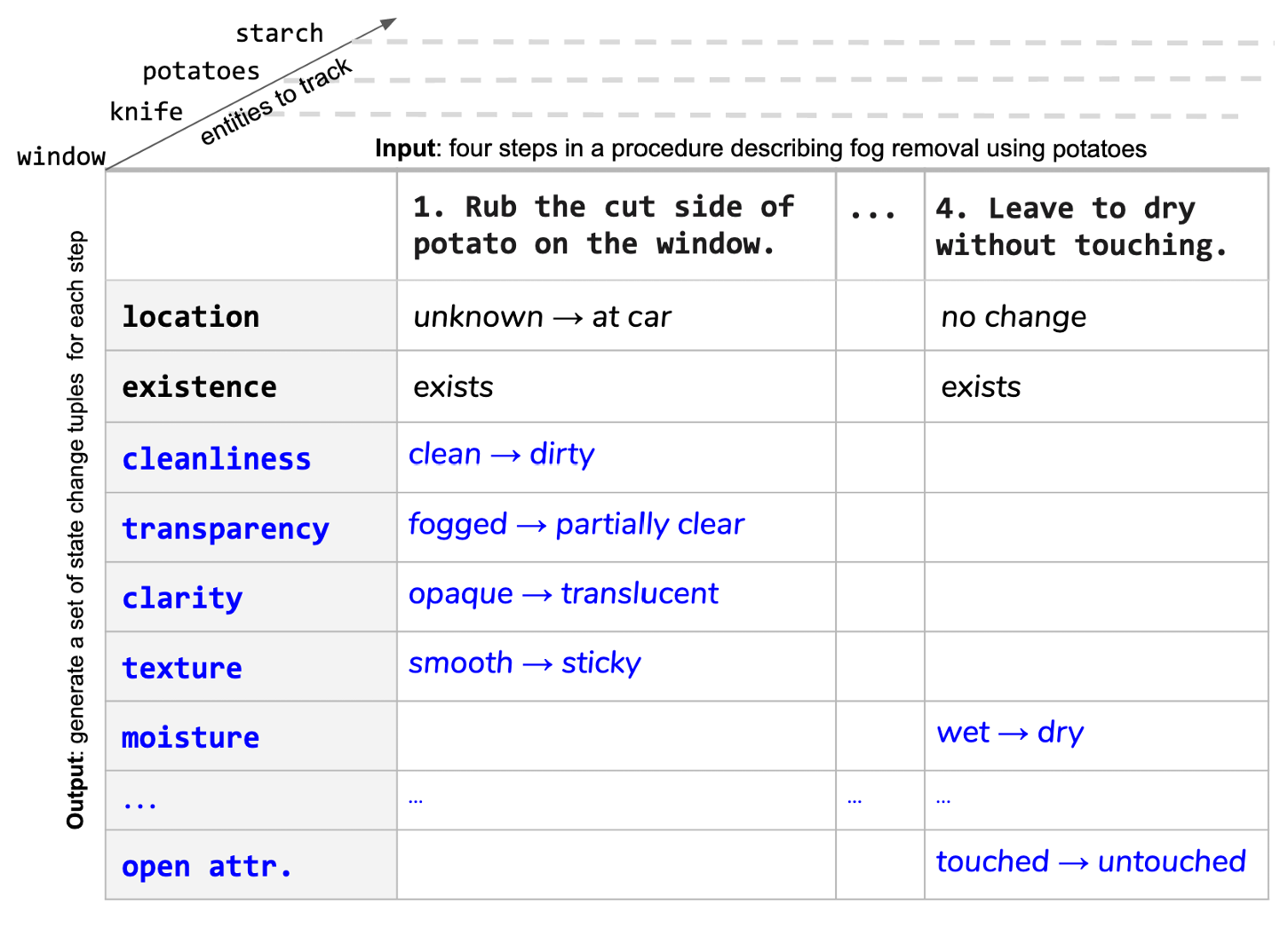}
    \caption{An example from the OpenPI dataset \citep{tandon-etal-2020-dataset}.}
    \label{fig:openpi}
\end{figure}

\section{Learning Entity States}
\label{sec:openpi}

My overarching goal is to train and evaluate LLMs to predict an entity schema given a procedure, so that this information can be used for reasoning in later sections. To do that, my collaborators and I start with the \openpi dataset \citep{tandon-etal-2020-dataset} with crowdsourced annotations of the entity states in procedural texts (Figure~\ref{fig:openpi}), identify its shortcomings, and
and propose an improved \openpitwo dataset.

\subsection{Background}
\label{sec:related_work}

Tracking entity states in procedural texts is closely related to many NLP reasoning tasks. To name a few, question answering (QA) about events (e.g., \textit{why use gloves when retrieving the tray from the oven}) often require knowledge of entity states (e.g., \textit{the tray becomes very hot while in the oven}) \citep{tandon-etal-2019-wiqa,spiliopoulou-etal-2022-events}; planning \citep{wang-etal-2022-scienceworld,brohan2023can} largely involves actions upon entities resulting in state changes. Procedural entity tracking is challenging in itself, requiring much understanding of an implicit environment as well as external knowledge of what events affect which entities, and how. 

Prior work on entity state tracking spans various disciplines of AI. For instance, object tracking, a sub-task of entity state tracking, has led to much work in both robotics \citep{wang2007simultaneous} and computer vision \citep{comaniciu2003kernel}. In NLP, early efforts focus on synthetic, closed-domain data \citep{weston2015towards, long-etal-2016-simpler} and more recent ones shift attention to real-world procedures \citep{bosselut2017simulating, dalvi-etal-2018-tracking,gupta-durrett-2019-tracking,du-etal-2019-consistent,mysore-etal-2019-materials} with a closed set of entities and attributes. The only open-ended dataset to our knowledge is still \openpi \citep{tandon-etal-2020-dataset} which we build on.

A small body of work on entity salience has focused on annotating entity salience in news articles and web pages for better information retrieval, recommendation, and linking \citep{gamon2013identifying,dunietz-gillick-2014-new,dojchinovski-etal-2016-crowdsourced,trani2018sel,wu-etal-2020-wn}. In contrast, we focus on entities in procedural texts, situating our work in script learning, robotic execution, automatic planning and reasoning, etc. Due to this mismatch of purpose, the definition, annotation process, and downstream applications of our entity salience and theirs are all fundamentally different.

\begin{figure}
    \centering
    \includegraphics[width=\columnwidth]{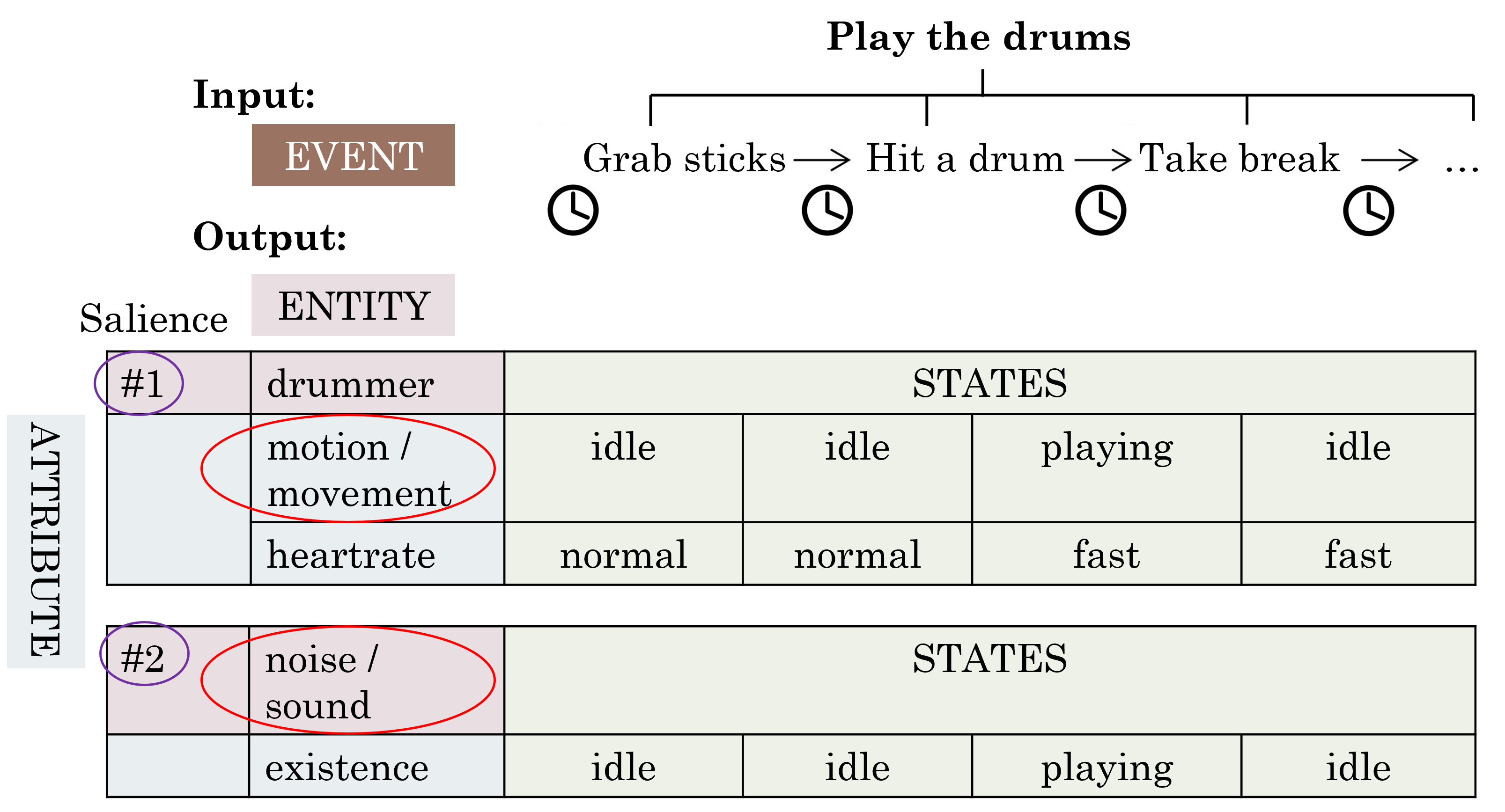}
    \caption{For each step in a procedure, \openpi annotates the state change of attributes of entities. Our \openpitwo additionally canonicalizes the entities and attributes (red circles) and includes their salience scores (purple circles). }
    \label{fig:openpi1v2}
\end{figure}

We propose the \openpitwo dataset which builds on \openpi \citep{tandon-etal-2020-dataset} (Open Procedural Inference), a large-scale dataset for tracking entity states in procedural texts from \url{wikiHow.com}. It contains annotations of entities, attributes, and state changes for each step (e.g., after the step ``set the pan in a heated oven'', the \textit{pan}'s \textit{temperature} was \textit{cool} before and \textit{hot} afterwards). \openpitwo features two critical improvements (see Figure~\ref{fig:openpi1v2} for a demonstration of key features of \openpi and \openpitwo):
\begin{enumerate}[topsep=0pt,itemsep=-1ex,partopsep=1ex,parsep=1ex,leftmargin=*]
    \item \textbf{Canonicalization}. Originally, different mentions of the same entity or attribute render evaluation difficult. Here, we prompt LMs to effectively cluster the entities and attributes.
    \item \textbf{Entity Salience}. Originally, all entities that undergo changes are listed in parallel. Here, we provide both human and model-predicted annotations of their salience.
\end{enumerate}
Regarding canonicalization, clustering paraphrases evidently allows for fairer evaluation. Moreover, as our task of predicting entities, attributes, and states is a generation task with imperfect and incomplete ground-truth references, we show that expanding each entity or attribute cluster with possible paraphrases (thus providing more references) is effective for reducing the false-negative rate. We then comprehensively report various state-of-the-art LMs' performance of entity tracking on \openpitwo. 

Regarding entity salience, we provide both manually annotated and automatically calculated labels. We evaluate them based on correlation with ground-truth data, and show that LMs can reliably predict entity salience with a close-to-human performance. We argue that salient entities acts as a means of \textit{compression} of the most critical information in procedural texts, similar to saliency maps in computer vision \citep{simonyan2013deep}. We proceed to qualitatively and quantitatively show that salient entities, as chain-of-though of LM prompting, benefit downstream tasks such as QA and classical planning, while reducing cost by excluding less important entities in the prompt.

\subsection{Canonicalization}\label{sec:canonicalization}
In the original \openpi dataset, the entities and attributes that undergo change were written by crowd workers. Consequently, the dataset contains different ways of expressing the same entity (e.g., \textit{coffee maker}, \textit{coffee machine}, \textit{espresso machine} in a coffee-making procedure) or attribute (e.g., \textit{texture}, \textit{smoothness}, \textit{sheen} of a paint). Canonicalization by clustering the entities and procedures is thus important for two reasons: 1) it facilitates evaluation especially in a generation setting, where a model might be wrongly penalized for predicting a paraphrase of some correct entity or attribute; 2) it facilitates further annotation of features such as salience (\S\ref{sec:salience}) of the entities and attributes.

\paragraph{Clustering Entities and Attributes}
While canonicalization seems straightforward, it is non-trivial in \openpitwo because clustering is highly context-dependent. For example, the entity \textit{torso} and \textit{paper chunk} usually have nothing to do with each other, but in fact refer to the same thing in a procedure of making a paper bird. 

\begin{table*}
\small
\centering
    \begin{adjustbox}{max width=\textwidth}
    \begin{tabular}{ll}
        \toprule
        Role & Content \\
        \midrule
        User & \makecell[l]{I am trying to make coffee. First, I put some coffee beans and tap water into the corresponding compartment \\
        of the espresso machine. Then, I select the desired type of coffee to make produced. Then I put a mug under \\
        the espresso machine and press start. Do you get it?} \\[4mm]
        Assistant & Yes. \\[1mm]
        User & \makecell[l]{
        We have the following objects: \textit{water}, \textit{coffee maker}, \textit{coffee machine}, \textit{mug}, \\\textit{espresso machine}. Group those that refer to the same thing. You must include all the provided entities. Do \\
        not add any entity that is not provided in the list.}  \\[4mm]
        Assistant & \makecell[l]{<start of generation> The grouped objects are:\\  - ['water']\\  - ['coffee maker', 'coffee machine', 'espresso machine']\\  - ['mug']} \\
        \bottomrule
    \end{tabular}
    \end{adjustbox}
    \caption{Our chosen prompt for entity and attribute clustering.}
\label{tab:canon_prompt}
\end{table*}

\paragraph{Clustering} Due to the contextual nature of the task, we prompt one of the SoTA LMs
\texttt{gpt-3.5-turbo} (a.k.a. ChatGPT)\footnote{\url{platform.openai.com/docs/models/gpt-3-5}} as shown in Table~\ref{tab:canon_prompt}. We use 3-shot prompting, meaning that the complete prompt includes three handwritten examples and the prompt header of the example to be inferred, only containing the ``User'' role. The temperature is as default (0.7) and so are other hyperparameters. We aggregate output from five runs of ChatGPT as the final entity cluster and three runs for attribute cluster, as doing so is found to be empirically superior than single-pass generation\footnote{With results from the 5 runs, individual Entity clusters are added to the final cluster based on their number of occurrences. For instance, if \texttt{(pan, pot)} occurred 5 times, then it will be added to the final cluster first.}.

To see if our model can cluster entities and attributes effectively, we evaluate the results using precision, recall, and F1 scores with exact match against a set of manually-labeled clusters from 20 procedures in the development set. 


\begin{table}
    \centering
    \begin{tabular}{lll} \toprule
              & Entity & Attribute \\ \midrule
    Cluster Recall    & .425   & .881      \\
    Cluster Precision & .593   & .906      \\
    Cluster F1        & .495   & .893     \\\bottomrule
    \end{tabular}
    \caption{Evaluation of entity and attribute clustering.}
    \label{tab:cluster_performance}
\end{table}

We see that ChatGPT scores better in clustering attributes compared to entities. Error analysis shows that two factors contribute to this performance discrepancy. First, most attributes describe the physical properties of an entity. Therefore, attribute clusters are less context-dependent compared to entity clusters. Second, many attributes are shared amongst entities. For instance, out of 1,145 attribute annotations in the development set, 204 of them are \textit{"location"}.

\begin{table*}
\centering
\small
\begin{tabular}{lllll|llll|ll}
\toprule
                 & \multicolumn{4}{c|}{schemata (global)}                                  & \multicolumn{4}{c|}{schemata (local)} & \multicolumn{2}{c}{states} \\ \midrule
                 & F1        & F1+exp       & BS               & BS+exp              & F1        & F1+exp       & BS              & BS+exp & acc. & BS \\ \midrule
\texttt{gpt-3.5-turbo}    & .151 & .249 & .843 & .869 & .025 & .039  & .798 & .804 & .074 & .600 \\
\texttt{text-davinci-003} & .362 & .450 & .891 & .920  & .130 & .155  & .798 & .810 & .225 & .682 \\
\texttt{LLaMA 65B}        & .129 & .174 & .799 &  .820 & .045 & .060 & .801 & .800 &   .102  & .577 \\ 
\bottomrule
\end{tabular}
\caption{Exact match F1 or accuracy and BERTScore on the schemata and states prediction sub-tasks, with and without cluster expansion. The schemata sub-task is evaluated both globally (per-procedure) and locally (per-step).}
\label{tab:perf_1}
\end{table*}

\paragraph{Cluster expansion} Though the existing entities and attributes are now clustered in \openpitwo, there may still be other paraphrases that a model might rightfully predict and wrongfully penalized for. Thus, we again prompt ChatGPT to expand the clusters by generating paraphrases given a cluster of entities or attributes (prompt omitted). 

To evaluate the quality of entities and attributes generated from the expansion, we manually rate 20 procedures and find that 83.3\% of the generated, paraphrased entities and 59.4\% attributes are correct.
This is largely because entity names are oftentimes self-explanatory and less context-dependent whereas the attribute names and their meanings are highly dependant on the context. 

\begin{table}
    \centering
    \small
    \begin{tabular}{lllll} \toprule
    & \multicolumn{4}{c}{complete} \\ \midrule
      & F1        & F1+exp       & BS               & BS+exp \\ \midrule
    \texttt{gpt-3.5-turbo} & .016 & .016 & .772 & .790  \\
    \texttt{text-davinci-003} & .034 & .034 &  .807  & .821  \\
    \texttt{LLaMA 65B} & .117 & .117 & .429 & .440 \\ 
\bottomrule     
    \end{tabular}
    \caption{Exact match F1 and BERTScore of complete sentences including an entity, an attribute, a pre-state, and a post-state, following the original \openpi paper. Canonicalization and expansion lead to little help for exact match as it is only done on entity and attribute clusters, while the state names can still be expressed in many ways, causing false negatives.}
    \label{tab:perf_2}
\end{table}

\begin{table}
    \small
    \centering
    \begin{tabular}{lllll}
    \toprule
    & Correct & No change & Nonsense & Missing \\ \midrule
    \texttt{003} & 585 (82.3\%) & 106 (15.0\%) & 	14 (2.0\%) & 383 (20.3\%) \\
    \texttt{3.5} & 303 (59.4\%) & 173 (33.9\%) & 34 (6.7\%) & 218 (42.7\%) \\ \bottomrule   
    \end{tabular}
    \caption{Error analysis on the schemata prediction task using two SoTA LLMs.}
    \label{tab:error_analyis}
\end{table}

\paragraph{Utility: Evaluation of Entity Tracking} \label{sec:eval_entity_tracking}

Just as the original \openpi, \openpitwo is meant to be used to benchmark (or train) models on entity tracking -- given a step in a procedure, predicting the state changes that entities and their attributes undergo. With the entities and attributes in \openpitwo now fully canonicalized, evaluation can be done more fairly. To start with, we follow \citet{tandon-etal-2020-dataset} and predict one \textbf{complete} sentence: ``\textit{attribute} of \textit{entity} is \textit{pre-state} before and \textit{post-state} afterwards'', which is then compared to such sentences in the ground-truth data (Table~\ref{tab:perf_2}). We further make the evaluation more fine-grained by formulating two sub-tasks: i. predicting \textbf{schemata}, namely the entities and their corresponding attributes given a step (e.g., given ``turn on the oven'', the ``temperature'' of the ``rack'' undergo state changes), and ii. predicting the change of \textbf{states} given a step, an entity and an attribute (e.g., given the previous information, the state change is from ``cool'' to ``hot''). This evaluation of first predicting a skeleton tensor of entities and attributes is highly practical, and stands in stark contrast with most previous work (\S\ref{sec:related_work}) in closed-domain entity tracking, where states are predicted using \textit{given} entities and attributes.

On the development set, we run three SoTA LMs: \texttt{gpt-3.5-turbo}, \texttt{text-davinci-003}\footnote{\url{platform.openai.com/docs/models/gpt-3-5}} \citep{brown2020language}, and the open-source LLaMA 65B \citep{touvron2023llama}. For each model, we start by separately tackling each of the two sub-tasks\footnote{To avoid error propagation, for states prediction, the ground-truth entities and attributes are provided.}; namely, a model first predicts attributes of entities (schemata) given a step, and then predicts a pre-state and a post state (states) given the gold entity-attribute pair.  All experiments are via 1-shot prompting. 

\begin{table*}
\small
\centering
    \begin{adjustbox}{max width=\textwidth}
    \begin{tabular}{ll}
        \toprule
        Role & Content \\
        \midrule
        User & \makecell[l]{Here are some instructions on making coffee.\\- Buy fresh coffee beans.\\- Grind the coffee beans.\\- ...\\Now, I will provide you with a series of objects, and you will assign scores on a scale of 1-5 to them based on\\their importance in the instruction. Your answer should strictly be a numerical score followed by a one-\\sentence explanation.}  \\
        Assistant & Sure, I can help you with that. Please provide the objects.\\
        User & Coffee bean \\
        Assistant & <generation> 5 - the coffee beans are the most important ingredient in making coffee. \\
        \bottomrule
    \end{tabular}
    \end{adjustbox}
    \caption{Our chosen prompt for predicting global or procedure-wide entity salience. For local salience, the wording is similar with only one step provided.}
\label{tab:salience_prompt}
\end{table*}


For all settings, we consider both exact match (F1 for schemata and complete sentence prediction and accuracy for states prediction) and BERTScore \citep{zhang2019bertscore} based on \texttt{deberta-xlarge-mnli} \citep{he2021deberta}. 

For the schemata prediction sub-task (Table~\ref{tab:perf_1}), the atomic unit to be evaluated is an entity-attribute pair. We consider both a \textit{global} evaluation, where predictions are made per-procedure (e.g., what attributes of entities undergo state changes in the procedure), and a \textit{local} evaluation, where predictions are made per-step. This categorization will reappear in \S\ref{sec:salience_eval}. Schemata prediction is naturally influenced by our entity and attribute clusters. Hence, for exact match we report F1 scores based on exact matches where any entity-attribute prediction that falls under an cluster, obtained by taking a Cartesian product of an entity cluster and an attribute cluster, is considered a true positive. For BERTScore, we calculate the maximum score of a prediction against all entity-attribute strings within all ground-truth clusters. Then, we report the mean score among all predictions as a macro average.

The states prediction sub-task (Table~\ref{tab:perf_1}) is much more straightforward as the entity-attribute pairs are provided and a model only needs to predict a pre-state and a post-state for each. Thus, we simply report the exact match accuracy and BERTScore for each state.

\paragraph{Discussion and Error Analysis}

We observe that the predicting attributes of entities that undergo state changes is a highly challenging task even for SoTA LMs. Although evidently, expansion of clusters improves performance (fairly, as we have shown that the generated paraphrases are mostly correct), false-negatives that result in underestimation of models cannot be eliminated entirely. One interesting observation is that \texttt{text-davinci-003} greatly outperforms the supposedly more superior \texttt{gpt-3.5-turbo}. To gain even more insights into models' behavior, we analyze the model output for the schemata prediction sub-task. For each step, we annotate each entity-attribute prediction based on three labels:

\begin{itemize}[topsep=0pt,itemsep=-1ex,partopsep=1ex,parsep=1ex]
    \item Correct, where the entity-attribute indeed go through some changes;
    \item Incorrect, because the entity-attribute actually does not go through any changes;
    \item Incorrect, because the entity-attribute is non-sensical.
\end{itemize}

In addition, we add any entity-attribute pairs that should have been predicted as going through some change, to measure models' recall. We randomly sample 20 procedures to perform this error analysis and the results are shown in Table~\ref{tab:error_analyis}.

Regarding precision, we find that while the majority of the predicted entities are correct, many of the predicted associated attributes are generic ones that do not undergo any change either locally or globally. For example, for the step ``Purchase a blackboard eraser'', the attributes predicted by \texttt{text-davinci-003} for the entity \textit{eraser} are \textit{location} (correct), \textit{cleanness} (static locally), \textit{shape}, and \textit{size} (static globally). The issue is much more pronounced with \texttt{gpt-3.5-turbo}, with predictions such as \textit{location} of \textit{seller}, \textit{name} of \textit{brand}, etc, despite that the prompt clearly explains the desired output with an example. We attribute such performance discrepancy to \texttt{gpt-3.5-turbo}'s decreased ability to follow examples and its inability to understand nuanced instructions. Regarding recall, both models fail to predict many attributes that the human annotator deems changing. Upon qualitative inspection, most of these missing attributes are no less salient than the predicted ones, suggesting that this issue cannot be explained away with only missing secondary attributes which may be plenty.

We leave to future work the resolution of these issues, which can be mitigated by re-prompting the models by validating if the predicted attributes indeed undergo changes, or simply have them predict the state changes altogether in the first place. 

\begin{table}
\centering
    \begin{tabular}{lll}
        \toprule
         & \multicolumn{1}{c}{Annotations} & \multicolumn{1}{c}{Predictions} \\
         & Human (A2)  & LM  \\ \midrule
        Global & .759 & .719  \\
        Local & .578 & .400  \\
        \bottomrule
    \end{tabular}
    \caption{Pearson' $r$ with human annotations (A1).}
\label{tab:correlation}
\end{table}

\subsection{Salience} \label{sec:salience}
The original \openpi is annotated with many parallel entities in each procedure. Often, they vary greatly by importance in accomplishing the task. For example, in a procedure of cooking a steak, entities \textit{fish}, \textit{oven}, \textit{gloves}, and \textit{spice rack} might all be involved, while some are more indispensable than the rest. In \openpitwo, we define two types of entity salience: the global salience refers to the importance of an entity in accomplishing the goal of the procedure, whereas the local salience refers to that in a step. 

\begin{figure*}
    \centering
    \includegraphics[width=\textwidth]{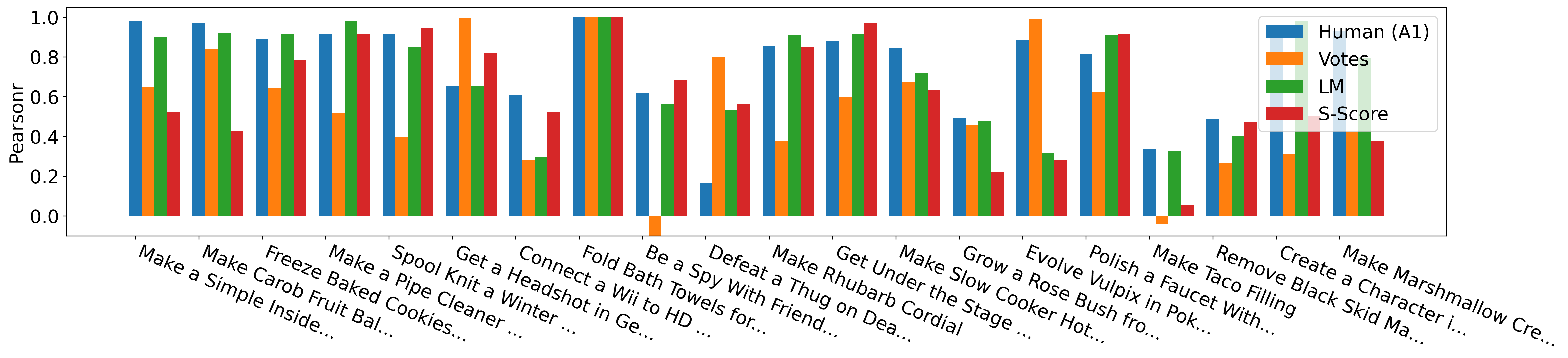}
    \caption{Per-procedure correlation of global entity salience between each set of annotations and the ground-truth human annotations.}
    \label{fig:salience_correlation}
\end{figure*}

\paragraph{Human Labeling} To first procure ground-truth salience labels, two authors (referred to as A1 and A2) annotated entity salience in the first 20 procedures in the development set as the gold standard of entity salience. We devise and follow these annotation criteria in a Likert scale:
\begin{enumerate}[topsep=0pt,itemsep=-1ex,partopsep=1ex,parsep=1ex]
    \item [5:] without or without mentioning this entity, the procedure or step cannot be done at all (e.g., \textit{lemon} in “Wash faucet with lemon”)
    \item [4:] without this entity, another entity of the same type can be used as a replacement, perhaps with worse outcome or more efforts (e.g., pan in “Sear a salmon” - can also use \textit{grill})
    \item [3:] without this entity, the procedure or step can be done in principal, though with slightly worse outcome or more efforts (e.g., \textit{glove} in “Cut off tough branches of a bonsai plant”)
    \item [2:] without this entity, the procedure or step can be done, though with negligibly worse outcome or more efforts (e.g., \textit{vacuum cleaner} in “Drill holes in the wall”)
    \item [1:] the entity appears in the procedure or step rather gratuitously, and the lack thereof makes no difference
    \item [0:] the entity is irrelevant to the procedure or step
\end{enumerate}

Subjectivity is inevitable even though we strive to minimize subjectivity using this fine-grained scale to capture nuanced situations (e.g., an entity that frequently appears that can be easily replaced versus one that appears only once but is irreplaceable). In later sections, we will see how this scale leads to reasonable inter-annotator agreement and favorable performance on downstream tasks.

\paragraph{LM Prediction} We prompt \texttt{gpt-3.5-turbo}, as before, to automatically predict salience. Table~\ref{tab:salience_prompt} shows an example prompt for predicting global salience. As before, we use the default hyperparameters with a temperature of 0.7. We parse the result by extracting the first digit from the generation as the score, and default to 1 whenever impossible.

\paragraph{Evaluation} \label{sec:salience_eval}

To first holistically evaluate the modelling of salience, we report pairwise Pearson's correlation coefficients between each set of labels above and the annotations of human A1. In Table~\ref{tab:correlation}, we report a ``macro correlation'', namely the mean of correlation of salience scores in each procedure.\footnote{To avoid NaN due to constant input array, a 0 is appended to each array as smoothing.} First, the correlation between the two annotators is high but imperfect, implying  subjectivity in the annotation of entity salience. In comparison, the LM predictions come close with especially impressive predictions for global salience. 

To understand when and how entity salience can be subjective among humans, in Figure~\ref{fig:salience_correlation} we show salience correlation for the first 20 procedures. Some with low inter-human correlation such as ``Defeat a Thug'' expose a \textit{document-frequency} problem: one human labels the entity ``you (actor)'' with a salience of 5, believing that without the actor the procedure cannot be undertaken at all, while the other labels 1, believing that the actor occurs in every procedure and thus the mention of it makes no difference. Other procedures such as ``Make Taco Filling'' expose a \textit{cluster-individual} problem: all individual ingredients are grouped under one cluster, for which annotators are divergent on the salience. While such issues can be mitigated by better defining entity salience in a potential future annotation process, it is much more practical to consider downstream applications.


\begin{table*}
\small
\centering
    \begin{adjustbox}{max width=\textwidth} 
    \begin{tabular}{ll}
        \toprule
        Role & Content \\
        \midrule
        User & \makecell[l]{A person walks into a dark room in a \textbf{freezing windy night}. In the very beginning before anything is done, \\how likely is it that she can see what's in the room? Answer on a scale of 1-5, with a 1-sentence explanation.}  \\[5pt]
        Assistant & 1 - it is highly unlikely for the person to see anything in a dark room without any source of light. \\[5pt]
        User & \makecell[l]{Next, she lights up a torch and hangs it near a window. At this point, how likely (same question)...} \\[5pt]
        Assistant & \makecell[l]{5 - hanging the torch near a window maximizes the distribution of light in the room, greatly enhancing the \\visibility and increasing the likelihood of seeing what's inside.} \\[5pt]
        \textsuperscript{1}User & \makecell[l]{Next, she \textbf{opens the window}. At this point, how likely (same question)...} \\[5pt]
        \textsuperscript{1}Assistant & \makecell[l]{\textbf{5} - opening the window allows \textbf{natural light} to enter the room, combining with the torchlight to provide \\ample illumination, greatly increasing the likelihood of seeing what's in the room.} \\[5pt] \midrule
        \textsuperscript{2}User & \makecell[l]{Next, she opens the window. The most salient entities for this step are the window, the wind, and the flames.\\How does each of these entities change?} \\[5pt]
        \textsuperscript{2}Assistant & \makecell[l]{When she opens the window, the following changes occur:\\Window: The \textbf{window is now open}, allowing for air and potentially some outside light to enter the room.\\Wind: The wind becomes more noticeable and \textbf{enters the room through the open window}, causing a draft\\and potentially affecting the flames of the torch.\\Flames: The flames of the torch may flicker or change direction due to the entry of wind through the open\\window, \textbf{altering their shape and intensity}.} \\[5pt]
        \textsuperscript{2}User & \makecell[l]{With such information, at this point, how likely is it that she can see what's in the room?} \\[5pt]
        \textsuperscript{2}Assistant & \makecell[l]{\textbf{3} - opening the window allows some outside light to enter, the presence of wind and potential \textbf{disruption to} \\\textbf{the flames} may still impede visibility to some extent.}
        \\ \bottomrule
    \end{tabular}
    \end{adjustbox}
    \caption{\texttt{gpt-3.5-turbo}'s performance on a CREPE-like example. The interactions with and without using entity salience are both shown. Critical information is illustrated in bold.}
\label{tab:crepelike_prompt}
\end{table*}

Using our methods above and adding the model predictions for canonicalization and salience to the \openpi dataset, we now fully present \openpitwo. As the procedures and entity state annotations have not changed, \openpitwo still has 55 procedures with 5.0 steps on average. These procedures are collected from wikiHow and their topics are everyday activities. \openpitwo also inherits the original entity-attribute-state changes annotated by crowd workers. After canonicalization, there are 356 canon entities each with 7.6 unique mentions and 5.5 expanded mentions on average, 3240 canon attributes, each with 3.0 unique mentions and 3.3 expanded mentions on average, and 1193 before-after states in the development set. The quality of clustering and expansion and be evidenced in \S\ref{sec:canonicalization}. Regarding salience, the global salience of entities has a mean of 3.5 and standard deviation of 1.4; the local salience of entities has a mean of 3.4 and standard deviation of 1.5.

Regarding the training set, \openpitwo is no different than \openpi, for one may either fine-tune an LLM to predict entity states (i.e., produce the entity schema, my proposed structured representation of a procedure) for LLMs with the pre-train then fine-tune paradigm. However, to take advantage of the powerful LLMs with the in-context learning paradigm, \openpitwo allows for effective \textit{model selection}, which entails not only selecting an LLM and its hyperparameters, but also prompt tuning. In other words, \openpitwo leads to LLMs that can best predict the entity schema of events. I will continue to discuss how that entity schema can effectively work with LLMs to outperform end-to-end usage in some challenging event reasoning tasks.

The work above was published in \citet{zhang2023openpi2}, in which I primarily contributed to all components except the canonicalization part to which Hainiu Xu primarily contributed. I have obtained approval from all collaborators to exclusively include this work in this thesis. 

\section{Application of Entity Schema}
\label{sec:crepe}

We now have LLMs, evaluated on \openpitwo, that can effectively predict entity states. Namely, an entity schema can be derived from procedural texts. We are now ready to tackle the task of causal event reasoning, exemplified in the beginning of this Chapter. We will see how a structured representation of said entity schema contributes to model performance.

Before diving into the application, I will first clarify how exactly the entity schema can interface with or be fed into LLMs. Recall that in Chapter~\ref{chap:relation}, the event-relation schema interfaces with LLMs via fine-tuning. This is possible because all the events are natural language sentences and the relations among events are learned via labeled data. Regarding the entity schema, one can similarly fine-tune LLMs to predict \textit{some part} of the schema, such as the pairwise relation between an event and an entity. However, I try to be more ambitious and attempt to feed the \textit{entire} entity schema as input to LLMs to facilitate their reasoning. In this past, this would not have been possible because LLMs were pre-trained only on natural language, unable to work with a structured representation such as a matrix, graph, or some data structure. Fortunately, at the time that work described in this Chapter took place, latest LLMs are also trained on structured data such as code, in addition to text. Therefore, these LLMs can not only take structured representations as input but also predict them. Along with this advance is the in-context learning paradigm with many advantages over the pre-train then fine-tune paradigm (Section~\ref{sec:llm_history}). Most noticeably, it reduces the need for manually labeled data for LLMs to make predictions. In summary, it is now possible for the the cutting-edge LLMs to not only predict by consume an entire entity schema. Even so, how exactly one should represent the schema (i.e., the form; e.g., as a matrix, graph, or Python code) is still a key design choice.

Unlike the form discussed above, the term \textit{structure} (e.g., structured reasoning, event structure, etc.) has a much deeper semantic meaning, referring to the underlying representation of an event. In summary, an unstructured representation of an event is necessarily in the form of natural language, whereas a structured representation can either take the form of natural language or symbolic language.

\subsection{Motivation}
The earlier example of whether ``\textit{one can safely touch the pan}'' is different from typical QA examples like ``\textit{how many states are there in the US}'' in that the former question is grounded to a particular environment. In other words, the answer depends on the exact configuration the environment, which in turn is decided by the events that have previous taken place. While the environment can be specified in many possible ways, procedural text, as we have extensively discussed in Chapter~\ref{chap:relation}, is a common descriptor of an environment that changes dynamically through a sequence of steps. Even so, the exact environment configuration is often implicit (e.g., we know that ``\textit{we boil the water}'', but we are not explicitly told that ``\textit{the water is hot}.'' Such a QA task is an instance of event reasoning - or, specifically, procedure reasoning - that this thesis focuses on. With these interesting challenges coupled with the added benefit of application to robotics \citep{brohan2023can} and household smart assistants such as Alexa \citep{Pennsylvania2022}, reasoning about procedures attracts great attention from the NLP community.

Most work on reasoning about procedural texts has focused solely on either predicting the properties of events (e.g., which event is more likely to happen) \citep{zhang-etal-2020-reasoning, yang-etal-2021-visual, tandon-etal-2019-wiqa} or tracking entity states (e.g., what is some property of an entity after some step) \citep{dalvi-etal-2018-tracking,tandon-etal-2020-dataset}, while the causal relation between events and entities is largely underexplored -- for example, whether ``\textit{there is a sizzling sound}'' is determined by the state of ``\textit{water}'' and ``\textit{oil}.'' Therefore, we claim that many event prediction tasks are multihop reasoning tasks that require the knowledge of intermediate entity states. Causal reasoning about events and entities differs from existing multihop reasoning tasks, such as \citet{yang-etal-2018-hotpotqa,dua-etal-2019-drop} whose reasoning process is explicitly formulated by a direct question (e.g., \textit{how old is the previous US president}); and \citet{geva-etal-2021-aristotle} whose supporting evidence is factual and static. In contrast, causal reasoning in procedures requires models to first figure out the relevant entity attributes, then infer their states based on the current context, and finally predict the event.

\begin{figure}
    \centering
    \includegraphics[width=0.6\columnwidth]{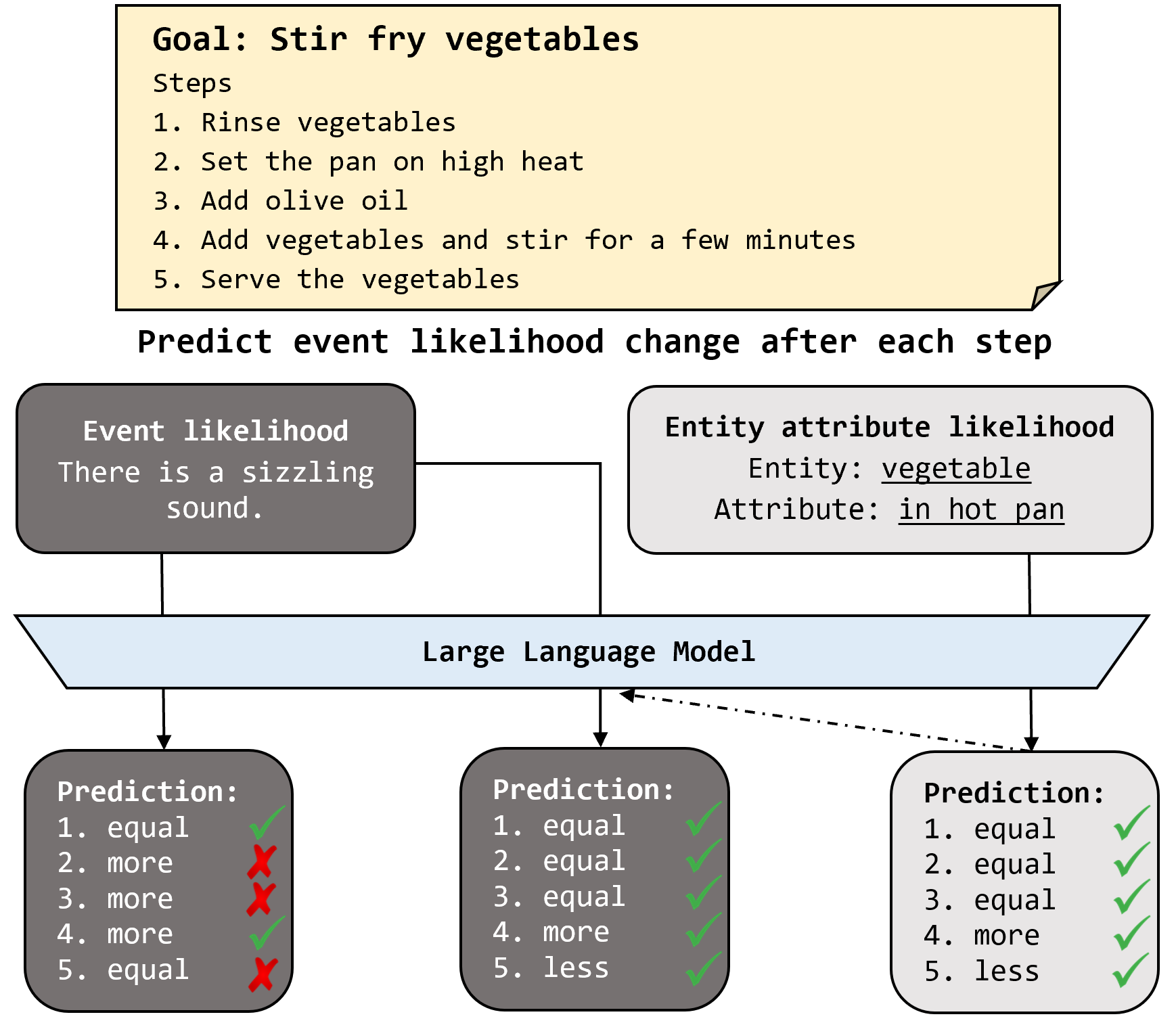}
    \caption{Example of our task CREPE. A procedure including a goal and some steps are provided. A model needs to predict the change in the likelihood of an event throughout the procedure. We show that predicting entity states as an intermediate step improves performance.}
    \label{fig:crepe}
\end{figure}

To this end, we propose the task of \textbf{C}ausal \textbf{R}easoning of \textbf{E}ntities and \textbf{E}vents in \textbf{P}rocedural Texts (\textbf{CREPE}), with an overview in Figure~\ref{fig:crepe}. Given a procedure consisting of a goal~(``\textit{stir fry vegetables}'') and some steps~(``\textit{rinse vegetable}''...), a model is to predict the likelihood of some unobserved events~(``\textit{there is a sizzling sound}'') after the execution of each step. This kind of hypothetical, counterfactual event reasoning is a high-level cognitive ability beyond pattern recognition and a manifestation of complex reasoning ability \citep{10.5555/3238230, pearl2019seven}. Counterfactual reasoning has a long history with formal methods \citep{forbus1984qualitative,lewis2013counterfactuals}. Less modern work exists in commonsense \citep{feng-etal-2021-empowering}, procedural texts \citep{tandon-etal-2019-wiqa}, and even computer vision \citep{yue2021counterfactual}. 

\subsection{Task and Hypothesis}
\label{sec:task}
I will first formally describe the task of CREPE. A procedure $P$ of length $n$ consists of a goal $G$ and some steps $s_1\dots s_n \in S$, each represented as a short sentence. Each procedure is associated with a set of hypothetical events $e_1\dots e_m \in E$ whose likelihood of happening changes throughout the procedure. The task is to predict the change of likelihood of a hypothetical event $e_j$ from step $s_{i-1}$ (the previous step) to step $s_i$ (the current step):
\[\delta_i = p\left(e_j|s_i,\dots,s_1, G\right) - p\left(e_j|s_{i-1},\dots,s_1, G\right)\]
The likelihood change $\delta_i$ is positive if the label is ``more likely'', negative if ``less likely'', or zero if ``equally likely''. 

In our work, we hypothesize that \textbf{the causal relation between entity changes and events} can be leveraged by LLMs to better perform counterfactual reasoning. In other words, any change of the likelihood of a hypothetical event is given rise to by changes of some entity attributes $a_1\dots a_m \in A$.
\[\delta_i = p(a_j|s_i,\dots,s_1, G) - p(a_j|s_{i-1},\dots,s_1, G)\]

\subsection{Dataset}
\label{sec:dataset}

Our CREPE benchmark dataset has two portions. The first is handcrafted and cross-validated by six authors of this paper. The annotation happens in 3 phases: (1) we first write down or acquire a procedure from the web; (2) we then annotate some hypothetical events whose likelihood of happening changes throughout the procedure, and how their likelihood change after each step; (3) for each event, we annotate a tuple of entity, attribute, and change that causes the event likelihood change. To obtain interesting and challenging data, we require annotators to write procedures covering a diverse range of topics and to prioritize events that undergo multiple likelihood changes, and those that involve information implicit from the steps. In our work, we strictly use this portion as the development set to inform all our experimental designs.

The second portion, designed to be drawn from a different distribution to minimize bias, was annotated by students in an Artificial Intelligence class at the University of Pennsylvania who participated in an extra-credit assignment. The students were given an overview of the project and some guidelines to annotate data with the aforementioned criteria. We carefully validated all resulting annotations by discarding or editing erroneous and inappropriate examples. In our work, we strictly use this portion as the test set to evaluate the generalization ability of our final models. The complete dataset and annotation instructions can be found in our public repository containing no personally identifiable information of any annotator. 

\begin{table}
    \centering
    \begin{tabular}{llll}
    \toprule
    \multicolumn{4}{c}{\textbf{Data Statistics}} \\
    \midrule
                                                      & Dev & Test & Total \\ \midrule
    Num. procedures                                   & 42    &  141    &  183     \\
    Num. steps                                        & 295    &  924    & 1219      \\
    Num. event changes & 144    & 180     &  324     \\
    Avg. step per procedure & 7.0 & 6.6  & 6.7 \\ 
    Avg. token per step & 6.8 & 6.8 & 6.8 \\
    \midrule
    \midrule
    \multicolumn{4}{c}{\textbf{Procedure Topics}} \\
    \midrule
                                                      & Dev & Test & Total \\ \midrule
    Recipe                                   & 10    &  33    &  43     \\
    Household                                        & 12    &  40    & 52      \\
    Craft & 4    & 17     &  21     \\
    Technology & 5 & 19  & 24 \\ 
    Travel & 4 & 4 & 8 \\
    Sports & 2 & 13 & 15 \\
    Others & 5 & 15 & 20 \\
    \bottomrule
    \end{tabular}
    \caption{Statistics of the CREPE dataset.}
    \label{tab:dataset_stats}
\end{table}

The statistics of CREPE are in Table~\ref{tab:dataset_stats}. In this work, we consciously focus on few-shot and in-context settings because our data annotation inevitably contains bias and limitation, and thus cannot be truly representative of counterfactual reasoning in every scenario. In such cases, we believe having a sizeable training set aggravates such biases and induces spurious artifacts.

The task of CREPE is essentially ternary classification, where the likelihood change of each event after each step is labeled as one of “more likely”, “less likely”, or “equally likely”. Here, all models have no access to the annotated entity state changes until later sections.

\subsection{Text Form}
To show the challenge CREPE brings to existing models, we first introduce some naive baselines.
\begin{itemize}[leftmargin=*]
    \item The \textbf{chance} baseline assigns random labels.
    \item The \textbf{majority} baseline always assigns the majority label “equally likely”.
\end{itemize}

\begin{figure}
    \centering
    \begin{center}
    \begin{tcolorbox} [top=2pt,bottom=2pt, width=\linewidth, boxrule=1pt]
    {\small {\fontfamily{zi4}\selectfont
    Goal: Wash sneakers\\
    Context: I remove shoelaces. I rinse.\\
    Question: What is the likelihood that my feet get wet by wearing the sneakers?\\
    Answer: likely
    }
    \par}
    \end{tcolorbox}
    \end{center}
    \caption{Our GPT-3 prompt, which is typical for a QA task. Each likelihood label is compared with the previous one to get the label for the change.}
    \label{fig:GPT-3_prompt}
\end{figure}

Next, we consider the following state-of-the-art LLMs as strong baselines, where all models are given exactly three examples in their prompt:

\begin{itemize}[leftmargin=*]
    \item \textbf{T5} \citep{raffel2020exploring} is one of the state-of-the-art LLMs. Given the goal, steps, and question formatted by a prompt template, we compare the probability of generating ``the answer is no$|$yes.'' We use \texttt{T0-3B}\footnote{\url{https://huggingface.co/t5-3b}} with 3 billion parameters.
    \item \textbf{T0} \citep{sanh2022multitask} is a variant of T5, fine-tuned on a large set of downstream tasks with natural language prompts. We adopt the same inference process as T5 described above. We use \texttt{T0pp}\footnote{\url{https://huggingface.co/bigscience/T0pp}} with 11 billion parameters.
    \item \textbf{GPT-3} \citep{brown2020language} is a series of LLMs that excels at few-shot learning using the prompting mechanism. We consider \texttt{text-curie-001} (7B parameters), \texttt{text-davinci-002}, \texttt{text-davinci-003}, and \texttt{ChatGPT} (all 175B parameters). We use default parameters with a temperature of 0 for deterministic predictions. An example of the prompt is shown in Figure~\ref{fig:GPT-3_prompt}.
    \item \textbf{GPT-3 finetuned on StrategyQA} is a GPT-3 \texttt{curie} model finetuned with StrategyQA \citep{geva-etal-2021-aristotle}, a dataset of factual multihop questions and their decomposition. StrategyQA is similar to our task in that estimating the change of event likelihood can also be decomposed into sub-tasks of estimating the change of state of related entities (Section~\ref{sec:injecting_entities}). 
\end{itemize}

\begin{table*}[!t]
\centering
    \small
    \begin{adjustbox}{max width=\textwidth}
\begin{tabular}{l|ll|llllllll|l}
\toprule
   & \multicolumn{2}{c|}{\textbf{Naive}} & \multicolumn{8}{c|}{\textbf{Large Language Models}}    & \textbf{Human}  \\ 
   & Cha.      & Maj.     & T5 & T0 & GPT3C & GPT3C+S &GPT3D2 &GPT3D3 &ChatGPT & \makecell{Codex\\(ours)} &  \\
Params & -          & -           & 3B & 11B & 13B & 13B   &175B &175B  & 175B  & 175B & -            \\ \midrule
Dev  & .262         & .297        & .343 & .336 & .346 & .341 & .350& .424 & .470 & \textbf{.585} & .868 \\ 
Test & .251         & .296       & .343 & .337 & .356 & .346& .533  & .423 & .462  & \textbf{.591} & -  \\ \bottomrule
\end{tabular}
\end{adjustbox}
\caption{Macro F1 of baseline models on the CREPE dataset. Human performance is not benchmarked on the test set as we strictly hold out its labels during all experiments. GPT3C represents the \texttt{text-curie-001} model. GPT3D2 represents the \texttt{text-davinci-002} model with an abnormal performance on the test set that we have confirmed but regrettably cannot explain. GPT3D3 represents the \texttt{text-davinci-003} model. GPT3C+S represents the GPT-3 \texttt{curie} model finetuned on StrategyQA. All of the above models work with textual prompts. Codex represents the \texttt{code-davinci-002} model and works with our proposed code-like prompts.}
\label{table:baseline}
\end{table*}

Table~\ref{table:baseline} shows that all state-of-the-art LLMs we have attempted achieve close-to-chance performance on CREPE around 0.350 F1, whereas \texttt{text-davinci-003} and \texttt{ChatGPT} which are known to be stronger at reasoning perform only slightly better. These results showcase that the CREPE task is clearly challenging for even the strongest LLMs when used in an end-to-end manner. Previously, we have established that LLMs can now work with both a natural language form and a symbolic language form of the data. While the above representation is of course an ordinary \textbf{text form}, we will next explore a symbolic form of the examples in CREPE that will later be essential for the application of the entity schema.

\subsection{Code Form}
\label{section:code_prompt}

\textbf{Codex} \citep{chen2021evaluating} is a variation of GPT-3 that was designed to be prompted with and to generate code, in addition to natural language texts. Shortly before the publication of this work, \citet{madaan-etal-2022-language} found that prompting Codex with some structured representation such as Python code. Inspired by this observation, we propose novel code representations of procedures and hypothetical events. Among many possibilities we experimented with, the representation with the best empirical performance is described below, later shown to greatly outperform all baseline models. The representation is exemplified in Figure~\ref{fig:codex_prompt}.

The procedure is represented as a class where the goal $G$ is the class name, followed by the steps $s_i$ as comments. Then, each step is defined as a member function, in which the hypothetical events $e_j$ are represented as objects with comments. Each event object has an attribute ``change'' whose value describes the change of the likelihood. During inference, Codex is provided with the prompt including three in-context examples and the current procedure up to the definition of the ``init'' function and predicts the definition of all step functions. Finally, we extract the assigned value of the ``change'' attribute as the event likelihood change $\delta_i$.

This prompt design effectively leverages the semantic similarity between procedures with entity states and functions with variables, by representing texts as function identifiers and comments. We use \texttt{code-davinci-002}\footnote{While OpenAI announced that \texttt{text-davinci-002} is based on \texttt{code-davinci-002} (\url{https://platform.openai.com/docs/model-index-for-researchers}), we empirically find the former to perform worse with our code prompt and thus only consider the latter with code prompt.} with 175B parameters and default hyperparameters with a temperature of 0.

\begin{figure}[!t]
    \centering
    \begin{tcolorbox} [top=2pt,bottom=2pt, width=\linewidth, boxrule=1pt]
\begin{tabular}{rp{15cm}}
\begin{lstlisting}[numbers=none, basicstyle=\fontfamily{cmtt}\small,style=pythoncode,belowskip=-\baselineskip,aboveskip=- 0.5\baselineskip,commentstyle=\color{codegreen}]
class Wash_Sneakers:
  # Init
  # Remove shoelaces
  # Rinse
  def __init__(self, event0):
    self.event0 = event0 # My feet get wet by wearing the sneakers.
  def remove_shoelaces(self):
    self.event0.change = "equally likely" # My feet get wet by wearing the sneakers.
  def rinse(self):
    self.event0.change = "more likely" # My feet get wet by wearing the sneakers.
\end{lstlisting}
\end{tabular}
\end{tcolorbox}
    \caption{Our best-performing Python code representation of a procedure and hypothetical events, for Codex.}
    \label{fig:codex_prompt}
\end{figure}

\paragraph{Results}
As CREPE is a ternary classification task, we report the macro F1 score across the three classes. As shown in Table~\ref{table:baseline}, T5 and T0 perform only slightly better (.343 and .336 F1) than chance (.297 F1). GPT-3, one of the most dominant models across a variety of NLP tasks, is no better (.336 F1), whereas finetuning it on another multihop reasoning dataset StrategyQA does not bring about any improvement (.341 F1). The latest GPT-3 models, \texttt{text-davinci-003} (.424 F1) and \texttt{ChatGPT} (.470 F1) which were released contemporarily with this paper, greatly outperform their predecessors. 

On the other hand, our code-representation of events as the prompt to Codex greatly outperforms all other models with .585 F1. As Codex is trained on public Github code in addition to the internet texts that GPT-3 is trained on, it is noteworthy that Codex can effectively reason about texts with code-like structures, for a procedure has many analogies to a class in object-oriented programming. 

\begin{table}[t!]
\centering
\begin{tabular}{lll}
\toprule
                  & \textbf{Dev} & \textbf{Test} \\ \midrule
Codex             &  \textbf{.585}   &  \textbf{.591}    \\
no step comments    & .377    &   .352   \\
no event comments   &  .576  &   .555   \\
nested function   &  .568   &   .572   \\
flat variables &   .338  &   .341   \\ \bottomrule
\end{tabular}
\caption{Macro F1 of the ablations of our Codex prompt.}
\label{table:ablation}
\end{table}

\paragraph{Ablation Studies}
To understand why the representation in our Codex prompt is effective, we perform an ablation study with various changes of the format to the representation, including:
\begin{itemize}[leftmargin=*]
    \item Remove steps comments in the beginning
    \item Remove event comments in step functions
    \item Use nested functions instead of a class
    \item Use flat variables to encode goals, steps, and events (no hierarchical class functions)
\end{itemize}
As seen in Table~\ref{table:ablation}, the hierarchical representation of procedures, steps, and events as classes or nested functions is critical. Besides, listing all the steps as comments helps, mimicking a programmer's textual explanation of a class or a function. 

\subsection{Injecting Entities}
\label{sec:injecting_entities}

When a human tries to predict whether the event ``\textit{one would get burnt by touching a pan}'' is likely, their reasoning process would first focus on some entities in the question~(e.g., ``the pan''), then attend to some attributes and states of that entity~(e.g., the temperature of the pan is hot), and finally draw a logical conclusion~(e.g., ``the pan being hot means one would get burnt by touching it.'') CREPE is constructed precisely with this thought process in mind. An entity-attribute-change tuple is annotated along with each event likelihood change. In this section, we study how to explicitly leverage the intermediate information to assist the prediction of event likelihood prediction.

In CREPE, the task of predicting event likelihood change can be seen as a case of multihop reasoning, where a model first decomposes the question into some open-ended sub-questions, answer these sub-questions, and aggregate them as a final answer. LLMs can be prompted to perform chain-of-thought (CoT) style reasoning \citep{nye2021show,wei2022chain}. Thus, we ask the question:
\begin{displayquote}
\textbf{Q1}. Can LLMs benefit from first \textbf{predicting} entity state changes, as a CoT, before predicting event likelihood changes?
\end{displayquote}

\paragraph{Text Form}
First, we prompt GPT-3 with \citet{wei2022chain}'s CoT paradigm and \citet{press2023measuring}'s self-ask paradigm, both of which are shown in Figure~\ref{fig:GPT-3_prompt_mh}. While self-ask relies on search engines for fact retrieval, we use LM generation instead as most of our entity state tracking questions are heavily context-dependent and unanswerable by any search engine. When writing demonstrations for few-shot learning, we impose the following logic progression for the follow-up questions: (1) initial followups shall ask questions on the state of entities that are directly related to the event; (2) followups following the entity state questions shall ask for the logical relationship between the entity states and the original event.

\begin{figure}[!t]
    \centering
    \begin{center}
    
    \begin{tcolorbox} [top=2pt,bottom=2pt, width=\linewidth, boxrule=1pt]
    {\small {\fontfamily{zi4}\selectfont
    Goal: Wash sneakers\\
    Context: I remove shoelaces. I rinse.\\
    Question: What is the likelihood that my feet get wet by wearing the sneakers?\\
    Answer: To get feet wet by wearing the sneakers, the sneakers must be wet. In the given context, the sneakers are wet. Therefore, comparing to the previous step, the likelihood change is "more likely".
    }
    \par}
    \end{tcolorbox}
    
    \begin{tcolorbox} [top=2pt,bottom=2pt, width=\linewidth, boxrule=1pt]
    {\small {\fontfamily{zi4}\selectfont
    Goal: Wash sneakers\\
    Context: I remove shoelaces. I rinse.\\
    Question: What is the likelihood that my feet get wet by wearing the sneakers?\\
    Follow up: Are the sneakers wet?\\
    Intermediate answer: Yes\\
    Follow up: Will my feet get wet by wearing wet sneakers?\\
    Intermediate answer: Yes \\
    Answer: likely
    }
    \par}
    \end{tcolorbox}    
    \end{center}
    \caption{Our GPT-3 prompt with intermediate questions, mimicking the CoT prompt (top) and the Self-Ask prompt (bottom).}
    \label{fig:GPT-3_prompt_mh}
\end{figure}

\begin{table*}[]
\centering
    \small
    \begin{adjustbox}{max width=\textwidth}
        
\begin{tabular}{c|c|cc|cccc|c}
\toprule
   & \multicolumn{1}{c|}{\textbf{Naive}} & \multicolumn{2}{c|}{\textbf{LLMs}} & \multicolumn{4}{c|}{\textbf{CoT Large Language Models}}    & \textbf{Human}  \\ \midrule
   & Majority     & GPT-3   & Codex     & GPT-3+CoT & GPT-3+self-ask & \makecell{Codex soft\\(ours)} & \makecell{Codex hard\\(ours)}  &  \\
Dev & .297     & .346     & .585        &  0.359 & .342 & .624   & \textbf{.667  }  & .868   \\ 
Test & .296    & .356      & .591       & 0.379    & .345 & \textbf{.626}   & .609   & -  \\\bottomrule
\end{tabular}
    \end{adjustbox}
\caption{Macro F1 of chain-of-thought models on the CREPE dataset. GPT-3 + CoT|self-ask represents the \texttt{text-davinci-002} model prompted with the CoT or self-ask style prompt. }
\label{table:baseline_cot}
\end{table*}

\paragraph{Code Form}
We modify our Codex prompt in Figure~\ref{fig:codex_prompt}, so that a sub-event is represented as a string variable whose declaration and value assignments are right before those of the hypothetical event. We refer to this as a \textit{soft representation} of entities (Figure~\ref{fig:codex_prompt_w_event}). 
During inference, Codex is provided with the code up to the step function header and predicts the entity and event changes for every step function. Our Codex model achieves the new best performance of .624 F1, outperforming the same model without predicted entities as CoT by .039 F1.

\begin{center}
\begin{tcolorbox} [top=2pt,bottom=2pt, width=\linewidth, boxrule=1pt]
\begin{tabular}{rp{15cm}}
\begin{lstlisting}[numbers=none, basicstyle=\fontfamily{cmtt}\small,style=pythoncode,belowskip=-\baselineskip,aboveskip=- 0.5\baselineskip,commentstyle=\color{codegreen}]
class Wash_Sneakers():
  # Init
  # Remove shoelaces
  # Rinse
  def init(self, event0, subevent0):
    self.event0 = event0 # My feet get wet by wearing the sneakers.
    self.event0.subevent = subevent0 # The sneakers are wet
  def remove_shoelaces(self):
    self.event0.subevent.change = 
      "equally likely" # The sneakers are wet
    self.event0.change = "equally likely" # My feet get wet by wearing the sneakers.
  def rinse(self):
    self.event0.subevent.change = 
       "more likely" # The sneakers are wet
    self.event0.change = "more likely" # My feet get wet by wearing the sneakers.
\end{lstlisting}
\end{tabular}
\end{tcolorbox}
\end{center}
\begin{figure}[t!]
    \centering
    \caption{Our Codex prompt with a soft representation of entity state changes as strings.}
    \label{fig:codex_prompt_w_event}
\end{figure}

The two approaches above both \textit{softly} represent the intermediate entity state changes as texts, either questions or statements. Here, LLMs are not enforced to generate intermediate reasoning steps that contain entities and attributes. To answer Q1 more precisely, we experiment with a \textit{hard entity representation} where the entity-attribute-change tuple is explicitly baked into the Codex prompt as shown in Figure~\ref{fig:codex_prompt_w_entity}. Here, each entity is represented as an object with an attribute and assigned value. The hard entity representation leads to a far superior performance of .667 F1 on the development set but generalizes worse on the test set with .609 F1. 

\begin{figure}[t!]
    \centering
    \begin{tcolorbox} [top=2pt,bottom=2pt, width=\linewidth, boxrule=1pt]
\begin{tabular}{rp{15cm}}
\begin{lstlisting}[numbers=none, basicstyle=\fontfamily{cmtt}\small,style=pythoncode,belowskip=-\baselineskip,aboveskip=- 0.5\baselineskip,commentstyle=\color{codegreen}]
class Wash_Sneakers():
  # Init
  # Remove shoelaces
  # Rinse
  def init(self, event0):
    self.sneakers = Sneakers()
    self.event0 = event0 # My feet get wet by wearing the sneakers.
  def remove_shoelaces(self):
    self.event0.change = "equally likely" # My feet get wet by wearing the sneakers.
  def rinse(self):
    self.sneakers.wet = True
    self.event0.change = "more likely" # My feet get wet by wearing the sneakers.
\end{lstlisting}
\end{tabular}
\end{tcolorbox}
    \caption{Our Codex prompt with a hard representation of entity states as variables, attributes, and values.}
    \label{fig:codex_prompt_w_entity}
\end{figure}

To recap, we have shown that LLMs can be prompted to exhibit a CoT that first predicts entity state changes and then event likelihood changes. Hence, our answer to \textbf{Q1} raised at the beginning of this subsection is `yes.'

\begin{table}[t!]
\centering
\begin{tabular}{lll}
\toprule
                  & \textbf{Dev} & \textbf{Test} \\ \midrule
Majority             &  .297   &  .296    \\ \midrule
GPT-3 CoT             &  .342   &  .345    \\
w/ gold entity changes    & .351    &   .380   \\
Codex CoT   &  .667  &   .609   \\
w/ gold entity changes   &  \textbf{.715}   &   \textbf{.722}   \\ \midrule 
Human             &  .868   &  -    \\\bottomrule
\end{tabular}
\caption{Macro F1 of GPT-3 and Codex with chain-of-thought provided with gold entity state changes.}
\label{table:gold_entities}
\end{table}

\paragraph{Annotated Entity States}
In the above section, we have shown how event likelihood prediction can be improved by first having the LLMs predict entity states as a CoT. These experiments mimic a realistic setting where information about entities is unavailable. However, in some scenarios, the entity states may be provided. For example, an embodied agent or a robot might have a reliable component that tracks entities; some practitioners might care about a small set of procedures in a narrow domain with annotated entity changes; or, some event schemata containing entity information could be used to predict unseen events. Here, we try to answer the following question:
\begin{displayquote}
\textbf{Q2}. Can LLMs effectively leverage \textbf{annotated} entity state changes to better predict event likelihood changes?
\end{displayquote}

Instead of having LLMs predict entity state changes, we provide the annotated entity state changes in the CREPE dataset to GPT-3 and Codex. Doing so has the additional benefit of verifying that entity state changes indeed causally benefit LLMs in predicting events.

As shown in Table~\ref{table:gold_entities}, our Codex representation with access to gold entity changes leads to improved performance of .715 F1 on the development set. In contrast, GPT-3 does not see any gain. Hence, the answer to \textbf{Q2} is `yes' for the code-trained LLMs but `no' for standard LLMs.

\paragraph{Predicted Entity States}

As we will discuss further in Section~\ref{sec:related_work}, entity state tracking is an established task in NLP with existing datasets and models. We have now predicted entity state changes using LLMs in a few-shot learning setting. It is then natural to pose the question:
\begin{displayquote}
\textbf{Q3}. Do existing entity state tracking models make predictions that lead to better performance on CREPE?
\end{displayquote}
Our definition of causal reasoning of events is directional since   we consider entity state changes as the cause of the change in event likelihoods. To this extent, we incorporate OpenPI \citep{tandon-etal-2020-dataset}, the only open-domain entity state tracking dataset in procedural texts, as a part of the pipeline. In OpenPI, the input is a goal, a step, and the output is tuples of an entity, a feature, and two attributes before and after the execution of the step. For example, after ``heat the pan [step]'', ``the temperature [feature] of the pan [entity] is cool [attribute] before and hot [attribute] afterward.'' While the original paper proposed a GPT2 model \citep{radford2019language}, we opt to finetune the superior GPT-3 Curie model on its data. After the model makes a prediction, we post-process it into the format of CREPE by discarding the feature and producing two entity-attribute-change pairs (e.g., pan-hot-``more likely'' and pan-cold-``less likely''). We provide Codex with only the entity changes when the entity is mentioned in the event. Further, to fit our prompt in the context window of Codex, we provide Codex with 5 entity state changes uniformly drawn from a pool of candidate choices at every step. The resulting OpenPI-prompted Codex gives a degraded macro F1 score of 0.553 on the development set and 0.496 on the testing set. Hence, our answer to \textbf{Q3} is `no,' suggesting that existing entity state tracking datasets may be insufficient for our causal reasoning task.

\subsection{Analysis}
In this section, we analyze potential factors that play a role in our Codex model's performance. We investigate three factors: (1) the number of steps in a procedure; (2) explicit mentions of event-related entity-of-interest (EoI) in a given step; and (3) the logical relation (entailment or contradiction) between the event likelihood change and its related entity state change. To study factor (1), we dichotomize procedures from the development set by the average length of the procedure. To investigate factors (2) and (3), we manually labeled the ground truth EoI mentioning and logical relation for the development dataset. Intuitively, estimating event likelihood in lengthy procedures and in steps where EoI is not explicitly mentioned would be difficult. Rather surprisingly, Codex shows no significant performance discrepancy under factors (2) and (3), and only a slight performance difference in factor (1). 

Further, the task of CREPE can be divided into two sub-tasks, first to identify whether an event likelihood change occurred at all, and then to classify the change as either more or less likely. 
We observe that CoT Codex outperforms Codex on both sub-tasks. For the classification task, in particular, CoT Codex obtained a .149 increase in macro F1 score from .805 to .954. This shows not only that CoT Codex is effective, but also that its bottleneck is identifying event likelihood change.

In summary, we present CREPE, a benchmark for causal reasoning about events and entities in procedural texts. After establishing that end-to-end LLMs such as GPT-3 perform close to chance, we discussed two means of improvement, both critical to this thesis. First, we show that a code-like representation of the data can be fed as input to LLMs and greatly improves the performance. Hinging on this exciting finding, Section~\ref{sec:symbolic_form} explores whether this symbolic form works across other NLP tasks. Second, we show that the entity schema contributes to the success in CREPE, for the information about entities clearly helps reason about the counterfactual events. 

The work above was published in \citet{zhang-etal-2023-causal}, in which I primarily contributed to all components. I have obtained approval from all collaborators to exclusively include this work in this thesis. 

\section{The Versatility of the Code Form}
\label{sec:symbolic_form}

\begin{figure}
    \centering
    \includegraphics[width=0.7\columnwidth]{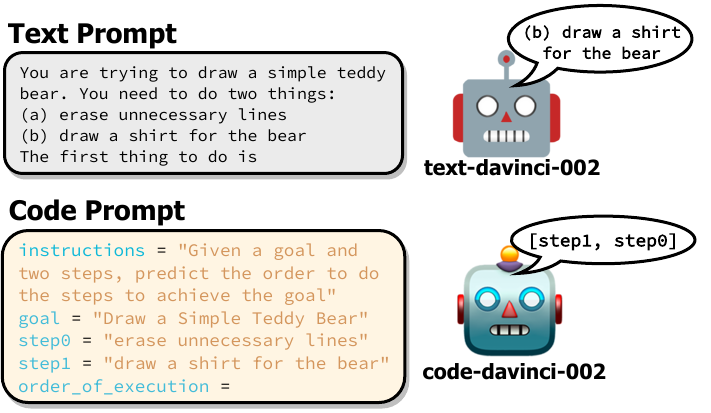}
    \caption{For certain tasks, prompting program-trained language models with code\textit{-like} representations works better than prompting with text.}
    \label{fig:page1}
\end{figure}

\subsection{Motivation}

Any work that attempts to have LLMs work with a structured representation must decide the \textit{form} that the representation takes (e.g., plain text, matrix, graph, Python code, etc.). Apart from our work above \citep{zhang-etal-2023-causal}, several concurrent work has found that prompting such LLMs with a code form (e.g., Python, JSON, PDDL) instead of text leads to performance improvements on structured common sense reasoning \citep{madaan-etal-2022-language}, event argument extraction \citep{wang-etal-2023-code4struct}, knowledge graph construction \citep{bi2023codekgc}, and story understanding \citep{dong2022corrpus}. Such results naturally lead us to ask whether prompting with the code form is the preferred way of interacting with code-trained LLMs \textit{in general}. While previous work is limited to reasoning tasks, in this work we analyze a broad selection of tasks (e.g., QA, sentiment, summarization) and systematically compare the performance of prompting LLMs with code vs. prompting with text\footnote{The code, prompts, and outputs for our experiments are public at \url{github.com/zharry29/curious_code_prompts}}.
We find that:
\begin{itemize}[noitemsep,nolistsep]
    \item With the exception of some reasoning tasks, code prompts do not outperform text prompts
    \item The style of code prompt has a large effect on performance for some but not all tasks.
    \item Fine-tuning on text instructions leads to relative improvements when using code prompts.
\end{itemize}

\subsection{Experimental Design}
\label{sec:experimental_design}
\paragraph{Model Selection}
For our text-based LM we use the original 175 billion parameter \texttt{davinci} model introduced by \citet{brown2020language}. For our LLM we use the newer \texttt{code-davinci-002} model which was explicitly trained on text and code. Neither model underwent any supervised instruction fine-tuning. In addition, we analyze performance on \texttt{text-davinci-002}, which is a variant of \texttt{code-davinci-002} trained explicitly on human demonstrations using supervised fine-tuning\footnote{\url{https://platform.openai.com/docs/model-index-for-researchers}}. We include this model to help us determine whether or not fine-tuning LLMs on text instructions affects their ability to interpret code prompts. All three models were queried through the OpenAI API\footnote{\url{https://openai.com/blog/openai-api}} and our experiments cost approximately \$2700 in total.

\begin{table*}[ht]
    \small
    \centering
    \begin{adjustbox}{max width=\textwidth}
    \begin{tabular}{l|l|l|l|l}
    \toprule
    Dataset & Task Category & Num. Eval Examples & Metric & Origin \\
    \midrule
    HellaSwag &Commonsense Reasoning&1000 / 10042&Accuracy&\citet{zellers-etal-2019-hellaswag}  \\
    wikiHow Goal-Step &Commonsense Reasoning&1000 / 1073&Accuracy&\citet{zhang-etal-2020-reasoning}  \\
    wikiHow Temporal &Commonsense Reasoning&1000 / 3100&Accuracy&\citet{zhang-etal-2020-reasoning} \\
    WinoGrande &Commonsense Reasoning&1000 / 1767&Accuracy&\citet{sakaguchi2021winogrande} \\
    OpenPI &Commonsense Reasoning&111 / 111&ROUGE-F1&\citet{tandon-etal-2020-dataset}\\
    ANLI & Natural Language Inference &1000 / 3000&Accuracy&\citet{nie-etal-2020-adversarial} \\
    Yelp &Sentiment Analysis&1000 / 10000&Pearson's r&\citet{ali-2018-character-level-convolutional} \\
    IMDb &Sentiment Analysis&1000 / 25000 &Accuracy&\citet{maas-etal-2011-learning}\\
    HotpotQA &Question Answering&1000 / 7405&Macro-F1 &\citet{yang-etal-2018-hotpotqa} \\
    SQuAD &Question Answering&1000 / 11873&Macro-F1 &\citet{rajpurkar-etal-2018-know} \\
    CNN/Daily Mail &Summarization&1000 / 13368&ROUGE-2&\citet{nallapati-etal-2016}\\
    XSUM &Summarization&1000 / 11332&ROUGE-2&\citet{narayan-etal-2018-dont}\\
    \bottomrule
    \end{tabular}
    \end{adjustbox}
    \caption{The 12 evaluation tasks. Macro F1 is based on \citet{rajpurkar-etal-2016-squad}. For each task, we randomly sample a fixed set of 1000 examples from its validation or test set for evaluation. For OpenPI we are limited to 111 examples.}
    \label{tab:all-tasks}
\end{table*}

\paragraph{Task Selection}
Following the methodology of \citet{sanhmultitask} we select tasks in a top-down fashion by first choosing the categories of interest (e.g. Question Answering, Sentiment Analysis, Summarization) and then selecting datasets from within those categories. We pay special attention to common sense and causal reasoning tasks as LLMs prompted with code have been shown to perform well on such tasks. The resulting 12 tasks are listed in Table~\ref{tab:all-tasks} and include Commonsense Reasoning, Natural Language Inference, Sentiment Analysis, Question Answering, and Summarization. 

\paragraph{Prompt Formulation}
We collect text prompts for each task using the PromptSource dataset \citep{bach2022promptsource}, a publicly available collection of crowd-sourced prompt templates. For tasks with many prompts, we randomly select one from those provided in the dataset. For a few tasks absent on PromptSource, we write the prompts ourselves.

For our code prompts, we manually write four custom code prompts per task. The code prompt types are as follows, from least to most Pythonic.

\begin{figure*}
    \centering
    \includegraphics[width=\columnwidth]{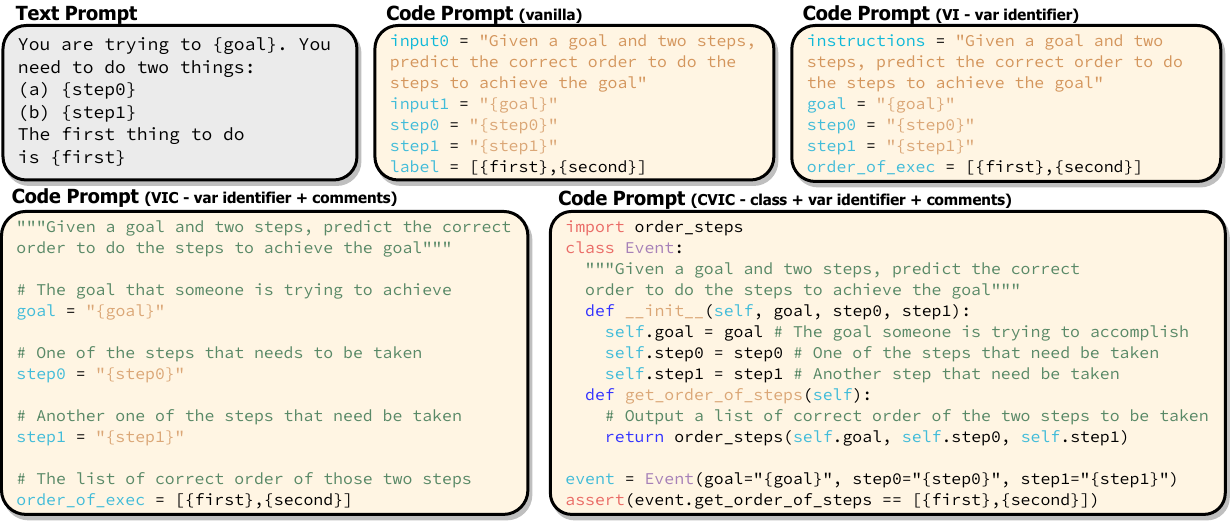}
    \caption{An example of the four styles of manually written code prompts used in our analysis (\texttt{Vanilla}, \texttt{VI}, \texttt{VIC}, and \texttt{CVIC}) for the wikiHow temporal ordering task. At test time, variables in braces are replaced with information from the dataset item (as shown in Figure \ref{fig:page1}). For this task, \texttt{\{goal\}}, \texttt{\{step0\}}, \texttt{\{step1\}} refer to the article title and the steps to order while \texttt{\{first\}} and \texttt{\{second\}} refer to the true ordering of the steps.}
    \label{fig:code-prompts}
\end{figure*}

\renewcommand{\theenumi}{\textbf{(\roman{enumi})}}
\begin{enumerate}[noitemsep,nolistsep]
    \item \textbf{Vanilla (\texttt{Vanilla})}: instructions and inputs are given as variables with generic names;
    \item \textbf{Var Identifier (\texttt{VI})}: instructions and inputs are given as variables with meaningful names;
    \item \textbf{Var Identifier + Comments (\texttt{VIC})}: instructions and inputs are given as variables with meaningful names along with comments explaining their purpose;
    \item \textbf{Class + Var Identifier + Comments (\texttt{CVIC})}: instructions and inputs are given as a task-specific \texttt{class}. Functionality is ``implemented'' as member functions.
\end{enumerate}

Figure \ref{fig:code-prompts} shows an example of the different styles of code prompts for the wikiHow temporal ordering task. Note that we attempt to write our code prompts such that we match the wording of the text-based PromptSource prompt as closely as possible.

At inference time, for each test example, we randomly sample in-context examples from the training set and add them to the context window until the maximum context length is reached. This process circumvents the bias caused by static in-context examples. We conduct an ablation study where we vary the random seed and show that this process produces consistent results. To see whether the findings in our Results section could be attributed to variance in the random sampling of in-context training examples per test example, we conduct five repeated runs using \texttt{code-davinci-002} with different random seeds each time and calculated the standard deviation across the five runs. We report our results in Table~\ref{tab:repeated-run-stats} and find that the choice of in-context examples accounts for very little of the observed variance across prompt type and context length. This finding is surprising as previous work has shown that the selection and ordering of in-context examples has a very large effect on the performance of models \cite{liu2021makes}. However, it seems that our approach of random sampling in-context examples per test item helps to lessen this inherent variance.

\begin{table}
    \small
    \centering
    \begin{tabular}{l|c|c}
    \toprule
        Dataset &Performance&$\sigma$\\
        \midrule
        Hellaswag&0.65, 0.67, 0.69, 0.67, 0.67&$\pm$0.01\\ 
        wikiHow-GS&0.51, 0.51, 0.51, 0.50, 0.51&$\pm$0.00\\ 
        wikiHow-T&0.62, 0.65, 0.63, 0.63, 0.62&$\pm$0.01\\ 
        Yelp&0.92, 0.92, 0.92, 0.92, 0.92&$\pm$0.00\\ 
        IMDb&0.94, 0.94, 0.94, 0.94, 0.94&$\pm$0.00\\ 
        WinoGrande&0.62, 0.64, 0.61, 0.62, 0.62&$\pm$0.01\\ 
        HotpotQA&0.35, 0.33, 0.35, 0.35, 0.35&$\pm$0.01\\ 
        ANLI&0.59, 0.58, 0.57, 0.60, 0.61&$\pm$0.01\\ 
        OpenPI&36.3, 38.1, 38.3, 37.7, 39.9&$\pm$1.16\\ 
        SQuAD&0.60, 0.62, 0.61, 0.60, 0.63&$\pm$0.01\\ 
        CNN/DM&11.7, 12.0, 12.4, 12.3, 12.0&$\pm$0.25\\ 
        XSUM&14.5, 14.9, 15.5, 15.2, 15.4&$\pm$0.36\\ 
        \bottomrule
    \end{tabular}
    \caption{Comparison across 5 repeated runs of the \texttt{code-davinci-002} model with text prompts using different random seeds for sampling in-context examples. We see minimal standard deviation ($\sigma$) between the runs.}
    \label{tab:repeated-run-stats}
\end{table}

\subsection{Results}

\minisection{What is the best type of code prompt?}
We compare performance across the four code prompt types from Section~\ref{sec:experimental_design} on all 12 tasks using \texttt{code-davinci-002} and report our results in Figure~\ref{fig:code_prompts}. We find that no single type of code prompt performs significantly better than the others across all tasks and that the relative difference in performance between code prompts also varies significantly across tasks. For example, on IMDb and SQuAD all code prompts have roughly even performance while for tasks such as wikiHow-Temporal and WinoGrande we see a near 14\% accuracy difference between the worst and best prompt. 

\begin{table}
    \small
    \centering
    \begin{tabular}{c|c|c|c|c}
    \toprule
    &\texttt{Vanilla} & \texttt{VI} & \texttt{VIC} & \texttt{CVIC} \\
    \midrule
    HellaSwag&3&2&1&4 \\
    wikiHow Goal-Step&4&2&1&3 \\
    wikiHow Temporal&4&3&2&1 \\
    Yelp&4&2&1&4 \\
    IMDb&1&3&1&4 \\
    WinoGrande&4&1&2&3 \\
    HotpotQA&4&3&2&1 \\
    ANLI&1&2&4&3 \\ 
    OpenPI&1&2&3&4 \\ 
    SQuAD&1&3&4&2 \\
    CNN/Daily Mail&4&2&3&1 \\
    XSUM&2&4&3&1 \\
    \midrule
    \midrule
    \textit{Mean}& 2.75 & 2.42 & 2.25 & 2.58 \\
    \textit{Standard Deviation}& 1.36 & 0.76 & 1.09 & 1.26 \\
    \bottomrule
    \end{tabular}
    \caption{Relative performance rank of the four code prompt types from Section~\ref{sec:experimental_design} across the 12 tasks. Ranks are calculated based on the results reported in Figure~\ref{fig:code_prompts}. We see that the ``Variable Identifier + Comments'' (\texttt{VIC}) style prompt performs the best out of all code prompt types on average.}
    \label{tab:rank-based-stat}
\end{table}

In Table~\ref{tab:rank-based-stat} we report the rank-based statistics of the four code prompt types from Section~\ref{sec:experimental_design} on our 12 tasks. Ranks are calculated based on the results reported in Figure~\ref{fig:code_prompts} of the main paper. The numbers in a row reflect the relative standing of each code prompt on the corresponding task. While we note that all code prompts perform within $\pm$0.5 ranks of each other on average, we see that on average the \texttt{VIC} prompt performs the best across all tasks and the \texttt{Vanilla} prompt performs the worst. Looking to the standard deviation section, we see that the \texttt{VI} prompt performs the most consistently across all tasks and that once again the \texttt{Vanilla} prompt performs the least consistently.

\begin{figure}[t!]
\centering
\begin{tikzpicture}
    \begin{axis}[ybar,width=5.75cm,height=6cm,enlarge x limits=0.2,
        symbolic x coords={HotpotQA,ANLI,SQuAD,OpenPI},xtick=data,yticklabel style={overlay,font=\tiny},legend style={font=\tiny},bar width=4.5pt,xticklabel style={text width=1.5cm,font=\tiny,align=center,max space between ticks=12,major tick length=0.1cm,},]
        \addplot[purple,fill=purple!30!white]
        coordinates{(HotpotQA,0.377) (ANLI,0.590) (SQuAD,0.598) (OpenPI,0.366)};
        \addplot[olive,fill=olive!30!white]
        coordinates{(HotpotQA,0.434) (ANLI,0.586) (SQuAD,0.581) (OpenPI,0.364)};
        \addplot[red,fill=red!30!white]
        coordinates{(HotpotQA,0.449) (ANLI,0.551) (SQuAD,0.579) (OpenPI,0.364)};
        \addplot[orange,fill=orange!30!white]
        coordinates{(HotpotQA,0.460) (ANLI,0.563) (SQuAD,0.594) (OpenPI,0.358)};    
    \end{axis}
\end{tikzpicture}
\begin{tikzpicture}
    \begin{axis}[ybar,width=3.8cm,height=6cm,enlarge x limits=0.5,symbolic x coords={CNN/DM,XSUM},xtick=data,xticklabel style={text width=1.5cm,font=\tiny,align=center,max space between ticks=12,major tick length=0.125cm,},yticklabel style={overlay,font=\tiny},legend style={font=\tiny},bar width=5pt,]
        \addplot[purple,fill=purple!30!white]
        coordinates{(CNN/DM,10.9) (XSUM,11.2)};
        \addplot[olive,fill=olive!30!white]
        coordinates{(CNN/DM,11.8) (XSUM,11.0)};
        \addplot[red,fill=red!30!white]
        coordinates{(CNN/DM,11.7) (XSUM,11.0)};
        \addplot[orange,fill=orange!30!white]
        coordinates{(CNN/DM,12.3) (XSUM,11.7)};
   \end{axis}
\end{tikzpicture}
\begin{tikzpicture}
   \begin{axis}[ybar,width=8.75cm,height=6cm,enlarge x limits=0.1,symbolic x coords={HellaSWAG,wikiHow goal-step,wikiHow temporal,WinoGrande,Yelp,IMDb},xtick=data,xticklabel style={text width=1.5cm,font=\tiny,align=center,max space between ticks=8,major tick length=0.1cm,ytick placement tolerance=-0.2},yticklabel style={overlay,font=\tiny,},legend style={font=\tiny},bar width=5pt,]
        \addplot[purple,fill=purple!30!white]
        coordinates{(HellaSWAG,0.495) (wikiHow goal-step,0.709) (wikiHow temporal,0.630) (WinoGrande,0.455) (Yelp,0.900) (IMDb,0.951)};
        \addplot[olive,fill=olive!30!white]
        coordinates{(HellaSWAG,0.571) (wikiHow goal-step,0.895) (wikiHow temporal,0.690) (WinoGrande,0.719) (Yelp,0.901) (IMDb,0.949)};
        \addplot[red,fill=red!30!white]
        coordinates{(HellaSWAG,0.606) (wikiHow goal-step,0.898) (wikiHow temporal,0.727) (WinoGrande,0.716) (Yelp,0.907) (IMDb,0.951)};
        \addplot[orange,fill=orange!30!white]
        coordinates{(HellaSWAG,0.487) (wikiHow goal-step,0.787) (wikiHow temporal,0.768) (WinoGrande,0.684) (Yelp,0.900) (IMDb,0.941)};
        \legend{\texttt{Vanilla}, \texttt{VI}, \texttt{VIC}, \texttt{CVIC}}
    \end{axis}
\end{tikzpicture}
\caption{Comparison of \texttt{code-davinci-002} across the four types of code prompts. Figures are split to allow for different y-axis scales. We see that different prompts do better on different tasks and while some tasks have high variance over prompt types, others do not.}
\label{fig:code_prompts}
\end{figure}

\begin{figure}
    \centering
    \includegraphics[width=0.7\columnwidth]{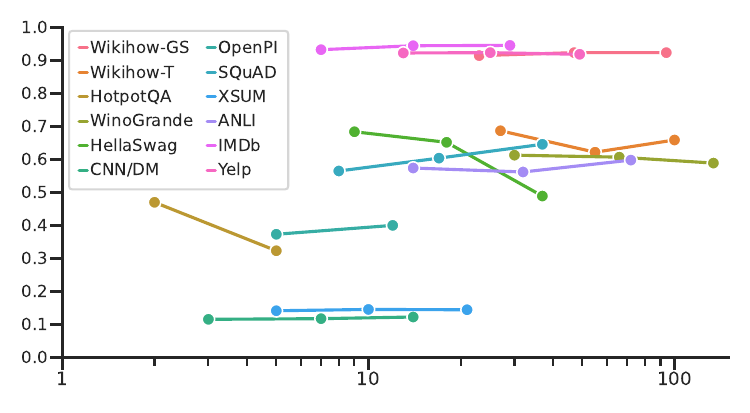}
    \caption{Performance score (y-axis) vs number of in-context examples (x-axis, in log scale) using code prompts (\texttt{VIC}) with \texttt{code-davinci-002}. We see that increasing number of examples does not always increase performance and in some cases makes it worse.}
    \label{fig:prompt_lengths}
\end{figure}

\paragraph{How many in-context examples should we include in our code prompt?}
We would like to also investigate how the number of in-context examples in the prompt affects models' ability to perform the task. We therefore conducted an experiment where we filled the context window of \texttt{code-davinci-002} with in-context examples up to 2000 tokens, 4000 tokens, and 8000 tokens and plotted the validation accuracy of the model with respect to the number of examples in Figure~\ref{fig:prompt_lengths}. 

Contrary to expectations, we find that the number of in-context examples has little effect on model performance for most tasks and actually has a \textit{negative} effect on some tasks. This is especially interesting given that previous work on in-context learning with text prompts finds roughly monotonic improvement from adding more in-context examples \citep{liu-etal-2022-makes}. While further research is necessary, it seems that code prompts may have different scaling behavior than text prompts when used in in-context learning. 

\paragraph{Which is better: code or text prompts?}
In our main experiment we compare the performance of the three GPT models on code prompts (\texttt{VIC} style) and text prompts across the 12 datasets. Given the results from Figure~\ref{fig:prompt_lengths}, we fill the context window of all models with in-context examples up to 4000 tokens to serve as a middle ground for comparing code and text prompts. We report the results of our main experiment in Table~\ref{tab:textvscode} and see several surprising trends.

First, we find that prompting LLMs with code leads to substantial increases in performance for certain few reasoning tasks but that this trend does not hold across all tasks---or even all reasoning tasks. For example, when using code prompts with \texttt{code-davinci-002}, we see a 10.5\% accuracy increase on wikiHow temporal ordering but a 2.6\% accuracy decrease on wikiHow goal-step inference despite both being commonsense reasoning tasks and having identical source material.

Second, we find that supervised instruction fine-tuning on natural language demonstrations does not hurt model performance on code. Rather, we observe that code prompts outperform text prompts on \textit{more} tasks when using \texttt{text-davinci-002} than when using \texttt{code-davinci-002} despite the fact that \texttt{text-davinci-002} received no additional fine-tuning on code instructions.

Finally, we find that LMs not explicitly trained on code can also benefit from code prompting on certain reasoning tasks. In particular, code prompts outperform text prompts on \texttt{davinci} for 3 out of our 12 tasks---the same proportion as \texttt{code-davinci-002}. The tasks that benefit from code prompts also seem to be largely consistent across the three types of models tested, suggesting some underlying trend as to which tasks systematically benefit from structured input.

\begin{table*}
    \small
    \centering
    \begin{adjustbox}{max width=\textwidth}
    \begin{tabular}{l|l|c|c|c|c|c|c|c|c|c}
    \toprule
        Dataset & Metric & \multicolumn{3}{c|}{\texttt{davinci}} & \multicolumn{3}{c|}{\texttt{code-002}} & \multicolumn{3}{c}{\texttt{text-002}}\\
        & & +Text & +Code & $\Delta$ & +Text & +Code & $\Delta$ & +Text & +Code &$\Delta$\\
        \midrule
        Hellaswag &Accuracy&0.321&0.307&{\color{lightred}-0.014}&0.652&0.606&{\color{lightred}-0.046}&0.717&0.773&{\color{cyan}+0.046}\\
        wikiHow goal-step&Accuracy&0.347&0.302&{\color{lightred}-0.045}&0.924&0.898&{\color{lightred}-0.026}&0.919&0.915&{\color{lightred}-0.004}\\
        wikiHow temporal&Accuracy&0.495&0.532&{\color{cyan}+0.037}&0.622&0.727&{\color{blue}+0.105}&0.688&0.761&{\color{cyan}+0.073}\\
        Yelp&Pearson $\rho$&0.913&0.896&{\color{lightred}-0.017}&0.924&0.907&{\color{lightred}-0.017}&0.919&0.904&{\color{lightred}-0.015}\\
        IMDb&Accuracy&0.872&0.935&{\color{cyan}+0.063}&0.945&0.951&{\color{cyan}+0.006}&0.940&0.952&{\color{cyan}+0.012}\\
        WinoGrande&Accuracy&0.513&0.500&{\color{lightred}-0.013}&0.607&0.716&{\color{blue}+0.109}&0.628&0.726&{\color{cyan}+0.098}\\
        ANLI&Accuracy&0.333&0.360&{\color{cyan}+0.027}&0.562&0.551&{\color{lightred}-0.011}&0.504&0.557&{\color{cyan}+0.053}\\
        HotpotQA&Macro-F1&-&-&-&0.470&0.449&{\color{lightred}-0.021}&0.490&0.350&{\color{lightred}-0.140}\\
        SQuAD&Macro-F1&0.482&0.466&{\color{lightred}-0.016}&0.604&0.579&{\color{lightred}-0.025}&0.670&0.656&{\color{lightred}-0.014}\\
        OpenPI&ROUGE-F1&-&-&-&37.33&36.36&{\color{lightred}-0.970}&35.60&31.30&{\color{darkred}-4.300}\\
        CNN/Daily Mail&ROUGE-2&9.28&9.13&{\color{lightred}-0.150}&11.74&11.67&{\color{lightred}-0.070}&13.63&13.55&{\color{lightred}-0.080}\\
        XSUM&ROUGE-2&9.38&6.83&{\color{darkred}-2.550}&14.51 &11.03&{\color{darkred}-3.580}&14.48&13.26&{\color{darkred}-1.220}\\
        \bottomrule
    \end{tabular}
    \end{adjustbox}
    \caption{Performance of the three LMs when using code prompts (+Code) vs. using text prompts (+Text). Blank cells indicate tasks for which single test examples could not fit in the context window. Color indicates how code prompts compare to text prompts. We see that while code prompts outperform text prompts for certain tasks (such as wikiHow temporal and WinoGrande) text prompts are better on average. We also find that instruction fine-tuning (\texttt{text-002}) allows for better code prompt utilization.}
    \label{tab:textvscode}
\end{table*}

Our rather anti-climatic findings above suggest that a symbolic form of data does not consistent benefit modern LLMs pre-trained with a mixture of text and code,\footnote{Our findings directly contradict a contemporaneous work \citep{mishra-etal-2023-prompting}, which experimented with a set of much smaller LLMs.} despite previously reported success of symbolic form on some event reasoning tasks. 

The work above was published in \citet{zhang-etal-2023-exploring}, in which I formulated the task and primarily contributed to a third of the tasks, with the efforts of data collection, model setup, and evaluation. I also performed the majority of the analysis. I have obtained approval from all collaborators to exclusively include this work in this thesis. 

\section{Summary}

In this chapter, I introduced the entity schema, which is a semi-symbolic structured representation of events. I start by motivating the need of involving entities in the process of causal reasoning about events. Then, I propose the \openpitwo that can be used to evaluate LLMs that can predict the entity schema. Using it, I show an improvement in the downstream task of event reasoning, and also drawing attention to the form that a structured representation should take to interface with LLMs. Finally, I show that LLMs can take advantage of pre-training on code and work better with a code form of the entity schema. However, such a code form does not consistently benefit general NLP tasks. 

At this point, my quest with ``structured reasoning'' has become more structured thanks to the semi-symbolic representation. However, the reliance on LLMs to make the final prediction lacks faithfulness. In other words, even were the representation to be completely correct, the LLM that takes it as input might still make a mistake. In the opening example of this chapter, even the knowledge of ``the pan is hot'' were to be given, the final answer that ``one should not touch the pan'' might not be reached, due to the non-deterministic nature of LLMs. To address this, in the next Chapter, I propose a fully symbolic representation that is no longer fed to LLMs, but deterministic algorithms.


\newpage

\chapter{WORLD MODEL: A SYMBOLIC LANGUAGE REPRESENTATION}
\label{chap:world}



In all previous sections, we have focused on defining and predicting a structured event representation that can be leveraged by LLMs in different ways. As LLMs are the only type of models in each pipeline, the performance is limited by their power given any task. Despite success in previously discussed reasoning tasks, there exist tasks that are more challenging. One example is text-based \textbf{symbolic planning}, where the conditions and the goal are described using natural language, and a model must pick a sequence of actions to achieve the task. For example, the following are the instructions of a simple block-moving task:
\begin{quote}
    On a table, I have five blocks stacked on top of each other. From the bottom to the top, we have blue, white, yellow, and brown. Your goal is to rearrange them into red, white, blue, yellow, and brown from top to bottom. You may only move a block with nothing on top of it, and you can put such a block on a table. What should you do?
\end{quote}
While even a human child is likely to find the task trivial by some trials, a state-of-the-art LLM like ChatGPT is completely incapacitated\footnote{Experiment is performed in January 2024 with OpenAI ChatGPT4.} \citep{valmeekam2023planbench}. Note that this is a symbolic reasoning task because even though the input is natural language, it describes a symbolic configuration, operating under a rule-based transition function, and the output space is also discrete and finite. Despite recent LLMs' emergent ability (see Section~\ref{sec:llm_history}) to perform some  symbolic reasoning tasks to some extent, this failure still stands in contrast with LLMs'  success in generating non-symbolic, unstructured plans, perhaps in response to a query like ``how to cook eggs.'' 


Let us also examine a second example, a grade-school math question:
\begin{displayquote}
    I have 5 apples. You have 4. How many do we have in total?
\end{displayquote}
Solving the problem requires the natural language ability to understand that `\textit{in total}' means addition as well as the symbolic reasoning ability to mathematically perform the addition. While LLMs can often extract the numerals and even come up with the correct formula, it systematically fails at arithmetic calculation\footnote{Closed-source, commercial tools like OpenAI ChatGPT in 2024 is likely a pipeline including but not limited to one or more LLMs. It has been able to correct answer this type of math questions by translating it to Python code, similar to the ideas in my own, my collaborators', and other work that I will discuss next. In the paper, we continue to refer to LLMs as the end-to-end models themselves, not pipelines that involve them.} \citep{lyu2023faithful}. To improve, one may attempt to work on LLMs that can better perform arithmetic calculations, or simply use a calculator to reliably reach the final answer (Figure~\ref{fig:math_example}), as long as the upstream LLM can provide a well-formed input.
\begin{figure}
    \centering
    \includegraphics[width=\columnwidth]{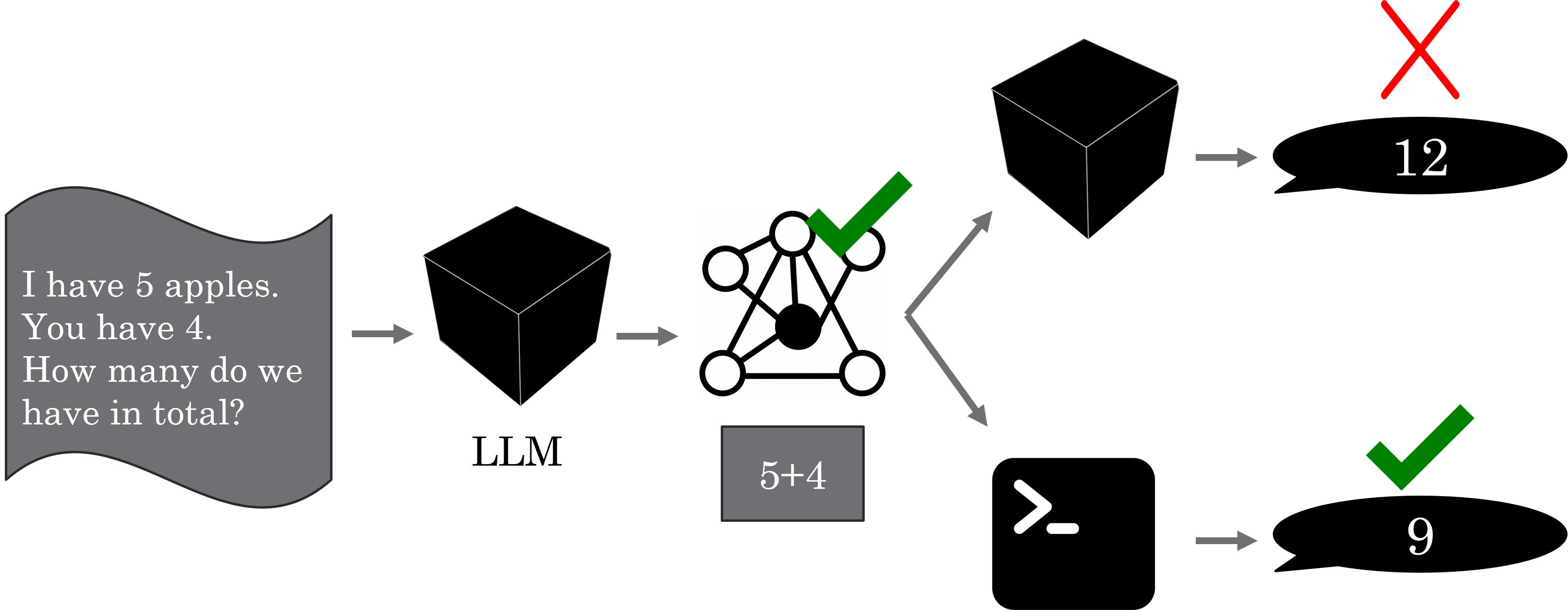}
    \caption{An example of a math question involving symbolic reasoning.}
    \label{fig:math_example}
\end{figure}

\begin{figure}
    \centering
    \includegraphics[scale=0.6]{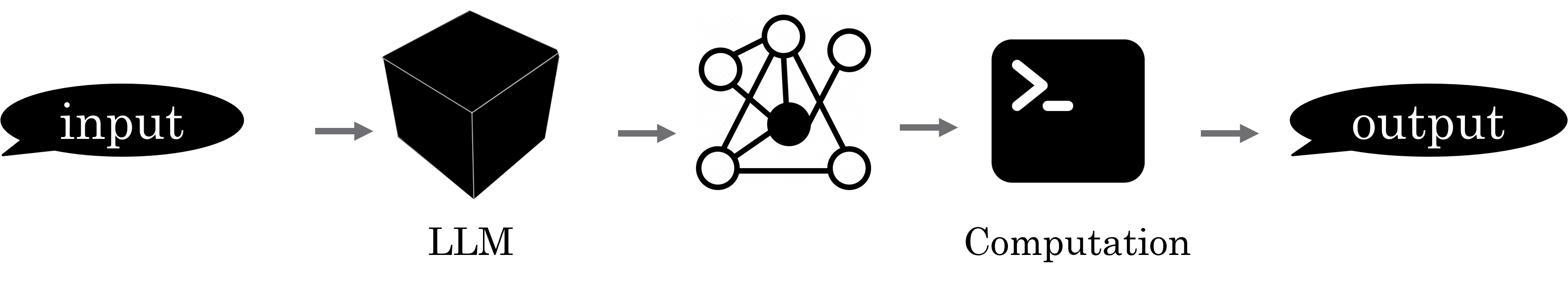}
    \caption{An illustration of my proposed pipeline leveraging a symbolic representation of the world pertaining to the problem. The representation is inferred by an LLM and fed to a computational tool such as an algorithm.}
    \label{fig:pipeline_sl}
\end{figure}

In this thesis, I have demonstrated many ways to use LLMs to solve a problem (Figure~\ref{fig:usages}), including an end-to-end usage (baseline), fine-tuning an LLM with structured data (Chapter~\ref{chap:relation}), and prompting an LLM with structured data (Chapter~\ref{chap:entity}). Following the idea above, I will now introduce a \textbf{neurosymbolic usage}, where an LLM no longer predict the final answer. Instead, its job ends at predicting the structured representation, with which some external computation is responsible for reaching the final answer. This approach is akin to program synthesis \citep{austin2021program} in the programming language community, using a generative model (e.g., LLM) to produce executable code. However, my work is practically scoped within problem solving and event reasoning, considering only specific representations for each downstream task.

\section{Preliminary Experiments: Neurosymbolic Usage with LLMs}
\label{sec:fcot}

As introduced in Section~\ref{sec:llm_history}, the ability to generate well-formed code is an emergent ability of LLMs \citep{feng-etal-2020-codebert,chen2021evaluating,roziere2023code}. However, the neurosymbolic usage of LLMs, namely, using the generated code a means to perform reasoning tasks, is a new concept proposed by \citet{lyu2023faithful}, a work that I secondarily contributed to, along with several contemporaneous work \citep{chen2022program,gao2023pal}. I will fleetingly discuss this work.

In \citet{lyu2023faithful}, I tackle the task of kinship deduction. Concretely, given a problem written in natural language, the answer is as a string-valued variable. I consider the CLUTRR \citep{sinha-etal-2019-clutrr} dataset that involves inferring the family relationship (e.g., ``grandson'') between two people from a short story (e.g., ``[Gabrielle] drove her daughter [Dorothy] to the hospital. [Dorothy]'s son [Vincent] showed up shortly after. How is [Vincent] related to [Gabrielle]?'', Figure~\ref{fig:clutrr}). This is a symbolic reasoning task where the performance of end-to-end LLMs drastically degrades as the complexity of the problem increases (i.e., more people are involved). In our neurosymbolic usage, we prompt the LM to generate Python code that essentially breaks down the question into sub-questions (``How is [Vincent] related to [Dorothy]'' and ``How is [Dorothy] related to [Gabrielle]''), as well as provide input extracts as rationales to support the answer (``[Dorothy]'s son [Vincent] showed up shortly after'', etc.) as comments. The code for each sub-question is a relational expression representing the relation between the mentioned entities, for example, \texttt{relation(Vincent, Dorothy)=son} denotes that Vincent is Dorothy's son. Once we have the code, a fully defined symbolic representation of the problem, the LLM's job is finished. We then use a simple relational inference engine that relies on a set of transitivity rules provided by \citet{zhang2022improved} among possible family relationships, e.g., \texttt{son@daughter=grandson} (the son of one's daughter is one's grandson). Our solver recursively applies these rules on $C_{SL}$ to derive $A$, and determine that Vincent is Gabrielle's grandson.

\begin{figure}
    \centering
    \includegraphics[width=0.9\columnwidth]{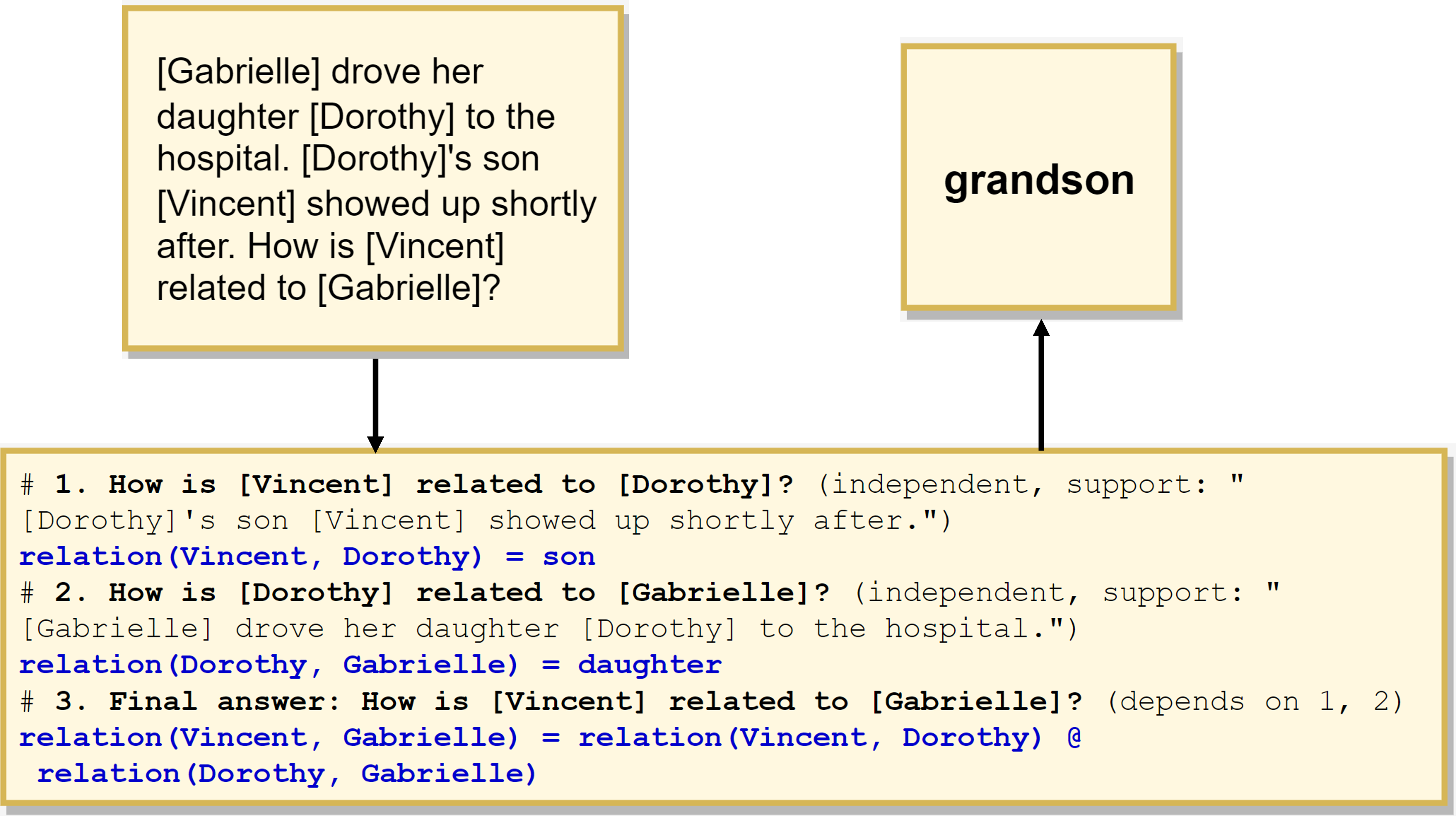}
    \caption{An example from the CLUTRR dataset.}
    \label{fig:clutrr}
\end{figure}

On the benchmark, I construct the prompt using 8 exemplars with $K \in \{2,3\}$, where $K$ is the number of intermediate steps, and test the models on the remaining examples with $K$ up to 10. For \texttt{code-davinci-002}, I observe a 45.7\% to 71.9\% performance leap compared to the end-to-end usage when using the neurosymbolic usage. We can interpret this result from two angles. Regarding performance, the external solver relieves the burden of problem solving from the LLMs, without which the accuracy degrades by 19.4\%. Regarding trustworthiness, due to the deterministic nature of the external solver, the answer is correct if and only if the interim representation (pairwise relationships) provided by the LLM is correct. Thus, a verifying agent (either human or automatic) may simply and locally check if each pairwise relationship is correctly extracted from the text. They may also trivially fix any error in the process. Such ability for a human user to interpret, verify, and correct the LLM output is a result of the neurosymbolic usage, not offered by the end-to-end usage.

At this point, I have introduced the neurosymbolic usage of LLMs and shown its preliminary success. In the next section, I will show that the neurosymbolic usage is an indispensable tool to push the limit of structured reasoning of events, by attempting a historically challenging task: planning. 

The work above was published in \citet{lyu2023faithful}, in which I contributed to all experiments on kinship deduction. 

Now that I have demonstrated the feasibility and appeal of the neurosymbolic usage of LLMs, I will return to the context of event-centric reasoning. In the next Section, I will explore the task of classical planning which instantiates said methodology. My proposal is to have LLMs generate not the solution, but a symbolic representation of the environment, namely a \textbf{world model} that can be executed by some external computation like a symbolic planner. I will next outline this methodology.


\section{Planning}
\label{sec:planning}

In this chapter, I will tackle the challenging task of planning that comes in two flavors: formal and informal. 

\subsection{Formal Planning}
In the AI community, planning is the task of finding a sequence of actions to achieve a goal in a given environment \citep{fikes1971strips,lavalle2006planning}. Over more than five decades of development, the task of planning has taken on many flavors. I will particular focus on the formulation of classical planning, where an initial and a goal configuration are defined as a collection of entity states. Each action gives rise to an event during which the states change in certain way. Therefore, the aim is to perform certain actions to drive the entity states in the environment to the goal configuration. Classical planning is both symbolic and formal, as all the involved concepts including entities, states, and actions are fully grounded to symbols (instead of natural language that we discussed in previous sections). Given a domain definition (including tuple of initial configuration, goal configuration, and actions), a plan can be found deterministically, if there is one. In other words, it does not require any inference or guesswork to find a plan. 

The Planning Domain Definition Language (PDDL) \citep{aeronautiques1998pddl} is a programming language designed for classical planning. A PDDL instance contains a domain file \df and a problem file \pf (Figure~\ref{fig:pddl_example}.\\
A \df defines the following elements:
\begin{itemize}[topsep=-2ex,itemsep=-1ex,partopsep=1ex,parsep=1ex,leftmargin=*]
    \item a header $H$, which consists of
    \begin{itemize}[topsep=-2ex,itemsep=-1ex,partopsep=1ex,parsep=1ex,leftmargin=*]
        \item types of entities (e.g., \textit{object}, \textit{location}, \textit{player})
        \item predicates (e.g., if object is \textit{at} a location)
        \item names of possible actions (e.g., \textit{boil water})
    \end{itemize}

    \item definitions of actions $A$, which consist of
    \begin{itemize}[topsep=-2ex,itemsep=-1ex,partopsep=1ex,parsep=1ex,leftmargin=*]
        \item parameters (e.g., water, pot) as a list of typed variables
        \item preconditions (e.g., water and pot belongs to player; water is not treated) as a conjunctive normal form of predicates
        \item effect (e.g., water is treated) as a conjunctive normal form of predicates
    \end{itemize}
\end{itemize}
A \pf defines the following elements:
\begin{itemize}[topsep=-2ex,itemsep=-1ex,partopsep=1ex,parsep=1ex,leftmargin=*]
    \item objects (e.g., rainwater)
    \item initial states (e.g., bucket is empty)
    \item goal states (e.g., bucket is filled with rainwater; rainwater is treated)
\end{itemize}
We say that a \df can \textit{solve} a \pf if there exists a sequence of actions $A_1,\dots,A_n$ that results in a transition from the initial state to the goal state. 

\begin{figure}
    \centering
    \includegraphics[width=0.7\textwidth]{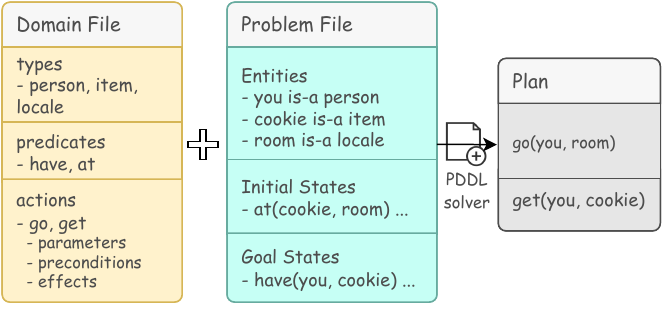}
    \caption{A PDDL solver produces a plan based on a minimal domain file and problem file. Previous work assumes the domain file as given, while we predict the action definitions in the domain file.}
    \label{fig:pddl_example}
\end{figure}

\subsection{Informal Planning}

However, there rarely exist formally defined, fully symbolic environments in the real life. Real-life environments are defined in many modalities; in the scope of NLP, we will focus on those defined using natural language texts. In Section~\ref{sec:script_generation}, the task of script generation can also be thought of as planning, since a user would like to perform the steps and reach the goal. To make an analogy to PDDL, the \df is equivalent to a textual description or model's parametric knowledge of the environment, and the \pf is equivalent to the goal and the inferred stats-quo. Because natural language is implicit and under-specified, such planning is informal as there are myriad undefined variables with regard to entities, states, and actions. The task of informal planning using natural language requires event reasoning to fill in these blanks. Ultimately, I believe this is also the most practical task to solve in the real world. 

\subsection{Neurosymbolic Method}

\begin{figure}
    \centering
    \includegraphics[width=\textwidth]{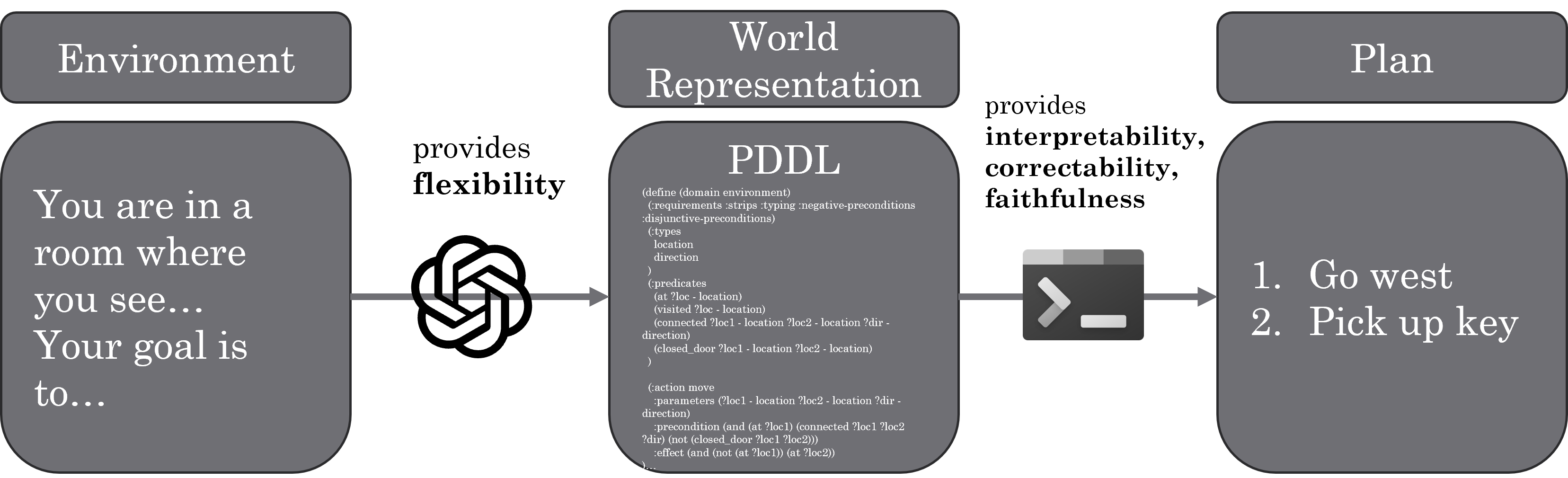}
    \caption{The proposed world-building planning methodology that learns a formal representation of an informal description of the environment.}
    \label{fig:world_building}
\end{figure}

For the task of planning with natural language, Section~\ref{sec:script_generation} has already shown the insufficiency of LLMs with both the end-to-usage and fine-tuning with relational structure. I now turn to a neurosymbolic methodology, where the setup of formal planning becomes a means to an end, introducing structure into the methodology. Concretely, my task is to \textbf{learn a structured domain definition} (i.e., \df and \pf) from unstructured natural language texts using LLMs. Once this is done correctly, a search-based planner can trivially find a plan, that may be presented to user after being converted to natural language or other modalities (Figure~\ref{fig:world_building}). In this methodology, an LLM's task is limited to \textbf{world-building}, coming up with a `mental representation' (that is structured and formal) while interacting with a textual environment (that is unstructured and informal). The actual task of finding the plan is delegated to an external solver, much like Section~\ref{sec:fcot}. 

As the output of planning is a sequence of actions, the task planning can be seen as an instance of event-centric reasoning. In my proposed framework of structured reasoning, the structured representation here, expressed by PDDL, is precisely an entity schema much like Section~\ref{sec:openpi} and \ref{sec:crepe}. Each action event is defined with preconditions and effects, both of which revolve around entities, or more precisely, their state changes. The process of modeling the actions is almost like the reverse of predicting entity states in a given procedure. The aim of the former is to \textit{infer} an entity schema in order to \textit{find} a plan. The aim of the latter is to \textit{reconstruct} an entity schema from an \textit{existing} plan in a post-hoc manner. 

From a high level, the approach I will take for planning is a neurosymbolic usage of LLMs that generate a structured representation, or entity schema, or world model, given textual context. I will show that this approach is superior to end-to-end usage of LLMs when it comes to both performance and trustworthiness. 

\subsection{\df and \pf}

Following the two-part design of PDDL, the \textit{structured domain definition} or \textit{world model} discussed above consist of two components: \df and \pf. The \pf, which specifies the initial and goal entities states, is much easier to learn from text \citep{lyu2023faithful,xie2023translating,liu2023llmp}. For example, the text:
\begin{displayquote}
    Block A is on top of Block B, which is on the table.
\end{displayquote}
needs to be translated into the following partial \pf:

\begin{lstlisting}[numbers=none, basicstyle=\fontfamily{cmtt}\small,style=pythoncode,belowskip=0.3\baselineskip,aboveskip=0.3\baselineskip,commentstyle=\color{codegreen}]
    on(A,B)
    on_table(B)
\end{lstlisting}

On the other hand, learning the \df is much more challenging. The crux of the challenge lies in action modeling, in which much information must be inferred from the text. For example, the text:
\begin{displayquote}
    You pick the lock.
\end{displayquote}
may correspond to the following partial \df:

\begin{lstlisting}[numbers=none, basicstyle=\fontfamily{cmtt}\small,style=pythoncode,belowskip=0.3\baselineskip,aboveskip=0.3\baselineskip,commentstyle=\color{codegreen}]
(:action pick-lock
    :parameters (?lock ?door ?room1 ?room2)
    :precondition (and
        (not (picked ?lock))
        (locked ?door)
        (not (accessible ?room1 ?room2))
    )
    :effect (and
        (picked ?lock)
        (not (locked ?door))
        (accessible ?room1 ?room2)
    )
)
\end{lstlisting}

Evidently, the action models in the \df are challenging to automatically generate, and require extensive manual effort to annotate for every domain. Much previous work tackles action modeling via plan traces \citep{yang2007learning,cresswell2013acquiring,lamanna2021online}, which is not our focus here. Few work that learns the \df from text \citep{yordanova2016learning,lindsay2017framer,hayton2017storyframer,hayton2020narrative,jin2022text,li2023automated}. These efforts have two issues. First, they involves complex pipeline approaches, many heavily templated and rule-based, and thus can unlikely generalize to diverse domains. Second, they have only targeted a static environment, performing a one-off translation or extraction from text to PDDL, much unlike the real world where an agent can interact with the environment. To overcome the first issue, LLMs with the newly gained ability to generate structured code seem like the tool of choice. However, it is known that LLMs struggle at generating low-resource domain specific languages \citep{10.1145/3618305.3623612} like PDDL. In Section~\ref{sec:plan_procedure}, I will discuss efforts to have LLMs generate correct and executable PDDL \df given complex text descriptions of an environment. To overcome the second issue, in Section~\ref{sec:plan_interaction}, I target interactive textual environments where an agent not only learns their structured representation, but also continue to grow and refine said representation. 

\section{Planning for Procedural Text}
\label{sec:plan_procedure}

A \df contains types, predicates, and actions (Figure~\ref{fig:pddl_example}). Like previous work, we specifically focus on action modeling of a \df. Given types, predicate, and names of actions of an arbitrary domain and some textual description, we predict the preconditions and effects of each action to complete the domain definition. As a real-life motivating example, a kitchen robot may have access to cookwares and ingredients as well as the actions it can perform like ``swinging a knife'', but it still needs to predict the precondition that it is only safe to do so in front of a cutting board and the effect that the ingredients will become diced. 
Once the domain is completely defined, along with the problem definition \pf, an off-the-shelf solver can deterministically find a plan given a query.

Next, we provide some first attempts towards the action modeling task. As PDDL for any domain is extremely costly to annotate, there is hardly any data for training a model. Therefore, we demonstrate how our task can be tackled by zero-shot prompting state-of-the-art LMs.

\paragraph{Methodology}
To predict action definitions in \df based on the header and some text \txt, we prompt an LM in a zero-shot manner \cite{brown2020language} by describing the task and providing the input. Note that few-shot prompting is prohibitively costly due to the excessive length of a resulting \df. 

We also incorporate the chain-of-thought (CoT) technique \cite{wei2022chain} that explicitly prompts an LM to perform three essential sub-tasks:

\begin{itemize}[topsep=-2ex,itemsep=-1ex,partopsep=1ex,parsep=1ex,leftmargin=*]
\item Summarization: describe each action, including the expected preconditions and effects;
\item Extraction: list the involved entities and their states before and after;
\item Translation: based on the information above, convert \txt to PDDL.
\end{itemize}


We experiment with two large LMs, \texttt{gpt-3.5-turbo-16k} and \texttt{gpt-4} dated June 2023 with max tokens of $8192$, temperature of $0.5$, and default hyperparameters otherwise. 

\paragraph{Dataset}

We introduce the \proc dataset of 27 different \txt-\df-\pfs tuples, drawing procedural texts from wikiHow articles of various topics. A class of graduate students in a U.S. university with prior knowledge of PDDL are each given a wikiHow article and annotate a \df and multiple corresponding \pfs from the article, each with a gold plan to solve it. On average, there are 13.33 defined actions in a \df and 8.07 instantiated actions in a gold plan. During prediction, we treat \txt as an annotated one-line summary of all annotated actions.

On average, it takes several hours to train each human annotator, and another several hours to produce a \txt-\df-\pfs tuple. Upon post-inspection, we notice that about 15\% are problematic and have to discard or fix them. This finding re-emphasize the great difficulty of annotating domain definitions and motivates our automatic prediction of \df. 

We partition the 27 examples into a 5:6:16 train-development-test splits. In this work, the train split is unused as all our methods are zero-shot; only the development set is used for error analysis; the test set is strictly held out for evaluation. 

\begin{figure*}[t!]
    \centering
    \includegraphics[width=\textwidth]{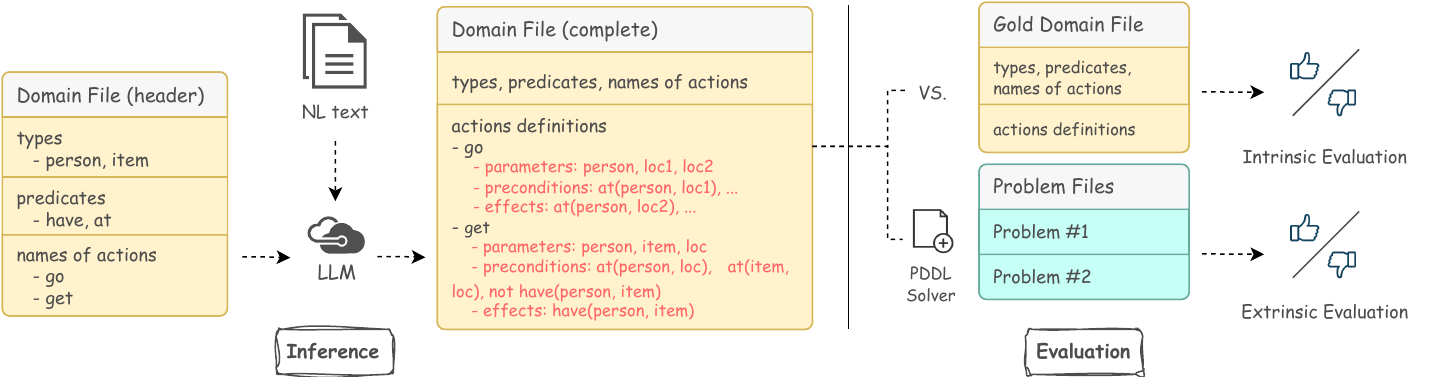}
    \caption{Our formulation of the \df-action prediction task. Given the domain file header that specifies the ontology, a model predicts action definitions including parameters, preconditions, and effects based on textual descriptions. During evaluation, the predicted \df is both compared to a reference and used to solve corresponding \pfs.}
    \label{fig:formulation}
\end{figure*}

\paragraph{Evaluation}

Now that a model generates the parameters, preconditions, and effects for each action, we have a complete \df. We evaluate it in two ways (Figure~\ref{fig:formulation}). \textit{Intrinsically}, we semantically compare the predicted $A$ with the ground-truth provided by our \openpitwo and report an action-wide accuracy. Equivalence of two action definitions does not depend on the naming of variables nor on the order within conjunctions. \textit{Extrinsically}, to measure actions' coherence, we use a BFS-based PDDL solver\footnote{\url{https://github.com/pucrs-automated-planning/pddl-parser}} to attempt to solve ground-truth \pfs with the predicted \df and report a success rate. An unsolved \pf is caused by (1.) no plan can be found, or (2.) the solver runs for more than 30 seconds, or (3.) the solver returns an error (usually a syntax error in generated PDDL).

\begin{table}[t!]
\centering
\begin{tabular}{llll}
\toprule
            & \multicolumn{1}{c}{Intrinsic} & \multicolumn{2}{c}{Extrinsic} \\ 
 Model    \%   & action acc.                  & \pf solve  & exact plan \\ \midrule
 \texttt{gpt-3.5}    & 0.2                         & 1.0       & 1.0         \\
 \texttt{gpt-4}       & 15.9                         & 33.7       & 4.2         \\
 \texttt{gpt-4} + CoT  & \textbf{18.1}                         & \textbf{35.8}       & \textbf{6.3}         \\
 gold & 100.0                             & 100.0             &  100.0            \\ \bottomrule
\end{tabular}
\caption{Performance of the \df-action prediction on the concatenation of the development and test set of \openpitwo. Metrics include action-wide accuracy, average edit distance of action definitions, the proportion of \pfs that can be solved, and the proportion of generated plans that exactly match the gold plans.}
\label{tab:results}
\end{table}

The intrinsic and extrinsic results are shown in Table~\ref{tab:results}. \texttt{gpt-3.5-turbo}, also known as ChatGPT which achieves impressive performance on many tasks, has a close-to-zero performance. In contrast, its predecessor \texttt{gpt-4} performs significantly better with 18\% action prediction accuracy and 36\% solve rate of \pfs based on the predicted \df. Still, the performance far worse than ideal, showing that even a simplified open-domain planning formulation is challenging to state-of-the-art LMs. 

CoT is helpful overall since it explicitly spells out many implicit entities and state changes in the extraction stage which are critical to predicting preconditions and effects. 
In most situations, the model summarizes the action and extracts the entity states correctly, though sometimes missing a few implicit entities. However, CoT's bottleneck lies in the translation stage, during which there are mainly three types of errors. 

\begin{enumerate}[topsep=-2ex,itemsep=-1ex,partopsep=1ex,parsep=1ex,leftmargin=*]
    \item mismatched predicates: the model uses \texttt{(at ?loc ?item)} instead of \texttt{(inventory ?item)};
    \item hallucinating predicates: the model creates a new predicate \texttt{(soaked ?item)} while neglecting the existing \texttt{(submerged ?item)};
    \item complicating predicates: the model adds unnecessary predicates \texttt{(inventory ?submerged\_item - item)} when already has \texttt{(inventory ?item)}.
\end{enumerate}

To address these, we leave to future work to demonstrate and standardize the translation process by clearly describing when an entity change or stage change is and is not needed, while also encouraging the model to strictly match the given predicates. 

\begin{table}[]
\centering
\begin{tabular}{lccc}
\toprule
Model \% & Parameter & Precondition & Effect \\ \midrule
\texttt{gpt-4} & 36.7 & 31.1 & 53.0 \\
\texttt{gpt-4} + CoT & 42.2 & 29.7 & 48.1 \\ \bottomrule
\end{tabular}
\caption{Marginalized intrinsic performance of parameter, precondition, and effect.}
\label{tab:results_intrinsic}
\end{table}

Finer-grained evaluation results are shown in Table~\ref{tab:results_intrinsic} to tease out the performance regarding such component within an action. It is clear that the LM is worse at predicting preconditions than at predicting effects. This is understandable as procedural texts like wikiHow tend to be less explicit about predictions than about effects (e.g., from \textit{bake for 10 minutes} it is obvious that the food will be baked, but it is unclear what state it had been in).

\begin{table}[t!]
    \centering
    \begin{tabular}{llll|ll}
    \toprule
           & \multicolumn{3}{c|}{Unsolved}      & \multicolumn{2}{c}{Solved} \\ \midrule
           & \makecell{Syntax\\Error} & \makecell{Bad\\Action} & \makecell{Good\\Action} & \makecell{Bad\\Plan}    & \makecell{Good\\Plan}    \\
\texttt{gpt-4}  & 3      & 7          & 2           & 0           & 3            \\ \bottomrule
    \end{tabular}
    \caption{Statistics of error types on the development set. }
    \label{tab:error_analysis}
\end{table}

To provide deeper insights into model performance, we manually inspect the model output of \texttt{gpt-4} on all 6 examples (15 \pfs) in the development set. We consider the following scenarios.

\textit{Syntax Error}: Model output may contain illegal expressions that cannot be parsed. For example, \texttt{(inventory ?player (clean ?strips))} is unacceptable because the arguments to a predicate must be atomic types, not another predicate.

\textit{Unsolved}: Whenever the predicted \df cannot solve a \pf, we identify the first problematic action that differs with the ground-truth. For example, if the action \texttt{cut\_plant} misses a critical effect of \texttt{(inventory ?player ?stalk)}, then other actions such as \texttt{graft\_stalk} requiring it cannot be executed. At times, there could be false negatives where the predicted action definitions are in fact reasonable but still cannot lead to a solution.

\textit{Solved}: The predicted \df may solve a \pf, but the plan may be different from the gold plan. It is naturally possible that the predicted plan is a fluke made possible by under-specified preconditions or over-exaggerated effects, as well as loopholes in the \pf leading to unreasonable shortcuts. For the example in Figure~\ref{fig:pddl_example}, a model could \textit{cheat} by defining the action \texttt{get} by not requiring the person and object to be in the same location; thus, the predicted plan would unreasonably omit the action \texttt{go}. However, at times, the predicted plan could also be a reasonable alternative.

The statistics of these errors on the development set is shown in Table~\ref{tab:error_analysis}. When no solution can be found, true negative is highly likely as the model indeed makes aforementioned mistakes during action prediction. When some solution is found, false positive is still possible as the predicted plan may be unreasonable. See attached materials for a complete error analysis of these examples. Our aforementioned future pipeline that separates summarization and translation would likely mitigate these errors.

Through our experiments, it is clear that LLMs are capable of generating PDDL, specifically \df, and specifically action models, given noisy procedural texts. This is already a leap from previous work devising specialized tools for each domain, mostly narrative texts. However, the problem remains that the current text-to-PDDL world building process is one-off and will not work if the environment dynamically develops. In the next section, I will focus on interactive environments where world building happens during exploration.

The work above was published in \citet{zhang2024proc2pddl}, in which I formulated the task and primarily contributed to the code to clean the dataset, run a PDDL planner, and perform evaluation. I also performed the majority of the analysis. I have obtained approval from all collaborators to exclusively include this work in this thesis. 

\section{Planning for Interactive Environments}
\label{sec:plan_interaction}

\paragraph{Motivation}

In Section~\ref{sec:openpi} and Section~\ref{sec:crepe}, I have proposed and discussed event reasoning tasks that deal with a dynamically changing environment. I have shown that these tasks put forward unique challenges, unlike other tasks we have seen, for event state-of-the-art LLMs. I will now explore the same idea for planning, where it is not possible to derive a plan once-and-for-all. Instead, an agent must iteratively interact with the environment, gain new sights, make plans, adjust them, and eventually reach the goal. 

\begin{figure}
    \centering
    \includegraphics[width=0.5\columnwidth]{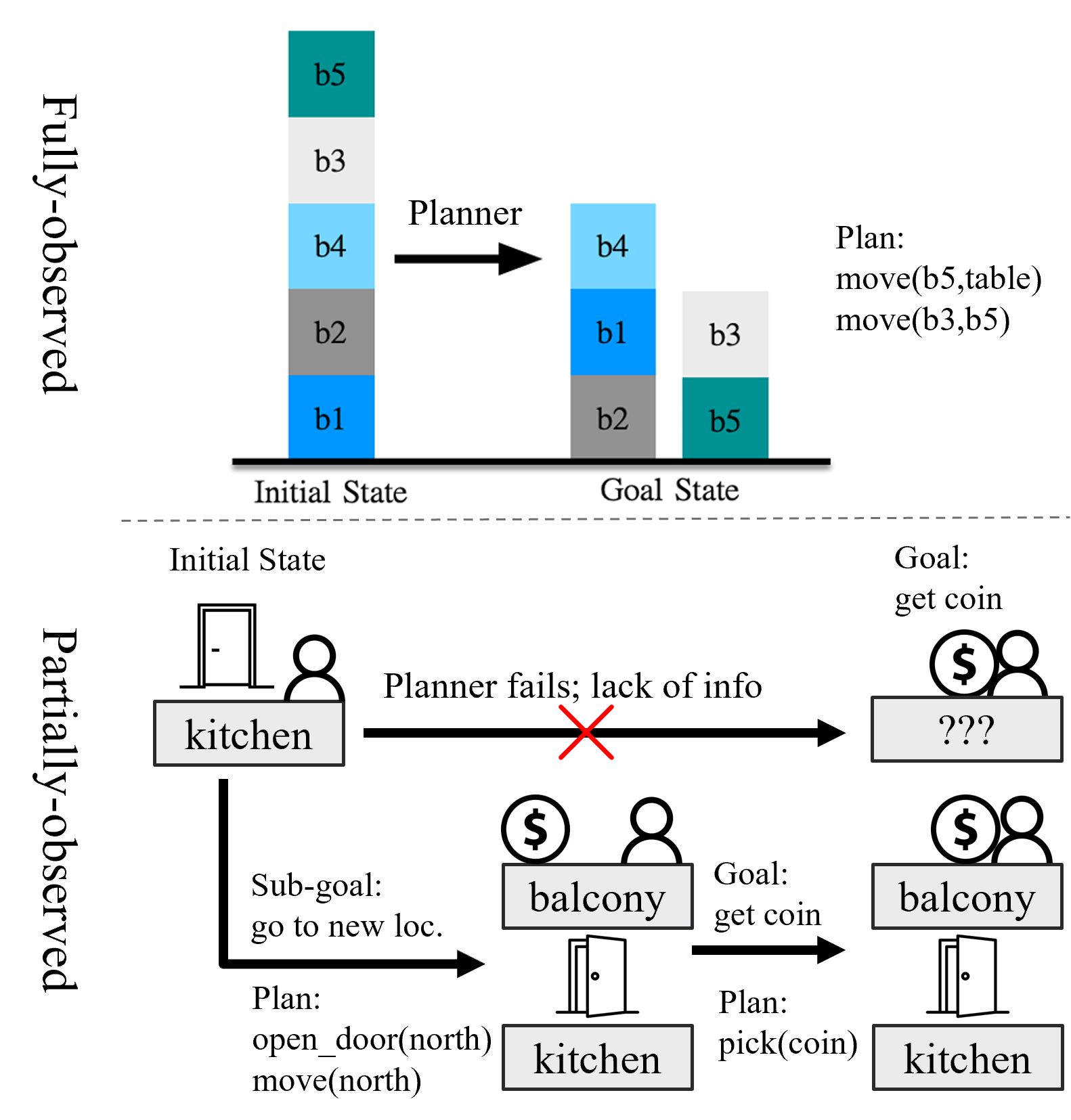}
    \caption{A fully-observed environment like BlocksWorld (upper, to rearrange objects from and to a given configuration) can be tackled by generating a PDDL problem file, while a partially observed one like Coin Collector (lower, to look for an object in an unknown location) cannot until sufficient exploration.}
    \label{fig:fully_partially_observed}
\end{figure}

All previous work on LLM generating PDDL has only experimented on \textbf{fully-observed} environments where all entity states are initially known. Take BlocksWorld as an example (Figure~\ref{fig:fully_partially_observed}, upper), both the initial and goal positions of the blocks are spelled out, in which case existing work use LLMs to translate or parse the NL description of the world state into a PDDL \pf \citep{lyu2023faithful,xie2023translating,liu2023llmp}. With such a complete problem file, a full plan can be found and executed to reach the goal. In contrast, most real-world environments are \textbf{partially-observed} (Figure~\ref{fig:fully_partially_observed}, lower), where the entity states dynamically get uncovered during exploration. Moreover, since the necessary initial and goal states might also be unknown (e.g., looking for an item without knowing where it is), the previous approach falls apart due to the impossibility to specify a complete problem file. This causes a chicken-and-egg problem where a plan is required for exploration, while exploration is required to build PDDL that results in a plan. Given this challenge, past work on partially-observed environments has only used LLMs to directly generate plans \cite{shinn2023reflexion,majumder2023clin} but not PDDL.

We propose \lego, a methodology to incrementally grow the PDDL by using LLMs to translate the NL observations from the environment into entity states expressed as a PDDL problem file. \lego solves the above stalemate of under-defined goal states by recursively falling back to a well-defined sub-goal. This way, a plan can be found to reach the sub-goal, leading to new observations obtained by exploring the environment. In this process, the problem file is incrementally built until a plan can be found for the final goal. 

\begin{figure}
    \centering
    \includegraphics[width=0.5\columnwidth]{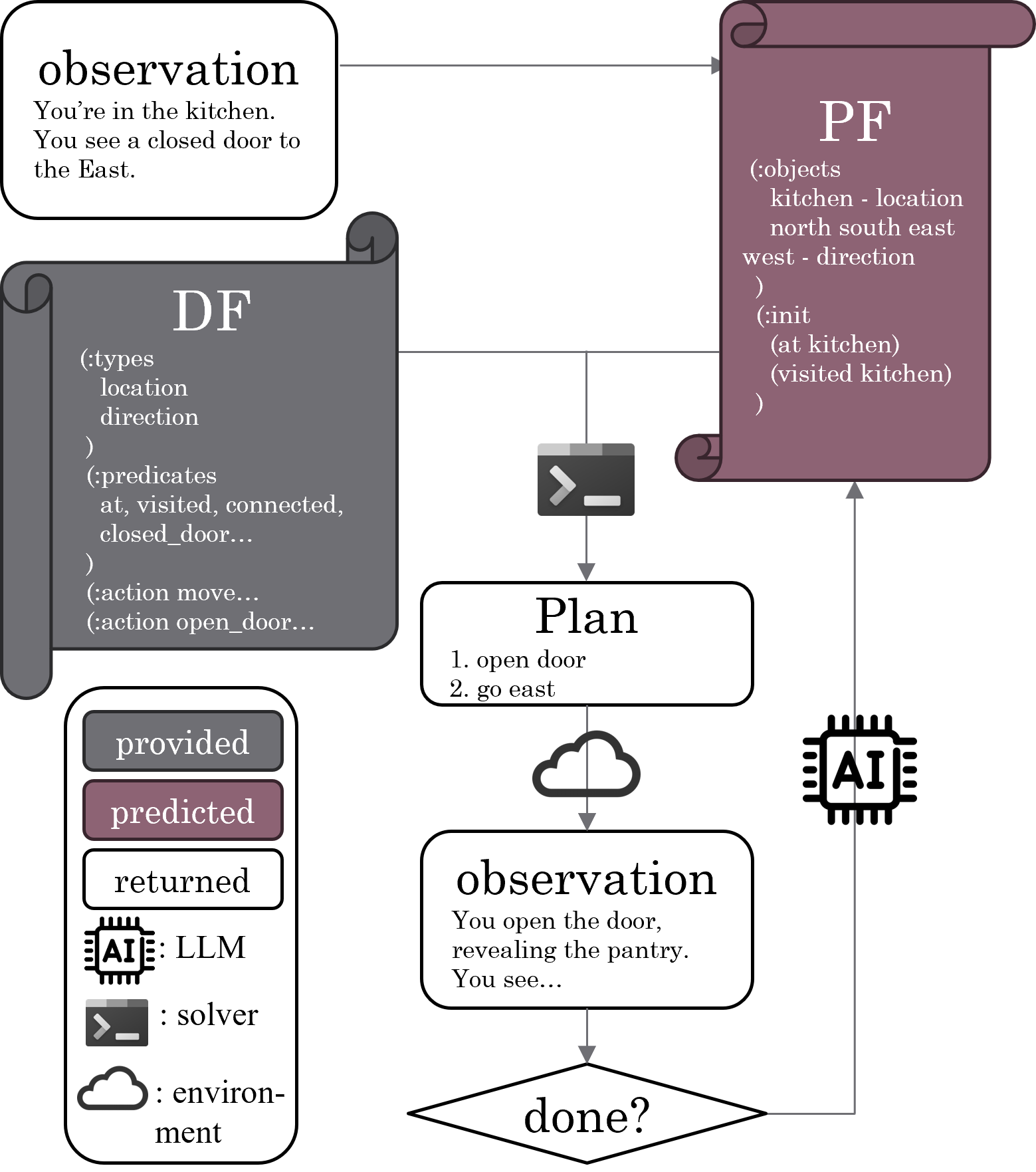}
    \caption{The pipeline of \lego. A PDDL problem file (PF) is incrementally built during exploration.}
    \label{fig:pipeline}
\end{figure}

\paragraph{Methodology}
We operate in a partially-observed simulated environment which functions as a multi-turn interaction between the environment and the agent (e.g., \textit{a game to find an item}). Specifically, the environment provides an NL observation (\textit{objects in a room}) along with a list of permitted actions (\textit{move, pick up}), then the agent selects on of these actions, and repeats. The environment can be seen as a finite state machine where each state consists of the conjunction of all entity states and determines the permitted actions. The agent succeeds when a goal state is reached (\textit{the sought item is in hand}); it fails when it cannot possibly reach goal state.

Like prior work in LLM generating PDDL, we assume that a domain file that defines the available actions is provided; this domain file can solve a problem file that defines the initial and goal entity states (\textit{where the agent is, where the item is, how are these two locations connected}) when possible to result in a plan (\textit{go west, pick up item}). Unlike prior work, it is impossible to initially construct a complete problem file that is necessary for reaching the goal until the agent has explored enough to have uncovered all necessary entity states (\textit{we don't know where the item is until we find it}). Therefore, our \lego assumes a plannable sub-goal for which partial plans can be found to make progress and iteratively build a problem file based on the new observations. Eventually, the problem is sufficiently defined and the final-goal can be planned for (Figure~\ref{fig:pipeline}).

Formally, the agent is initially in state $s_1$ and presented with the first observation $o_1$. The agent then construct an initial problem file $PF_1$ to plan for the final-goal $G$. 
\begin{equation}
    PF_1 = \{LLM(o_1), G\}
\end{equation}
If this problem file can be solved by the provided domain file with a solver, a plan containing one or more actions is found. 
\begin{equation}
    Plan_1 := (a_1^1, a_1^2, \dots) = solver(DF,PF_1)
\end{equation}
If a plan cannot be found due to a lack of information in the problem file, the goal $G$ is swapped out by an immediate sub-goal $G'$, and the solver retries. The actions in the plan are then sequentially executed, resulting in a list of new observations.
\begin{equation}
    (o_2^1, o_2^2, \dots) = exec(s_1, a_1^1, a_1^2, \dots)
\end{equation}
Thus begins the second iteration. Using the new observations, the previous problem file is regenerated (referred to as \textbf{PDDL-gen}).
\begin{equation}
    PF_2 = \{LLM(PF_1, o_2), G\}
\end{equation}
The process goes on until one observation fulfills the termination condition.

Unlike prior work that generates the problem file once, \lego's having LLMs iteratively generating the problem file often result in inconsistencies and errors (e.g., missing a connectivity relation between two rooms, using the name a room in a relation without declaring the room, missing a parenthesis, etc.). To tackle this, we have the LLMs only predict the change in the problem file (i.e., the change of entity states), which we deterministically applied to the previous problem file (referred to as \textbf{PDDL-edit}).
\begin{equation}\tag{4'}
    \Delta_{2} = LLM(PF_1, o_2),\ PF_2 = PF_1 + \Delta_{2}
\end{equation}

We will compare our two approaches above with the baseline where LLMs directly generate an action (referred to as \textbf{Action-gen}).
\begin{equation}\tag{2'}
    Plan_i = LLM(o_i)
\end{equation}

\paragraph{Environments}
We experiment with two goal-oriented, partially-observed simulated environments, or text games, that span a variety of difficulty and flavor. 

\noindent \textbf{Coin Collector} \cite{yuan2019counting} focuses on navigation, which is an indispensable element of most simulations. The agent's task is to explore rooms, some connected by locked doors, and find a coin, similar to the running example above. Just as previously discussed, the previous approach on generating a PDDL problem file cannot be applied to Coin Collector because the location of the coin is unknown until the agent enters the same room as the coin. Therefore, the sub-goal structure for this tasks is defined as:
\begin{enumerate}[topsep=0pt,itemsep=-1ex,partopsep=1ex,parsep=1ex]
    \item pick up coin (requires the location of the coin)
    \item go to a room that has not been visited (reveals location of the coin)
\end{enumerate}

The sub-goal of ``going to an unvisited room'' results in monotonously increasing progress to the final-goal of ``finding the coin''. In similar search-related tasks, this singular sub-goal or strategy suffices, though it may not work for all situations.

\noindent \textbf{Cooking World} \cite{ijcai2020p207} subsumes Coin Collector with more complex tasks. The agent' task is to first explore rooms to find ingredients required by a recipe, much like Coin Collector. Next, it should cook the ingredient in some specified location using some specified appliance. Finally, when all ingredients are cooked correctly, a meal can be successfully prepared. Therefore, the sub-goal structure for this tasks is defined as:
\begin{enumerate}[topsep=0pt,itemsep=-1ex,partopsep=1ex,parsep=1ex]
    \item prepare meal (requires having obtained each ingredient and located relevant appliances)
    \item pick up each ingredient (requires the location of each ingredient; obtains ingredients)
    \item go to a room that has not been visited (reveals location of ingredients and appliances)
\end{enumerate}

\begin{table*}[]
\centering
\small
\resizebox{\columnwidth}{!}{%
\begin{tabular}{l|l|lll|lll}
\toprule
         & \multicolumn{1}{l|}{random} & \multicolumn{3}{c|}{GPT 3.5 Turbo} & \multicolumn{3}{c}{GPT 4 Turbo} \\ \midrule
         & & Action-gen  & PDDL-gen$^\dag$  & PDDL-edit$^\dag$ & Action-gen    & PDDL-gen$^\dag$   & PDDL-edit$^\dag$   \\
Coin    & 4\%  & 68\% & 26\% & 28\% & \textbf{94\%} & 58\% & 78\%    \\
Cooking (easy) & 0\% & 0\% & 70\% & 68\% & 4\% & 94\% & \textbf{98\%} \\
Cooking (hard) & 0\% & 0\% & 4\%  & 6\% & 0\% & 16\% & \textbf{46\%}            \\\bottomrule
\end{tabular}}
\caption{The percentage where the agent succeeds by taking no more than the maximum steps on the test set. The $^\dag$ sign specifies methods under our proposed \lego methodology. }
\label{tab:planning_results}
\end{table*}

\lstset{
  breaklines=true
}

\begin{figure*}
    \centering
    \small
    \begin{lstlisting}
< examine cookbook
> Gather all following ingredients and follow the directions to prepare this tasty meal.
Ingredients:
  black pepper, block of cheese, salt, red apple, yellow potato
Directions:
  slice the block of cheese, dice the red apple, chop the yellow potato, grill the yellow potato, prepare meal
< look around
> You are in the kitchen. In one part of the room you see a stove. There is also an oven. You also see a fridge that is closed. In another part of the room you see a counter that has a knife, a red apple, a raw yellow potato, and a cookbook on it. In one part of the room you see a kitchen cupboard that is closed. There is also a cutlery drawer that is closed. You also see a trash can that is closed. In another part of the room you see a dishwasher that is closed. In one part of the room you see a dining chair, that has nothing on it. 
To the South you see a closed plain door. To the West you see the corridor. 
< open fridge
> You open the fridge. The fridge contains a block of cheese.
< take block of cheese
> You take the block of cheese.
< take knife
> You take the knife.
< take red apple
> You take the red apple.
< take yellow potato
> You take the yellow potato.
< open kitchen cupboard
> You open the kitchen cupboard. It's empty inside.
< move west
> You are in the corridor. In one part of the room you see a key holder, that has nothing on it. There is also a shoe cabinet that is closed. You also see a umbrella stand, that has nothing on it. In another part of the room you see a hat rack, that has nothing on it. In one part of the room you see a coat hanger, that has nothing on it. 
To the North you see a closed wood door. To the East you see the kitchen. To the West you see a closed sliding patio door. 
< open shoe cabinet
> You open the shoe cabinet. It's empty inside.
< open door to north
> You open the wood door, revealing the bedroom. 
< move north
> You are in the bedroom. In one part of the room you see a dressing table, that has nothing on it. There is also a desk chair, that has nothing on it. You also see a desk, that has nothing on it. In another part of the room you see a chest of drawers that is closed. In one part of the room you see a wardrobe that is closed. There is also a night stand, that has nothing on it. You also see a bed, that has nothing on it. 
Through an open wood door, to the South you see the corridor. 
    \end{lstlisting}
\end{figure*}

\begin{figure*}
    \centering
    \small
    \begin{lstlisting}
< open chest of drawers
> You open the chest of drawers. It's empty inside.
< open wardrobe
> You open the wardrobe. It's empty inside.
< move south
> You are in the corridor. (description omitted)
< open door to west
> You open the sliding patio door, revealing the backyard. 
< move west
> You are in the backyard. In one part of the room you see a barbeque. There is also a workbench, that has nothing on it. You also see a patio chair, that has nothing on it. In another part of the room you see a patio table, that has nothing on it. In one part of the room you see a clothes line, that has nothing on it. There is also a garden, that has nothing on it. 
Through an open sliding patio door, to the East you see the corridor. 
< move east < move east
> You are in the kitchen. (description omitted)
< open door to south
> You open the plain door, revealing the pantry. 
< move south
> You are in the pantry. In one part of the room you see a folding chair, that has nothing on it. There is also a shelf that has some black pepper, and some salt on it. 
Through an open plain door, to the North you see the kitchen. 
< take black pepper
> You take the black pepper.
< take salt
> You take the salt.
< dice red apple
> You dice the red apple.
< slice block of cheese
> You slice the block of cheese.
< chop yellow potato
> You chop the yellow potato.
< move north
> You are in the kitchen. (description omitted)
Through an open plain door, to the South you see the pantry. To the West you see the corridor. 
< move west < move west
> You are in the backyard. (description omitted)
< cook yellow potato in barbeque
> You grill the yellow potato with the barbeque.
< move east < move east
> You are in the kitchen. (description omitted)
< prepare meal
> Adding the meal to your inventory.
< eat meal
> You eat the meal.  It is delicious.
    \end{lstlisting}
    \caption{An example trajectory performed by GPT 4 Turbo and PDDL-edit on Cooking World (hard).}
    \label{fig:cooking_trajectory}
\end{figure*}

To better understand these simulations, an example trajectory for Cooking World is shown in Figure~\ref{fig:cooking_trajectory}.

\paragraph{Evaluation}
For both simulations, we use the implementation from \citet{jansen2022textworldexpress}. For Coin Collector, we use the most complex setting; for Cooking World, we consider an easy and a hard setting with varying number of locations and ingredients. 

For the choice of LLM, we consider \texttt{gpt-3.5-turbo-1106} (GPT 3.5 Turbo) and \texttt{gpt-4-1106-preview} (GPT 4 Turbo) across baseline methods (i.e., Action-gen, PDDL-gen, and PDDL-edit). For Action-gen, we prompt the LLM with a full description of the simulation, and for PDDL methods, with a hand-annotated domain file containing well-defined actions. For the PDDL-edit setting, we prompt the LLM to generate templated edits (add, replace, and delete lines in the problem file). The prompt of each method include a 1-shot demonstration of the output format. 

Regarding \textbf{performance}, Table~\ref{tab:planning_results} shows a drastic performance degradation of Action-gen moving from Coin Collector (only 2 valid actions: move, open door each with 4 direction arguments) to the much more complex Cooking World (with 8 more actions with infinite possible arguments, like processing an ingredient). Moreover, in Cooking World, an agent would fail if an ingredient is processed incorrectly (e.g., fried instead of grilled, was not chopped before roasted). Therefore, LLMs generating actions on the fly are more likely to make irrevocable mistakes and fail the task. In contrast, our two-stage PDDL generation approaches ensure the correctness of the plan to process the ingredients (in the second stage) \textit{assuming} that the ingredients are gathered and that the appliances are identified (in the first stage). Logically, the failures of \lego indicates an inconsistency between the environmental observation and the problem file. For example, the connectivity of the rooms may not be updated correctly upon entrance to a new room, causing no plan or invalid plans to be found. By lessen the burden on LLMs, PDDL-edit notably ameliorates but cannot eliminate this issue. On Coin Collector, issues frequently arise in a loop, where opening a new door leads to a visited room. Notably, GPT3.5 is far worse than GPT4 in generating PDDL, in line with the observations by \citet{zhang2024proc2pddl} and \citet{silver2023generalized}. 

Regarding \textbf{efficiency}, Figure~\ref{fig:coin_step_count} shows that on Coin Collector, PDDL-edit is no less efficient than Action-gen on 7 out 8 examples (red crosses are often lower than the blue circles) in the development set where PDDL-edit terminates successfully. Scaling up to the entire test set, with GPT4, PDDL-edit has an average step to success of 7.8 compared to Action-gen's 13.6 among successful attempts, a 43\% improvement on efficiency. Among these steps, 3.3 of Action-gen are invalid (e.g., moving through a closed door) compared to merely 0.2 of \lego, a significant difference when trials and errors are expensive. \lego also shows better \textbf{stability}. In Figure~\ref{fig:coin_step_count}, PDDL-edit exhibits a much smaller variance across runs than Action-gen. For example, if the coin happens to be immediately to the west of the initial room, deciding to go west initially would result in a prompt success, while exploring the east portion initially would result in a notable detour. Our approach of PDDL generation leaves only the task of parsing environmental configuration to the LLM, while the planning task is done deterministically by the solver, leading to more consistent plans across runs.

\begin{figure}[t!]
    \centering
    \includegraphics[width=0.7\columnwidth]{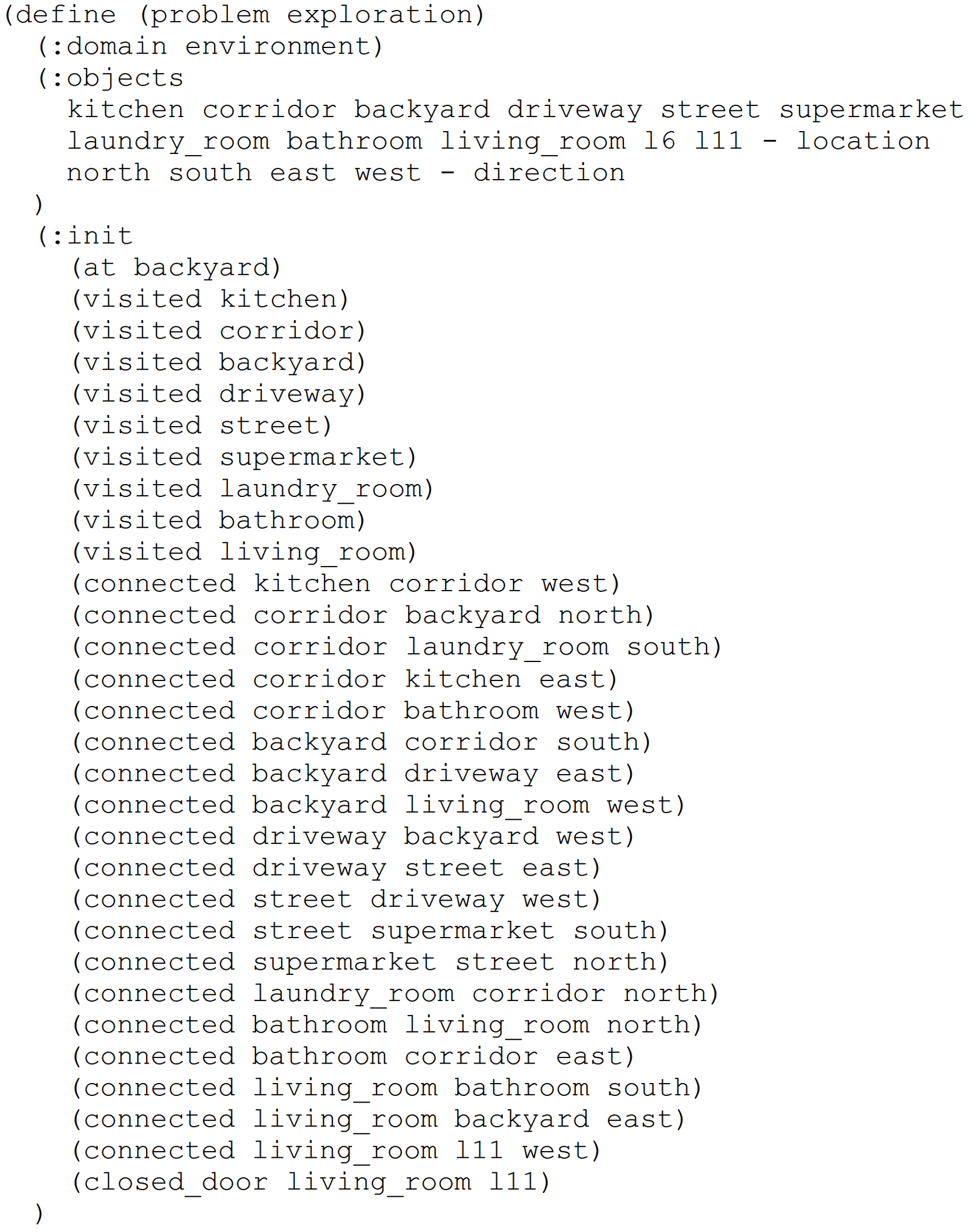}
    \caption{An example PDDL problem file learned throughout exploration in Coin Collector.}
    \label{fig:coin_pf}
\end{figure}

Regarding \textbf{interpretability} and \textbf{correctability}, the black-box nature of LLMs results in no faithful interpretation behind the decisions (c.f., thought-process). In Coin Collector, for example, if the coin has not be found at the maximum permitted steps, a problematic Action-gen trajectory is almost impossible to manually correct unless a human is to plot a map and keep track of the exploration. On the other hand, both PDDL-gen and PDDL-edit guarantees the correctness of the plan assuming that the generated or edited problem file is correct. Hence, upon failure, a human only needs to inspect and correct the most recent observation and the PDDL. For PDDL-edit, the job is even easier as only the change in the problem needs to be considered. An example learned problem file can be found in Figure~\ref{fig:coin_pf}.

\pgfplotstableread{
x         y    y-max  y-min
0  35.4 6.6   6.6
1 25 12.2   12.2
2     17.9 16.6   16.6
3  0 0   0
4 9.6 6.8   6.8 
5     3 0   0
6  10.6 10.9   10.9
7 12 10.6   10.6
8     20.4 7.3   7.3
9  0 0   0
}{\plangen}

\pgfplotstableread{
x         y    y-max  y-min
1 7 0   0
2     3.4 0.9   0.9
3  0 0   0
4 7 0   0
5     3 0   0
7 1 0   0
8 25 9.2 9.2
9  0 0   0
}{\pddlgen}

\begin{figure}
    \centering
    \begin{tikzpicture}[scale=0.9] 
    \begin{axis} [symbolic x coords={0,1,2,3,4,5,6,7,8, 9},xtick=data, xlabel={example ID}, ylabel={num. steps to success}]
    \addplot[only marks, mark=o, blue] 
      plot[error bars/.cd, y dir=both, y explicit]
      table[x=x,y=y,y error plus expr=\thisrow{y-max},y error minus expr=\thisrow{y-min}] {\plangen};
    \addplot[only marks, mark=x, red] 
      plot[error bars/.cd, y dir=both, y explicit]
      table[x=x,y=y,y error plus expr=\thisrow{y-max},y error minus expr=\thisrow{y-min}] {\pddlgen};
    \legend{Action-gen,PDDL-edit}
    \end{axis}
    \end{tikzpicture}
    \caption{On Coin Collector, the mean and standard deviation of number of steps to success (less is better) for each development example, each over 5 trials with different random seeds of \texttt{gpt-4-1106-preview}, comparing Action-gen and PDDL-edit. The error bar represents the sample standard deviation. On example 0 and 6, PDDL-edit fails and thus not shown.}
    \label{fig:coin_step_count}
\end{figure}

In conclusion, we propose \lego, the first approach to use LLMs to generate PDDL instead of actions while exploring partially-observed environments. We quantitatively show the improvement of performance, efficiency and stability, while qualitatively argue the benefit of interpretability and correctability by construction of this method. Future work should improve its flexibility by automatically predicting, instead of annotating, the domain file and the sub-goal hierarchy of each task, and show such flexibility on more involved simulations.

The work above was published in \citet{zhang-etal-2024-pddlego}, in which I formulated the task and primarily contributed to all components. I have obtained approval from all collaborators to exclusively include this work in this thesis. 

\section{Summary}

In this chapter, I introduced a neurosymbolic approach to use LLMs to generate a structured representation, before that representation is executed by a symbolic solver. In the context of planning, LLMs are only in charge of world modeling but not deciding upon the actions. Compared to the semi-symbolic event-entity schema discussed in Chapter~\ref{chap:entity}, this pipeline promises more determinism and interpretability by having the symbolic tools play a bigger role. For tasks where symbolic deduction plays a significant role, such as kinship deduction and classical planning discussed in this chapter, I have demonstrated how the synergy of LLMs and symbolic solvers lead to improved performance.

The ability of LLMs to generate executable structured representations (e.g., PDDL, Python) offers an exciting outlook for neurosymbolic methods as a whole that can tackle many tasks that include but are not limited to event reasoning. However, I have also shown significant challenges regarding the struggle of generating low-resource domain-specific languages (syntactically correct) as well as maintaining the consistency of the world representation during iterative generation (semantically correct). Furthermore, it awaits to be tested how well LLMs can generate such representation in a variety of domains.

\chapter{Conclusion}

This thesis has focused on automatically reasoning about events, a line of work that spans many fronts of efforts in NLP including question answering, commonsense inference, symbolic reasoning, etc. I start by establishing that LLMs are suitable tools for event reasoning due to their strong ability of cross-domain adaptation. I then systematically explore LLMs' failure on a variety of tasks and datasets. To address this, I propose the general methodology to combine LLMs with some structured event representation. I introduce three types of such representation depending on various factors such as the task, the capability of the LLMs, and the need for flexibility versus determinism. Each of these representations is used as a component of a pipeline with models including but not limited to LLMs. Across a variety of tasks, I have demonstrated the benefit of the synergy between LLMs and structured representations. Notably are the following takeaways. 

\minisection{Event reasoning is semi-symbolic in nature.} The choice to study the task of event reasoning in this thesis is practically motivated. However, event reasoning is representative of a superset of semi-language, semi-symbolic tasks (examples: the math question and relational inference tasks described in Section~\ref{sec:fcot}; non-examples: creative writing, summarization, emotion detection) that are ubiquitous and challenging in real-life scenarios. When using generative models (e.g., LLMs) for these tasks, we practitioners must decide the trade-off between language-modeling and symbol-modeling. This trade-off usually depends on the resources provided and the metric to optimize for. For example, to answer the question ``\textit{Is it safe to touch the pan?}'' in Chapter~\ref{chap:entity}, one may fully rely on an unstructured data representation and end-to-end LLMs if the LLMs are capable, the example is well-represented in the their training data, and stochasticity is acceptable. However, one may be better off decomposing the question into symbols as a multi-hop reasoning problem if the LLMs cannot directly answer the question, the example is out-of-distribution of the training data, and determinism must be guaranteed. 

\minisection{The long-tail problem thwarts fast-advancing models.} The work discussed in this thesis was done over a span of five years (2019 to 2024) where models like LLMs have witnessed extraordinary growth in capability. Despite their improvement, the strength and weakness of these data-driven models have roughly remained constant: they become increasingly capable with problems that are well-represented in the training (or pre-training) data, but remain challenged by those that are not. This is known as the long-tail problem. In event reasoning and similar tasks, the long-tail problem can emerge in two ways. The first is the context-wise complexity. In Chapter~\ref{chap:relation}, the questions are relatively short, generic (e.g., should one first \textit{use the washer} or \textit{user the dryer})), and heavily represented in the training data, so that they are more likely to be solved by larger-scale LLMs. In Chapter~\ref{chap:entity}, the questions become more specific (e.g., Is it \textit{safe to touch the pan} after doing step A, B, and C), while in Chapter~\ref{chap:world} even more so (e.g., find a plan to \textit{retrieve a coin} given a configuration of a maze). For AI technology to be more useful for a variety of users and circumstances (especially those who are underrepresented), the long-tail problem should remain in focus. In this thesis, in the context of using LLMs, I have explored both the in-context end-to-end approach and the structured approach, empirically arguing the latter is superior in many scenarios. 

\minisection{Neurosymbolic methods are empowered by code-trained LLMs.} My work described in Chapter~\ref{chap:entity} and~\ref{chap:world} are novel and special cases of neurosymbolic methods because it leverages LLMs to consider and predict the symbols. As discussed in those chapters and related work, this has not been possible until approximately 2021 when LLMs are trained to work with structured data such as code. Historically, the advantage of neurosymbolic methods is their precision and determinism while the disadvantage is their brittleness across domains. Similarly, historically, the advantage of data-driven neural methods is their flexibility and the disadvantage is their inability to work with structured data. In the present day, neurosymbolic methods do not just compete with, but are empowered by code-trained LLMs as symbol generators and validators. In Chapter~\ref{chap:world}), the preliminary results show that this line of working is promising despite challenges.

The three methodologies introduced in this thesis each call for future work that includes examining more domains, integrating future LLMs, and so on. However, perhaps a even more crucial direction is to develop a ``hierarchical organizer'' that can reduce the amount of hard-coding when designing these pipelines. For example, a high-level switch may decide to use the neurosymbolic usage for a task where symbolic reasoning plays a significant role. Then, a mid-level switch may decide what to symbolically model and what to rely on end-to-end generation. Finally, a low-level switch may decide on implementation details, such as whether to include techniques like chain-of-thought or self-refine. Though ambitious, the prospect of hierarchical planning and problem solving will surely lead to AI systems' ability and versatility.

\end{mainf}

    

\begin{bibliof}
\bibliography{bibliography,anthology}
\end{bibliof}
\end{document}